\documentclass[journal, final]{IEEEtran}

\IEEEoverridecommandlockouts

\usepackage[utf8]{inputenc} 
\usepackage[T1]{fontenc}
\usepackage{microtype}
\usepackage{textcomp}
\usepackage{xcolor}

\usepackage{amsmath, amssymb, amsfonts, amstext, amsthm, amsbsy}
\usepackage{mathtools}
\usepackage{mathrsfs}
\usepackage{bm}
\usepackage{dsfont}
\usepackage{stmaryrd}
\usepackage{xfrac}
\usepackage{nicefrac}
\usepackage{relsize}
\allowdisplaybreaks[1]

\usepackage{algorithm}
\usepackage{algpseudocode}

\usepackage{graphicx}
\usepackage{subcaption}
\usepackage{caption}
\usepackage{booktabs}
\usepackage{multirow}
\usepackage{array}
\usepackage{tabularx}
\usepackage{longtable}
\usepackage{makecell}
\usepackage{colortbl}
\usepackage{float}

\usepackage{setspace}
\usepackage{cuted}
\usepackage{multicol}
\usepackage{enumitem}
\usepackage{footmisc}
\usepackage{chngcntr}
\usepackage{tocloft}
\usepackage{sansmath}

\usepackage{tikz}
\usetikzlibrary{calc}

\usepackage{cite}
\usepackage{doi}
\usepackage{bibunits}
\defaultbibliographystyle{IEEEtran}
\defaultbibliography{references}

\usepackage{hyperref}
\hypersetup{
    colorlinks=true,
    linkcolor={red!50!black},
    citecolor={blue!50!black},
    urlcolor={black}
}

\DeclareMathAlphabet{\mathpzc}{OT1}{pzc}{m}{it}

\def\BibTeX{{\rm B\kern-.05em{\sc i\kern-.025em b}\kern-.08em
    T\kern-.1667em\lower.7ex\hbox{E}\kern-.125emX}}

\def\H{\mathrm{H}} 
\def\I{\mathrm{I}} 
\def\D{\mathrm{D}} 

\newcommand{\orcid}[1]{%
    \href{https://orcid.org/#1}{\includegraphics[width=10pt]{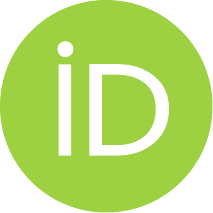}}%
}

\newtheorem{theorem}{Theorem}

\newtheorem{definition}{Definition}

\newtheorem{remark}{Remark}

\def\markov{\hbox{$\--$}\kern-1.5pt\hbox{$\circ$}\kern-1.5pt\hbox{$\--$}}

\makeatletter
\newcommand*{\centernot}{\mathpalette\@centernot}
\def\@centernot#1#2{%
  \mathrel{%
    \rlap{%
      \settowidth\dimen@{$\m@th#1{#2}$}%
      \kern.5\dimen@
      \settowidth\dimen@{$\m@th#1=$}%
      \kern-.5\dimen@
      $\m@th#1\not$%
    }%
    {#2}%
  }%
}
\makeatother

\renewcommand\thesection{\arabic{section}}
\renewcommand\thesubsection{\arabic{section}.\arabic{subsection}}

\def\thesectiondis{\thesection.}
\def\thesubsectiondis{\thesectiondis\arabic{subsection}.}

\makeatletter
\def\@seccntformatinl#1{\csname the#1dis\endcsname\hskip 1em\relax}
\makeatother

\makeatletter
\newcommand*{\rom}[1]{\expandafter\@slowromancap\romannumeral #1@}
\makeatother

\definecolor{headercolor}{gray}{0.82}
\definecolor{zebracolor}{gray}{0.95}
\definecolor{lightgray}{gray}{0.93}

\makeatletter
\newcommand{\AppendixOnlyTOC}{%
  \section*{Appendix Contents}%
  \@starttoc{atoc}%
}
\makeatother

\newcommand{\appsection}[1]{%
  \section{#1}%
  \addcontentsline{atoc}{section}{\protect\numberline{\thesection}#1}%
}

\newcommand{\appsubsection}[1]{%
  \subsection{#1}%
  \addcontentsline{atoc}{subsection}{\protect\numberline{\thesubsection}#1}%
}

\begin{document}

\begin{bibunit}

\title{Deep Privacy Funnel Model:\\From a Discriminative to a Generative Approach\\with an Application to Face Recognition}

\author{Behrooz~Razeghi~$\!{\orcid{0000-0001-9568-4166}}$ $\!^\ast$,~
			Parsa~Rahimi~$\!{\orcid{0000-0001-7927-268X}}$,~
			and~S\'{e}bastien~Marcel~$\!{\orcid{0000-0002-2497-9140}}$~
\thanks{$^\ast$ Corresponding Author.}
\thanks{B.~Razeghi is with the Harvard University, US (e-mail: behroozrazeghi@seas.harvard.edu); work done while at the Idiap Research Institute, Switzerland.
P.~Rahimi and S.~Marcel are with the Idiap Research Institute, Switzerland (e-mail:~\{parsa.rahimi, sebastien.marcel\}@idiap.ch).
P.~Rahimi is also with the École Polytechnique Fédérale de Lausanne (EPFL), Switzerland.
S.~Marcel is also with the Université de Lausanne, Switzerland.}%
\thanks{This manuscript is an extended version of our paper accepted in 2024 IEEE International Conference on Acoustics, Speech, and Signal Processing~\cite{razeghi2024dvpf}.}
\thanks{\urlstyle{sf}The source code is publicly available at \url{https://github.com/BehroozRazeghi/DeepPrivacyFunnelModel}.}
}

\maketitle

%
\begin{abstract}
In this study, we apply the information-theoretic Privacy Funnel (PF) model to face recognition and develop a method for privacy-preserving representation learning within an end-to-end trainable framework. Our approach addresses the trade-off between utility and obfuscation of sensitive information under logarithmic loss. We study the integration of information-theoretic privacy principles with representation learning, with a particular focus on face recognition systems. We also highlight the compatibility of the proposed framework with modern face recognition networks such as AdaFace and ArcFace.
In addition, we introduce the Generative Privacy Funnel ($\mathsf{GenPF}$) model, which extends the traditional discriminative PF formulation, referred to here as the Discriminative Privacy Funnel ($\mathsf{DisPF}$). The proposed $\mathsf{GenPF}$ model extends the privacy-funnel framework to generative formulations under information-theoretic and estimation-theoretic criteria. Complementing these developments, we present the deep variational PF (DVPF) model, which yields a tractable variational bound for measuring information leakage and enables optimization in deep representation-learning settings. The DVPF framework, associated with both the $\mathsf{DisPF}$ and $\mathsf{GenPF}$ models, also clarifies connections with generative models such as variational autoencoders (VAEs), generative adversarial networks (GANs), and diffusion models.
Finally, we validate the framework on modern face recognition systems and show that it provides a controllable privacy--utility trade-off while substantially reducing leakage about sensitive attributes. To support reproducibility, we also release a PyTorch implementation of the proposed framework.
\end{abstract}

\section{Introduction}
\label{Sec:Introduction}

\vspace{-5pt}

\IEEEPARstart{I}{n} face recognition, an important challenge is to balance privacy preservation with utility. This challenge is particularly relevant in representation learning, where improving privacy often comes at the cost of reducing the usefulness of the learned representation for downstream tasks. Existing privacy-preserving representation-learning approaches for face recognition do not explicitly characterize this privacy--utility trade-off from an information-theoretic perspective. This limitation motivates the development of methods for \textit{identifying}, \textit{quantifying}, and \textit{mitigating} privacy risks in face recognition systems.

Our work studies this problem through the lens of the information-theoretic Privacy Funnel (PF) model applied to face recognition systems. We develop an end-to-end framework for privacy-preserving representation learning, in which the privacy--utility trade-off is quantified under logarithmic loss. The formulation can also be extended to other loss functions on positive measures. This provides a principled way to connect information-theoretic privacy with representation learning in face recognition.
The proposed framework is compatible with recent face recognition architectures, including AdaFace and ArcFace, and can therefore be integrated with current face recognition pipelines.

We further introduce the Generative Privacy Funnel ($\mathsf{GenPF}$) model and the Deep Variational Privacy Funnel (DVPF) framework. The $\mathsf{GenPF}$ model extends the Privacy Funnel formulation to a generative setting. The DVPF framework introduces a variational bound on the information-leakage term, which makes the Privacy Funnel objective tractable in deep representation learning. The proposed framework can also be combined with prior-independent privacy-enhancing mechanisms, such as differential privacy, thereby allowing prior-dependent and prior-independent protections to be used jointly.
The proposed framework supports both raw-image and embedding-based inputs. In the present paper, however, we focus on a controlled embedding-based plug-and-play setting in which pre-trained recognition backbones are kept fixed and the privacy module is learned on top of the extracted embeddings. Raw-image and fine-tuning scenarios are supported by the general framework, but are not studied exhaustively here.

%
\begin{figure*}[t]
    \centering
    \begin{subfigure}[h]{0.38\textwidth}
        \includegraphics[height= 3.1cm]{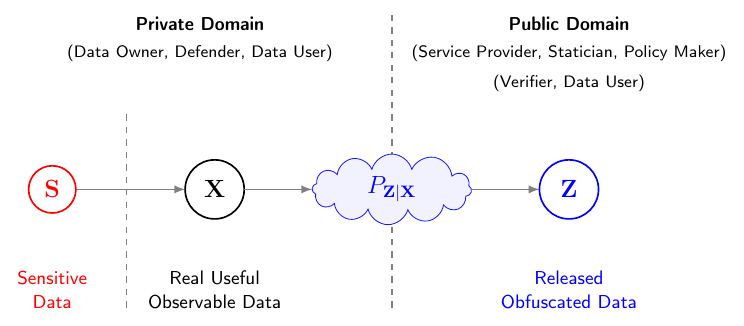}%
        \vspace{-1pt}
        \caption{}
        \label{fig:discriminativePF_setup}
    \end{subfigure}%
~~~~~~~~~~~~
    \begin{subfigure}[h]{0.38\textwidth}
        \includegraphics[height= 3.1cm]{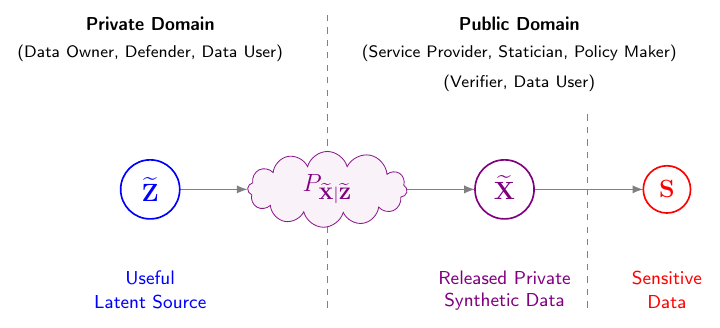}%
        \vspace{-1pt}
        \caption{}
        \label{fig:generativePF_setup}
    \end{subfigure}
    \vspace{-6pt}
    \caption{High-level schematic comparison of privacy funnel models: \textbf{(a)} discriminative paradigm; \textbf{(b)} generative paradigm.}
    \vspace{-7pt}
    \label{Fig:Fist_diagram_DisPF_GenPF}
\end{figure*}

Our work is connected to two main research directions: privacy funnel methods and disentangled representation learning. In the privacy funnel literature, existing work includes methods that reduce leakage of sensitive information as well as optimization-based approaches for solving privacy funnel formulations more efficiently, such as the difference-of-convex method in \cite{huang2024efficient}; see also \cite{de2022funck, huang2024efficient}. In disentangled representation learning, several related works address representation control and bias mitigation. For example, \cite{tran2017disentangled} studies disentangled representations for pose variation, \cite{gong2020jointly} considers bias mitigation across demographic groups, \cite{park2021learning} develops a model for reducing AI discrimination while preserving task-relevant information, and \cite{li2022discover} proposes DebiAN, which mitigates bias without using protected-attribute labels. In a related direction, \cite{suwala2024face} introduces PluGeN4Faces for facial attribute manipulation with identity preservation.
For extended discussion see Appendix~\ref{AppxSec:Navigating_DataPrivacy_Paradigm} and Appendix~\ref{AppxSec:Preliminaries}.

%
%

\subsection{Key Contributions}
\label{ssec:Contributions}

\vspace{-1pt}

Our research makes the following contributions to the field:
\begin{itemize}[leftmargin=1em]
\item 
\textbf{Privacy Funnel Modeling for Face Recognition:}
We study privacy-preserving representation learning for face recognition using the information-theoretic PF model. To the best of our knowledge, this is among the first end-to-end PF-based formulations developed for modern face recognition pipelines. The framework is compatible with recent state-of-the-art face recognition architectures, including ArcFace \cite{arcface2019} and AdaFace \cite{kim2022adaface}.
\item 
\textbf{Generative Privacy Funnel Model:}
We introduce the Generative Privacy Funnel ($\mathsf{GenPF}$) model as a generative extension of the standard Privacy Funnel formulation, which we denote by the Discriminative Privacy Funnel ($\mathsf{DisPF}$). This formulation provides a framework for studying privacy-preserving data generation under information-theoretic and estimation-theoretic criteria. We further study a specific $\mathsf{GenPF}$ formulation in the context of face recognition.
%
\item
\textbf{Deep Variational Privacy Funnel Framework:}
We develop the Deep Variational Privacy Funnel (DVPF) framework for privacy-preserving representation learning. The framework introduces a tractable variational treatment of the information-leakage term, which makes the Privacy Funnel objective amenable to optimization in deep models. We also discuss its connections to common generative-modeling frameworks, including VAEs, GANs, and diffusion-based models.
Furthermore, we have applied the DVPF model to the advanced face recognition systems.
%
\end{itemize}

\vspace{-7pt}

%
%
\subsection{Outline}
\label{ssec:Outline}

\vspace{-1pt}

In Sec.~\ref{Sec:PF_Model}, we present the discriminative and generative perspectives of the PF model. We then present the deep variational PF model in Sec.~\ref{Sec:DeepVariationalPF_Model}. 
Experimental results are provided in Sec.~\ref{Sec:Experiments}. Finally, conclusions are drawn in Sec.~\ref{Sec:Conclusion}. 

\vspace{-7pt}

%
%
\subsection{Notations}

\vspace{-1pt}

Throughout this paper, random variables are denoted by capital letters (e.g., $X$, $Y$), deterministic values are denoted by small letters (e.g., $x$, $y$), random vectors are denoted by capital bold letter (e.g., $\mathbf{X}$, $\mathbf{Y}$), deterministic vectors are denoted by small bold letters (e.g., $\mathbf{x}$, $\mathbf{y}$), alphabets (sets) are denoted by calligraphic fonts (e.g., $\mathcal{X, Y}$), and for specific quantities/values we use sans serif font (e.g., $\mathsf{x}$, $\mathsf{y}$, $\mathsf{C}$, $\mathsf{D}$). Also, we use the notation $\left[ N \right]$ for the set $\{ 1, 2, \cdots, N\}$. $\H \left( P_{\mathbf{X}} \right) \coloneqq \mathbb{E}_{P_{\mathbf{X}}} \left[ - \log P_{\mathbf{X}} \right]$ denotes the Shannon entropy; $\H \left( P_{\mathbf{X}} \Vert Q_{\mathbf{X}} \right) \coloneqq \mathbb{E}_{P_{\mathbf{X}}} \left[ - \log Q_{\mathbf{X}} \right]$ denotes the cross-entropy of the distribution $P_{\mathbf{X}}$ relative to a distribution $Q_{\mathbf{X}}$; and $\H \left( P_{\mathbf{Z} \mid \mathbf{X}} \Vert Q_{\mathbf{Z} \mid \mathbf{X}} \mid P_{\mathbf{X}}  \right) \coloneqq \mathbb{E}_{P_{\mathbf{X}}}  \mathbb{E}_{P_{\mathbf{Z} \mid \mathbf{X} }} \left[ - \log Q_{\mathbf{Z} \mid \mathbf{X}} \right]$ denotes the cross-entropy loss for $Q_{\mathbf{Z} \mid \mathbf{X}}$. The relative entropy is defined as $\D_{\mathrm{KL}} \left( P_{\mathbf{X}} \Vert Q_{\mathbf{X}} \right)  \coloneqq \mathbb{E}_{P_{\mathbf{X}}} \big[ \log \frac{P_{\mathbf{X}}}{Q_{\mathbf{X}}} \big] $. The conditional relative entropy is defined by $\D_{\mathrm{KL}} \left( P_{\mathbf{Z} \mid \mathbf{X}} \Vert  Q_{\mathbf{Z} \mid \mathbf{X}}  \mid P_{\mathbf{X}} \right) \coloneqq  \mathbb{E}_{P_{\mathbf{X}}} \left[ \D_{\mathrm{KL}} \left( P_{\mathbf{Z} \mid \mathbf{X}= \mathbf{x}} \Vert  Q_{\mathbf{Z} \mid \mathbf{X} = \mathbf{x}} \right)  \right]$ and the mutual information is defined by $\I \left( P_{\mathbf{X}}; P_{\mathbf{Z} \mid \mathbf{X}} \right)  \coloneqq \D_{\mathrm{KL}} \left( P_{\mathbf{Z} \mid \mathbf{X}} \Vert  P_{\mathbf{Z} }  \mid P_{\mathbf{X}} \right)$. We abuse notation to write $\H \left( \mathbf{X} \right) = \H \left( P_{\mathbf{X}} \right)$ and $\I \left( \mathbf{X}; \mathbf{Z} \right) = \I \left( P_{\mathbf{X}}; P_{\mathbf{Z} \mid \mathbf{X}} \right)$ for random objects $\mathbf{X} \sim P_{\mathbf{X}}$ and $\mathbf{Z} \sim P_{\mathbf{Z}}$. We use the same notation for the probability distributions and the associated densities. 
%


\vspace{-5pt}

\section{Privacy Funnel Model:\\Discriminative and Generative Paradigms}
\label{Sec:PF_Model}

%
\begin{figure*}[!t]
    \centering
        \begin{subfigure}[h]{0.3\textwidth}
        \includegraphics[scale=0.522]{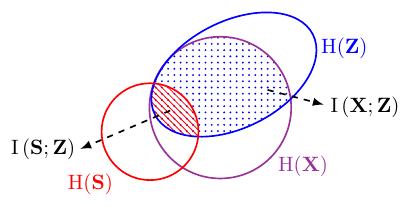}%
          \vspace{-7pt}
        \caption{}
        \label{fig:Idiagram1}
    \end{subfigure}%
~
      \begin{subfigure}[h]{0.3\textwidth}
        \includegraphics[scale=0.54]{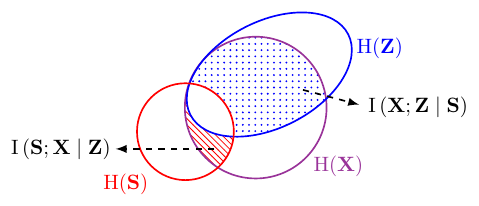}%
        \vspace{-7pt}
        \caption{}
        \label{fig:Idiagram2}
    \end{subfigure}
~~~~~~
      \begin{subfigure}[h]{0.3\textwidth}
        \includegraphics[scale=0.567]{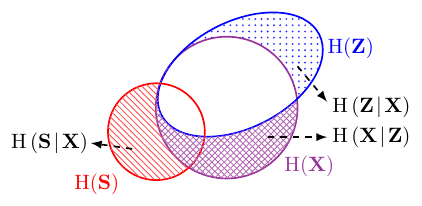}%
        \vspace{-7pt}
        \caption{}
        \label{fig:Idiagram3}
    \end{subfigure}
   \vspace{-6pt}
    \caption{Information diagrams for $ \mathbf{S}  \markov \mathbf{X} \markov \mathbf{Z}$. (a) entropy $\H \left(\mathbf{S}\right)$, $\H \left(\mathbf{X}\right)$, $\H \left(\mathbf{Z}\right)$, and preserved useful information in the disclosed representation $\I \left(\mathbf{X}; \mathbf{Z}\right)$
    and information leakage $\I \left(\mathbf{S}; \mathbf{Z}\right)$. (b) preserved useful non-sensitive information $\I \left(\mathbf{X}; \mathbf{Z} \mid \mathbf{S}\right)$ and residual sensitive information $\I \left(\mathbf{S}; \mathbf{X} \mid \mathbf{Z}\right)$. (c) sensitive attribute uncertainty $\H \left( \mathbf{S} \! \mid \! \mathbf{X} \right)$, useful information decoding uncertainty $\H \left( \mathbf{X} \! \mid \! \mathbf{Z} \right)$, and encoding uncertainty $\H \left( \mathbf{Z} \! \mid \! \mathbf{X} \right)$. 
    }
    \label{Fig:I-diagram}
   \vspace{-10pt}
\end{figure*}

\subsection{Measuring Privacy Leakage and Utility Performance}
\label{SubSec:MeasuringPrivacyLeakageUtilityPerformance}

\vspace{-1pt}

Let $(\mathbf{S},\mathbf{X}) \sim P_{\mathbf{S},\mathbf{X}}$, where $\mathbf{S}$ denotes sensitive information and $\mathbf{X}$ denotes useful or observable data. Any privacy mechanism that releases a variable $\mathbf{W}$ induces joint distributions $P_{\mathbf{S},\mathbf{W}}$ and $P_{\mathbf{X},\mathbf{W}}$. We measure privacy leakage through a privacy-risk functional $\mathcal{C}_{\mathsf{S}} : \mathcal{P}\!\left(\mathcal{S}\times\mathcal{W}\right)\rightarrow \mathbb{R}_{+}$, which quantifies the leakage about $\mathbf{S}$ contained in the released variable $\mathbf{W}$. Utility is quantified through a well-characterized and task-dependent functional $\mathcal{C}_{\mathsf{U}} : \mathcal{P}\!\left(\mathcal{X}\times\mathcal{W}\right)\rightarrow \mathbb{R}$, which evaluates how well $\mathbf{W}$ preserves the information in $\mathbf{X}$ that is relevant to the downstream task.
Depending on the sign convention, $\mathcal{C}_{\mathsf{U}}$ may be interpreted either as a utility reward to be maximized or as a utility loss to be minimized. In this work, we use the Shannon mutual information (MI) criterion, for which privacy leakage is measured by $\I(\mathbf{S};\mathbf{W})$ and utility is measured by $\I(\mathbf{X};\mathbf{W})$. 

\vspace{-4pt}

\subsection[Discriminative Privacy Funnel ($\mathsf{DisPF}$) Model]{Discriminative Privacy Funnel Model: Optimizing Information Extraction Under Privacy Constraints}

Given correlated random variables $\mathbf{S}$ and $\mathbf{X}$ with joint distribution $P_{\mathbf{S},\mathbf{X}}$, the objective in the classical discriminative PF method \cite{makhdoumi2014information} is to derive a representation $\mathbf{Z}$ of \textit{useful} data $\mathbf{X}$ through a stochastic mapping $P_{\mathbf{Z}\mid \mathbf{X}}$ such that: \textbf{(i)} $\mathbf{S} \markov \mathbf{X} \markov \mathbf{Z}$ forms a Markov chain; \textbf{(ii)} $\mathbf{Z}$ is maximally informative about $\mathbf{X}$; and \textbf{(iii)} $\mathbf{Z}$ is minimally informative about $\mathbf{S}$; see Fig.~\ref{fig:discriminativePF_setup}.

The classical PF method therefore characterizes the trade-off between privacy leakage $\I(\mathbf{S};\mathbf{Z})$ and revealed useful information $\I(\mathbf{X};\mathbf{Z})$. For a leakage budget $R^{\mathrm{s}}\geq 0$, this trade-off is given by
\vspace{-5pt}
\begin{align}
\label{Eq:DisPF_functional_MI}
\mathsf{DisPF\text{-}MI}\!\left(R^{\mathrm{s}},P_{\mathbf{S},\mathbf{X}}\right)
\coloneqq
\sup_{\substack{P_{\mathbf{Z}\mid \mathbf{X}}:\\ \mathbf{S}\markov \mathbf{X}\markov \mathbf{Z}}}
\quad & \I(\mathbf{X};\mathbf{Z}) \\
\mathrm{subject~to}\quad
& \I(\mathbf{S};\mathbf{Z}) \le R^{\mathrm{s}} . \nonumber
\end{align}
The $\mathsf{DisPF\text{-}MI}$ curve is obtained by varying $R^{\mathrm{s}}$ over its feasible range.
A standard scalarization of \eqref{Eq:DisPF_functional_MI} is obtained through the Lagrangian objective
\vspace{-3pt}
\begin{equation}
\label{Eq:PF_LagrangianFunctional}
\mathcal{L}_{\mathsf{DisPF\text{-}MI}}\!\left(P_{\mathbf{Z}\mid \mathbf{X}},\alpha\right)
\coloneqq
\I(\mathbf{X};\mathbf{Z})-\alpha\,\I(\mathbf{S};\mathbf{Z}),
\qquad \alpha \ge 0 .
\end{equation}

Yeung's $\I$-measure provides a set-theoretic representation of Shannon information quantities \cite{yeung1991new, razeghi2023bottlenecks}. Under the Markov constraint $\mathbf{S}\markov \mathbf{X}\markov \mathbf{Z}$, we have $\I(\mathbf{S};\mathbf{Z}\mid \mathbf{X})=0$. Hence, under the sign convention used here,
$\I(\mathbf{S};\mathbf{X};\mathbf{Z})
=
\I(\mathbf{S};\mathbf{Z})-\I(\mathbf{S};\mathbf{Z}\mid \mathbf{X})
=
\I(\mathbf{S};\mathbf{Z})
\ge 0$,
which is reflected by the corresponding $\I$-diagram in Fig.~\ref{Fig:I-diagram}.

\vspace{5pt}

\begin{figure*}[!t]
    \centering
    \begin{subfigure}[h]{0.35\textwidth}
        \includegraphics[width=\textwidth]{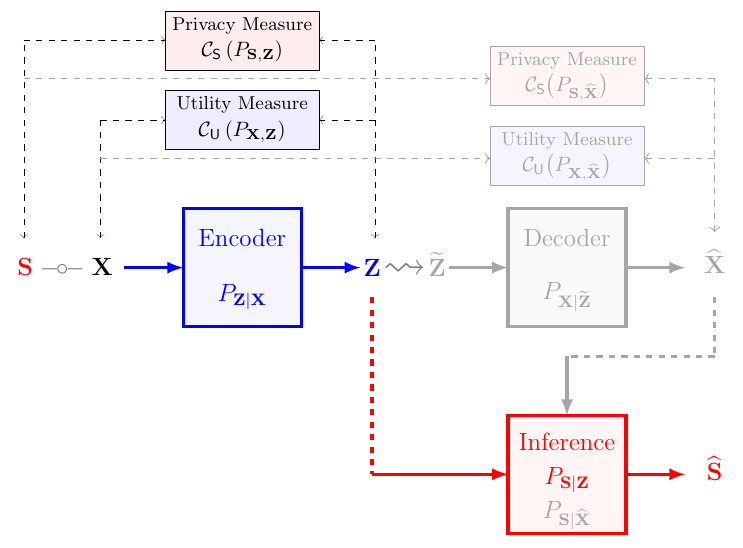}%
        \vspace{-2pt}
        \caption{Discriminative Privacy Funnel $\mathsf{DisPF}$.}
        \label{fig:discriminativePF_model}
    \end{subfigure}%
\hspace{70pt}
    \begin{subfigure}[h]{0.35\textwidth}
        \includegraphics[width=\textwidth]{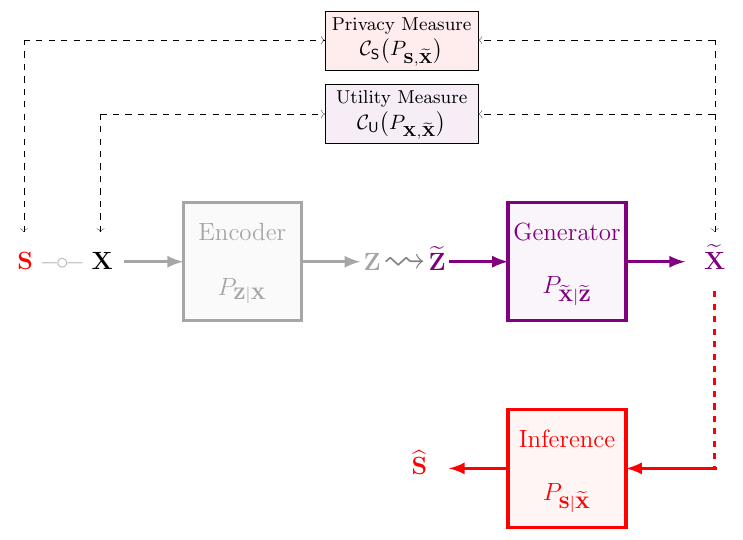}%
        \vspace{-2pt}
        \caption{Generative Privacy Funnel $\mathsf{GenPF}$.}
        \label{fig:generativePF_model}
    \end{subfigure}
    \caption{Comparative overview of generalized privacy funnel (PF) approaches: \textbf{(a)} the established discriminative (classical) model; \textbf{(b)} the proposed generative model.}
    \label{Fig:DisPF_GenPF_models}
\end{figure*}

\noindent
\textbf{Discriminative Privacy Funnel with General Loss Functions:}
Consider an extension of the standard discriminative PF objective to a broader class of loss functions. The goal of this general discriminative PF formulation is to obtain a representation $\mathbf{Z}$ of the \textit{useful} data $\mathbf{X}$ through a probabilistic mapping $P_{\mathbf{Z}\mid \mathbf{X}}$ (see Fig.~\ref{fig:discriminativePF_setup} and Fig.~\ref{fig:discriminativePF_model}). This objective is subject to the following requirements:\vspace{-1pt}
\begin{itemize}[leftmargin=2.2em]
    \item[(i)]
    The variables satisfy the Markov chain $\mathbf{S} \markov \mathbf{X} \markov \mathbf{Z}$.
    \item[(ii)]
    The utility loss $\mathcal{C}_{\mathsf{U}} \left( P_{\mathbf{X} , \mathbf{Z}} \right)$ is minimized, so that $\mathbf{Z}$ preserves the information in $\mathbf{X}$ that is relevant to the utility task.
    \item[(iii)]
    The privacy-risk functional $\mathcal{C}_{\mathsf{S}} \left( P_{\mathbf{S} , \mathbf{Z}} \right)$ is minimized, so that $\mathbf{Z}$ limits the leakage about the sensitive information~$\mathbf{S}$.
\end{itemize}
Equivalently, one may impose a constraint on the privacy-risk functional. Thus, for a given privacy budget $R^{\mathrm{s}} \geq 0$, the trade-off can be represented by the $\mathsf{DisPF}$ functional:\vspace{-3pt}
\begin{align}\label{Eq:DisPF_functional_general}
\mathsf{DisPF} \left( R^{\mathrm{s}}, P_{\mathbf{S}, \mathbf{X}}\right)
\coloneqq
\mathop{\inf}_{\substack{P_{\mathbf{Z} \mid \mathbf{X}}: \\ \mathbf{S} \markov \mathbf{X} \markov \mathbf{Z}}}
&\ \mathcal{C}_{\mathsf{U}} \left( P_{\mathbf{X} , \mathbf{Z}} \right) \\
\mathrm{subject~to} \quad
&\ \mathcal{C}_{\mathsf{S}} \left( P_{\mathbf{S} , \mathbf{Z}} \right) \leq R^{\mathrm{s}} . \nonumber
\end{align}
The MI formulation in \eqref{Eq:DisPF_functional_MI} is recovered by taking
$\mathcal{C}_{\mathsf{U}}\!\left(P_{\mathbf{X},\mathbf{Z}}\right) = -\I(\mathbf{X};\mathbf{Z})$
and 
$\mathcal{C}_{\mathsf{S}}\!\left(P_{\mathbf{S},\mathbf{Z}}\right)=\I(\mathbf{S};\mathbf{Z})$.

\vspace{3pt}

\begin{remark}
The stochastic mapping $P_{\mathbf{Z}\mid \mathbf{X}}$ may represent either a \textbf{domain-preserving transformation} or a \textbf{non-domain-preserving transformation}, as illustrated in Fig.~\ref{fig:discriminativePF_model}. In a domain-preserving transformation, such as image-to-image obfuscation, the released variable $\mathbf{Z}$ remains in the same domain as $\mathbf{X}$ but is modified to suppress sensitive information. In a non-domain-preserving transformation, such as image-to-embedding conversion, $\mathbf{Z}$ lies in a different representation space.
If a decoder is introduced, producing a reconstruction $\widehat{\mathbf{X}}$ from $\mathbf{Z}$, then utility and privacy should be evaluated on the variable that is actually used or released in the application. Accordingly, utility may be measured either through $\mathcal{C}_{\mathsf{U}}(P_{\mathbf{X},\mathbf{Z}})$ or, where applicable, after the decoding phase indicated in gray in Fig.~\ref{fig:discriminativePF_model}, through $\mathcal{C}_{\mathsf{U}}(P_{\mathbf{X},\widehat{\mathbf{X}}})$. Similarly, privacy leakage may be quantified either through $\mathcal{C}_{\mathsf{S}}(P_{\mathbf{S},\mathbf{Z}})$ or, in the decoded setting, through $\mathcal{C}_{\mathsf{S}}(P_{\mathbf{S},\widehat{\mathbf{X}}})$.
\end{remark}

%

\vspace{-5pt}
\vspace{-3pt}

\subsection[Generative Privacy Funnel ($\mathsf{GenPF}$) Model]{Generative Privacy Funnel Model: Optimizing Data Synthesis Under Privacy Constraints}

The generative PF ($\mathsf{GenPF}$) model addresses the problem of releasing synthetic data under explicit privacy constraints. Let $\widetilde{\mathbf{X}}$ denote the released synthetic data and let $\widetilde{\mathbf{Z}}$ denote a latent variable used by the synthetic mechanism. To define the induced joint laws $P_{\mathbf{X},\widetilde{\mathbf{X}}}$ and $P_{\mathbf{S},\widetilde{\mathbf{X}}}$, the generative mechanism must specify how $\widetilde{\mathbf{Z}}$ is coupled to the original data. In the general case, we therefore consider an encoder--generator construction of the form
$P_{\mathbf{S},\mathbf{X},\widetilde{\mathbf{Z}},\widetilde{\mathbf{X}}}
=
P_{\mathbf{S},\mathbf{X}}\,
P_{\widetilde{\mathbf{Z}}\mid \mathbf{X}}\,
P_{\widetilde{\mathbf{X}}\mid \widetilde{\mathbf{Z}}}$,
which induces the Markov chain
$\mathbf{S}\markov \mathbf{X}\markov \widetilde{\mathbf{Z}}\markov \widetilde{\mathbf{X}}$,
and hence also $\mathbf{S}\markov \mathbf{X}\markov \widetilde{\mathbf{X}}$.

The objective of the $\mathsf{GenPF}$ model is to generate synthetic data $\widetilde{\mathbf{X}}$ that preserve task-relevant information from the original data $\mathbf{X}$ while limiting leakage about the sensitive information $\mathbf{S}$; see Fig.~\ref{fig:generativePF_setup} and Fig.~\ref{fig:generativePF_model}. Using the general loss-function formalism introduced above, this objective is subject to the following requirements:
\begin{itemize}[leftmargin=2em]
    \item[(i)]
    The variables satisfy the Markov chain $\mathbf{S}\markov \mathbf{X}\markov \widetilde{\mathbf{Z}}\markov \widetilde{\mathbf{X}}$.
    \item[(ii)]
    The utility loss $\mathcal{C}_{\mathsf{U}}\big(P_{\mathbf{X},\widetilde{\mathbf{X}}}\big)$ is minimized, so that $\widetilde{\mathbf{X}}$ preserves the information in $\mathbf{X}$ that is relevant to the utility task.
    \item[(iii)]
    The privacy-risk functional $\mathcal{C}_{\mathsf{S}} \big(P_{\mathbf{S},\widetilde{\mathbf{X}}} \big)$ is minimized, so~that $\widetilde{\mathbf{X}}$ limits the leakage about the sensitive information $\mathbf{S}$.
\end{itemize}

Accordingly, for a given privacy budget $R^{\mathrm{s}} \geq 0$, the trade-off can be represented by the $\mathsf{GenPF}$ functional:
\vspace{-3pt}
\begin{align}
\label{Eq:GenPF_functional_general}
\mathsf{GenPF}\big(R^{\mathrm{s}},P_{\mathbf{S},\mathbf{X}}\big)
\coloneqq
\inf_{\substack{P_{\widetilde{\mathbf{Z}}\mid \mathbf{X}},\,P_{\widetilde{\mathbf{X}}\mid \widetilde{\mathbf{Z}}}:\\
\mathbf{S}\markov \mathbf{X}\markov \widetilde{\mathbf{Z}}\markov \widetilde{\mathbf{X}}}}
\quad &
\mathcal{C}_{\mathsf{U}}\big(P_{\mathbf{X},\widetilde{\mathbf{X}}}\big)
\\
\mathrm{subject~to}\quad
&
\mathcal{C}_{\mathsf{S}}\big(P_{\mathbf{S},\widetilde{\mathbf{X}}}\big)\le R^{\mathrm{s}} .
\nonumber
\end{align}

\vspace{-5pt}

\begin{remark}
As illustrated in Fig.~\ref{fig:generativePF_model}, the generative PF model may include an explicit encoding step, represented in gray, through the conditional distribution $P_{\widetilde{\mathbf{Z}}\mid \mathbf{X}}$. In this case, the released synthetic data are obtained by passing the encoded representation through the generator $P_{\widetilde{\mathbf{X}}\mid \widetilde{\mathbf{Z}}}$. More generally, the model may also operate directly from a latent prior when no encoder is used. In that case, however, samplewise utility criteria based on $P_{\mathbf{X},\widetilde{\mathbf{X}}}$ require an explicit coupling between the original and synthetic data.
\end{remark}

\noindent
\textbf{Generative Privacy Funnel with MI Criterion:}
When the synthetic mechanism induces a nontrivial coupling between $\mathbf{X}$ and $\widetilde{\mathbf{X}}$, a MI formulation~is
\begin{align}
\label{Eq:GenPF_functional_MI}
\mathsf{GenPF\text{-}MI}\!\left(R^{\mathrm{s}},P_{\mathbf{S},\mathbf{X}}\right)
\coloneqq
\sup_{\substack{P_{\widetilde{\mathbf{Z}}\mid \mathbf{X}},\,P_{\widetilde{\mathbf{X}}\mid \widetilde{\mathbf{Z}}}:\\
\mathbf{S}\markov \mathbf{X}\markov \widetilde{\mathbf{Z}}\markov \widetilde{\mathbf{X}}}}
\quad &
\I(\mathbf{X};\widetilde{\mathbf{X}})
\\
\mathrm{subject~to}\quad
&
\I(\mathbf{S};\widetilde{\mathbf{X}})\le R^{\mathrm{s}} .
\nonumber
\end{align}
If the generator is deterministic, then $\widetilde{\mathbf{X}}=g(\widetilde{\mathbf{Z}})$ and $P_{\widetilde{\mathbf{X}}\mid \widetilde{\mathbf{Z}}}$ is induced by $g$.

\vspace{-7pt}

\begin{remark}
The latent code $\widetilde{\mathbf{Z}}$ plays different roles across generative models. It may represent the latent variable in a VAE, the $\mathcal{W}$ space in StyleGAN, a latent code obtained through StyleGAN inversion, or the latent/noise representation used in diffusion models.
\end{remark}

\vspace{-7pt}

\subsection{Threat Model}
\label{Ssec:ThreatModel}

Our threat model is based on the following assumptions:
\begin{itemize}[leftmargin=1em]
\item
We consider an adversary interested in inferring a sensitive attribute $\mathbf{S}$ associated with the data $\mathbf{X}$. The attribute $\mathbf{S}$ may be a deterministic or randomized function of $\mathbf{X}$.
We limit $\mathbf{S}$ to a discrete attribute, which accommodates most scenarios of interest, such as a facial feature or an identity attribute.
\item
The adversary observes the released variable $\mathbf{W}$, where $\mathbf{W}=\mathbf{Z}$ in the discriminative setting and $\mathbf{W}=\widetilde{\mathbf{X}}$ in the generative setting. The release mechanism induces the Markov chain $\mathbf{S}\markov \mathbf{X}\markov \mathbf{W}$.
\item
We adopt Kerckhoffs' principle, so the privacy mechanism is public knowledge. In particular, the adversary knows the mechanism selected by the defender, namely $P_{\mathbf{Z}\mid \mathbf{X}}$ in the discriminative setting or the synthetic mechanism in the generative setting.
\end{itemize}

For extended discussion see Appendix~\ref{AppxSec:connection_PF_others}.

%
%
\section{Deep Variational Privacy Funnel}
\label{Sec:DeepVariationalPF_Model}

\subsection{Information Leakage Approximation}

\vspace{-2pt}

We provide parameterized variational approximations for information leakage, including an explicit tight variational bound and an upper bound. 
This approximation is designed to be computationally tractable and easily integrated with deep learning models, which allows for a flexible and efficient evaluation of privacy guarantees.
To better understand the nature of information leakage, we can express $\I \left( \mathbf{S}; \mathbf{Z} \right)$ as:\vspace{-2pt}
\begin{subequations}\label{I_sz_decomposition}
\begin{align}
\I \left( \mathbf{S}; \mathbf{Z} \right) &=  \I \left( \mathbf{X}; \mathbf{Z} \right) - \I \left( \mathbf{X}; \mathbf{Z} \mid \mathbf{S} \right) \\
&= 
\I \left( \mathbf{X}; \mathbf{Z} \right) - \H \left( \mathbf{X} \mid \mathbf{S} \right) + \H \left(  \mathbf{X} \mid \mathbf{S}, \mathbf{Z}\right). 
\end{align}
\end{subequations}
The conditional entropy $\H \left( \mathbf{X} \mid \mathbf{S} \right)$ is originated from the nature of data since it is out of our control. 
It can be interpreted as `\textit{useful information decoding uncertainty}'.
Now, we derive the variational decomposition of $\I \left( \mathbf{X}; \mathbf{Z} \right)$ and $\H \left(  \mathbf{X} \mid \mathbf{S}, \mathbf{Z}\right)$. 
The mutual information $\I \left( \mathbf{X}; \mathbf{Z} \right)$ can be interpreted as `\textit{information complexity}' or `\textit{encoder capacity}' \cite{razeghi2023bottlenecks}. It can be decomposed~as:\vspace{-5pt}
\begin{align}
\label{I_xz_decomposition}
\I \left( \mathbf{X}; \mathbf{Z} \right)  = 
\D_{\mathrm{KL}} \left( P_{\mathbf{Z} \mid \mathbf{X}} \Vert Q_{\mathbf{Z}} \mid P_{\mathbf{X}} \right) - \D_{\mathrm{KL}} \left( P_{\mathbf{Z}} \Vert Q_{\mathbf{Z}}\right),
\end{align}
where $Q_{\mathbf{Z}}: \mathcal{Z} \rightarrow \mathcal{P}\left( \mathcal{Z}\right)$ is variational approximation of the latent space distribution $P_{\mathbf{Z}}$. 
The conditional entropy $\H \left( \mathbf{X} \! \mid \! \mathbf{S}, \mathbf{Z} \right)$ can be decomposed as:
\begin{subequations}
\label{ConditionalEntropy_X_given_SZ}
\begin{align}
\label{ConditionalEntropy_X_given_SZ_a}
&  \! \!\!\! \! \H \left( \mathbf{X} \! \mid \! \mathbf{S}, \mathbf{Z} \right) \\ 
\! \! &   \! \!\!\! =  \!- \mathbb{E}_{P_{\mathbf{S}, \mathbf{X}, \mathbf{Z}}} \left[ \log P_{\mathbf{X} \mid \mathbf{S}, \mathbf{Z}} \right]\\
\! \! &  \! \!\! \! =  \! -\mathbb{E}_{P_{\mathbf{S}, \mathbf{X}}}  \! \left[  \mathbb{E}_{P_{\mathbf{Z} \mid \mathbf{X}}}  \! \left[ \log Q_{\mathbf{X} \mid \mathbf{S}, \mathbf{Z}} \right] \right]  \! - \!\D_{\mathrm{KL}}  \! \left( P_{\mathbf{X}\mid \mathbf{S}, \mathbf{Z} }  \Vert Q_{\mathbf{X} \mid \mathbf{S}, \mathbf{Z}} \right)\\
\! \! &   \! \!\! \! \leq -\mathbb{E}_{P_{\mathbf{S}, \mathbf{X}}} \left[  \mathbb{E}_{P_{\mathbf{Z} \mid \mathbf{X}}} \left[ \log Q_{\mathbf{X} \mid \mathbf{S}, \mathbf{Z}} \right] \right] \\
\! &   \! \!\! \! = \H \left( P_{\mathbf{X} \mid \mathbf{S,Z}} \Vert  Q_{\mathbf{X} \mid \mathbf{S,Z}}  \mid P_{\mathbf{S,Z}} \right)
\eqqcolon \H^{\mathrm{U}} \! \left( \mathbf{X} \! \mid \! \mathbf{S}, \mathbf{Z} \right), \label{ConditionalEntropy_X_given_SZ_b}
\end{align}
\end{subequations}

\noindent
where $Q_{\mathbf{X}\mid \mathbf{S}, \mathbf{Z}}\! :\! \mathcal{S} \! \times \! \mathcal{Z} \! \rightarrow \! \mathcal{P}\left( \mathcal{X}\right)$ is variational approximation of the optimal uncertainty decoder distribution $P_{\mathbf{X}\mid \mathbf{S}, \mathbf{Z}}$, and the inequality in \eqref{ConditionalEntropy_X_given_SZ_b} follows by noticing that $ \D_{\mathrm{KL}}  ( P_{\mathbf{X}\mid \mathbf{S}, \mathbf{Z} }  \Vert Q_{\mathbf{X} \mid \mathbf{S}, \mathbf{Z}}  )$ $\geq 0$. 
Using \eqref{I_sz_decomposition}, \eqref{I_xz_decomposition} and \eqref{ConditionalEntropy_X_given_SZ}, the variational upper bound of information leakage is given as:\vspace{-4pt}
\begin{multline}\label{I_SZ_upperBound}
    \I \left( \mathbf{S}; \mathbf{Z} \right) \leq \D_{\mathrm{KL}} \left( P_{\mathbf{Z} \mid \mathbf{X}} \Vert Q_{\mathbf{Z}} \mid P_{\mathbf{X}} \right) - \D_{\mathrm{KL}} \left( P_{\mathbf{Z}} \Vert Q_{\mathbf{Z}}\right)  \\ + \H^{\mathrm{U}} \! \left( \mathbf{X} \! \mid \! \mathbf{S}, \mathbf{Z} \right). 
\end{multline}

\vspace{-3pt}

Having the variational upper bound of information leakage, we now approximate the parameterized variational bound using neural networks. 
Let $P_{\boldsymbol{\phi}} (\mathbf{Z} \! \mid \! \mathbf{X})$ represent the family of encoding probability distributions $P_{\mathbf{Z} \mid \mathbf{X}}$ over $\mathcal{Z}$ for each element of space $\mathcal{X}$, parameterized by the output of a deep neural network $f_{\boldsymbol{\phi}}$ with parameters $\boldsymbol{\phi}$. 
Analogously, let $P_{\boldsymbol{\varphi}}  \left( \mathbf{X} \! \mid \! \mathbf{S}, \mathbf{Z} \right)$ denote the corresponding family of decoding probability distributions $Q_{\mathbf{X} \mid \mathbf{S}, \mathbf{Z}}$, driven by $g_{\boldsymbol{\varphi}}$. 
Lastly, $Q_{\boldsymbol{\psi}} (\mathbf{Z})$ denotes the parameterized prior distribution, either explicit or implicit, that is associated with $Q_{\mathbf{Z}}$. 

Using \eqref{I_xz_decomposition}, the parameterized variational approximation of $\I \left( \mathbf{X}; \mathbf{Z} \right)$ can be defined as:\vspace{-5pt}
\begin{multline}\label{Eq:I_XZ_phi_psi}
\I_{\boldsymbol{\phi}, \boldsymbol{\psi}}   \left( \mathbf{X}; \mathbf{Z} \right) \coloneqq  \D_{\mathrm{KL}} \! \left( P_{\boldsymbol{\phi}} (\mathbf{Z} \! \mid \! \mathbf{X}) \, \Vert \, Q_{\boldsymbol{\psi}}   (\mathbf{Z}) \mid P_{\mathsf{D}}(\mathbf{X}) \right)\\ -  \D_{\mathrm{KL}} \! \left( P_{\boldsymbol{\phi}}(\mathbf{Z}) \, \Vert \, Q_{\boldsymbol{\psi}} (\mathbf{Z})\right).
\end{multline}
The parameterized variational approximation of conditional entropy $\H^{\mathrm{U}} \left( \mathbf{X} \mid \mathbf{S}, \mathbf{Z} \right)$ in \eqref{ConditionalEntropy_X_given_SZ_b} can be defined as:
\begin{eqnarray}
 \H_{\boldsymbol{\phi}, \boldsymbol{\varphi}}^{\mathrm{U}} \left( \mathbf{X} \! \mid \! \mathbf{S}, \mathbf{Z} \right) \coloneqq - \mathbb{E}_{P_{\mathbf{S}, \mathbf{X}}} \left[  \mathbb{E}_{P_{\boldsymbol{\phi}} (\mathbf{Z} \mid \mathbf{X})} \left[ \log P_{\boldsymbol{\varphi}}  (\mathbf{X} \! \mid \! \mathbf{S}, \mathbf{Z}) \right] \right] .
\end{eqnarray}
Let $\I_{\boldsymbol{\phi}, \boldsymbol{\xi}} \left( \mathbf{S}; \mathbf{Z} \right)$ denote the parameterized variational approximation of information leakage $\I \left(\mathbf{S}; \mathbf{Z} \right)$. 
Using \eqref{I_SZ_upperBound}, an upper bound of $\I_{\boldsymbol{\phi}, \boldsymbol{\xi}}  \! \left( \mathbf{S}; \mathbf{Z} \right)$ can be given as:\vspace{-4pt}
\begin{subequations}\label{Eq:I_SZ_phi_Xi_UpperBound}
\begin{align}
\!\!\!\!\! \I_{\boldsymbol{\phi}, \boldsymbol{\xi}} (\mathbf{S}; \mathbf{Z})  &\leq  \!\!\!\!\!\!\!\!\!\!
\underbrace{\I_{\boldsymbol{\phi}, \boldsymbol{\psi}}  \left( \mathbf{X}; \mathbf{Z} \right)}_{\mathrm{Information~Complexity}}  \!\! + \!  \underbrace{\H_{\boldsymbol{\phi}, \boldsymbol{\varphi}}^{\mathrm{U}}   \left( \mathbf{X} \! \mid \! \mathbf{S}, \mathbf{Z} \right)}_{\mathrm{Information~Uncertainty}} \!\!\!\!  + \, \mathrm{c} \\ 
&\eqqcolon \; \I_{\boldsymbol{\phi}, \boldsymbol{\psi}, \boldsymbol{\varphi}}^{\mathrm{U}}   \left( \mathbf{S}; \mathbf{Z} \right) + \mathrm{c},
\end{align}
\end{subequations}
where $\mathrm{c}$ is a constant term, independent of the neural network parameters.

This upper bound encourages the model to reduce both the information complexity, represented by $\I_{\boldsymbol{\phi}, \boldsymbol{\psi}}  \left( \mathbf{X}; \mathbf{Z} \right)$, and the information uncertainty, denoted by $\H_{\boldsymbol{\phi}, \boldsymbol{\varphi}}^{\mathrm{U}}  \left( \mathbf{X} \! \mid \! \mathbf{S}, \mathbf{Z} \right)$. Consequently, this leads the model to `forget' or de-emphasize the sensitive attribute $\mathbf{S}$, which subsequently reduces the uncertainty about the useful data $\mathbf{X}$. In essence, this nudges the model towards an accurate reconstruction of the data~$\mathbf{X}$.

Now, let us derive another parameterized variational bound of information leakage $\I_{\boldsymbol{\phi}, \boldsymbol{\xi}} \left( \mathbf{S}; \mathbf{Z} \right)$. 
We can decompose $\I_{\boldsymbol{\phi}, \boldsymbol{\xi}}  \left( \mathbf{S}; \mathbf{Z} \right)$ as follows:\vspace{-4pt}
\begin{subequations}\label{Eq:I_SZ_phi_xi_SecondDecomposition}
\begin{align}
&\I_{\boldsymbol{\phi}, \boldsymbol{\xi}}  \left( \mathbf{S}; \mathbf{Z} \right)\nonumber\\
& =
\mathbb{E}_{P_{\mathbf{S}, \mathbf{X}}} \! \left[  \mathbb{E}_{P_{\boldsymbol{\phi}} \left( \mathbf{Z} \mid \mathbf{X} \right) } \! \left[ \log P_{\boldsymbol{\xi}}  \left( \mathbf{S} \! \mid \! \mathbf{Z} \right) \right] \right]
+ \mathbb{E}_{P_{\mathbf{S}}} \left[ \log P_{\boldsymbol{\xi}} (\mathbf{S}) \right] \!\!
\\ 
& \qquad \qquad\qquad\qquad\qquad\qquad\qquad\;\;\, - \mathbb{E}_{P_{\mathbf{S}}} \big[ \log \frac{P_{\mathbf{S}}}{P_{\boldsymbol{\xi}} (\mathbf{S})}\big]  \nonumber
\\
\label{Eq:I_SZ_phi_xi_SecondDecomposition_b}
& =
 \underbrace{- \H_{\boldsymbol{\phi}, \boldsymbol{\xi}} \left( \mathbf{S} \! \mid  \! \mathbf{Z} \right) 
+ \H \left( P_{\mathbf{S}} \, \Vert \, P_{\boldsymbol{\xi}} (\mathbf{S}) \right)}_{\mathrm{Prediction~Fidelity}}\,\,\, 
- \!\!\!\! \underbrace{ \D_{\mathrm{KL}} \left( P_{\mathbf{S}} \, \Vert \, P_{\boldsymbol{\xi}} (\mathbf{S}) \right)}_{\mathrm{Distribution~Discrepancy}}
\end{align}
\end{subequations}
where $P_{\boldsymbol{\xi}} (\mathbf{S} \! \mid \! \mathbf{Z})$ denotes the corresponding family of decoding probability distribution $Q_{\mathbf{S} \mid \mathbf{Z}}$, where $Q_{\mathbf{S} \mid \mathbf{Z}} : \mathcal{Z} \rightarrow \mathcal{P} (\mathcal{S})$ is a variational approximation of optimal decoder distribution $P_{\mathbf{S} \mid \mathbf{Z}}$.

\noindent
Let us interpret the MI decomposition in Eq~\eqref{Eq:I_SZ_phi_xi_SecondDecomposition_b}:
\begin{itemize}[leftmargin=1em]
    \item 
    Negative conditional cross-entropy $- \H_{\boldsymbol{\phi}, \boldsymbol{\xi}} \left( \mathbf{S} \! \mid  \! \mathbf{Z} \right) $: 
    This term aims to maximize the uncertainty in predicting $\mathbf{S}$ given $\mathbf{Z}$. 
    $\H_{\boldsymbol{\phi}, \boldsymbol{\xi}} \left( \mathbf{S} \! \mid  \! \mathbf{Z} \right)$ can be as low as $0$ when $\mathbf{S}$ is deterministically predictable given $\mathbf{Z}$. This means that knowing $\mathbf{Z}$ gives us full information about $\mathbf{S}$. A negative sign encourages the  model (encoder) to increase the entropy of $\mathbf{S}$ given $\mathbf{Z}$, which means making $\mathbf{S}$ less predictable when you know $\mathbf{Z}$.  In the case of a discrete sensitive attribute $\mathbf{S}$, the conditional entropy is maximized when all the conditional distributions $P_{\mathbf{S \mid \mathbf{Z}= \mathbf{z}}}$ are uniform.
    The maximum entropy is $\log_2 \vert \mathcal{S} \vert$, where $\vert \mathcal{S} \vert$ is the number of possible states (or values, or classes) for $\mathbf{S}$. 
    This means the adversary, lacking any additional information, can do no better than `\textit{random guessing}'. This scenario equates to a potential lower boundary for $- \H_{\boldsymbol{\phi}, \boldsymbol{\xi}} \left( \mathbf{S} \! \mid  \! \mathbf{Z} \right) $ at $- \log_2 \vert \mathcal{S} \vert$.
    \item 
    Cross-entropy $\H \left( P_{\mathbf{S}} \, \Vert \, P_{\boldsymbol{\xi}} (\mathbf{S}) \right)$: This term encourages the classifier to produce correct predictions for $\mathbf{S}$. The minimum value is equal to the entropy of $P_{\mathbf{S}}$, i.e., $\H(P_{\mathbf{S}})$, which is achieved when $P_{\boldsymbol{\xi}} (\mathbf{S}) =  P_{\mathbf{S}}$. Given that $\mathbf{S}$ is discrete, the maximum value is $\log_2 \vert \mathcal{S} \vert$.
    \item 
    Distribution discrepancy
    $\D_{\mathrm{KL}} \left( P_{\mathbf{S}} \,\Vert\, P_{\boldsymbol{\xi}} (\mathbf{S}) \right)$:
    This term ensures the model's inferred distribution, $ P_{\boldsymbol{\xi}} (\mathbf{S})$, aligns tightly with the actual distribution $P_{\mathbf{S}}$. Ideally, the divergence measure, $\D_{\mathrm{KL}} \left( P_{\mathbf{S}}  \Vert  P_{\boldsymbol{\xi}} (\mathbf{S}) \right)$, is minimized to zero when $ P_{\boldsymbol{\xi}} (\mathbf{S})$ aligns perfectly with $P_{\mathbf{S}}$.
\end{itemize}

\vspace{-2pt}
\noindent
By pushing both $\H_{\boldsymbol{\phi}, \boldsymbol{\xi}} \left( \mathbf{S} \! \mid  \! \mathbf{Z} \right) $ and $\H \left( P_{\mathbf{S}} \, \Vert \, P_{\boldsymbol{\xi}} (\mathbf{S}) \right)$ to their maximum values of $\log_2 \vert \mathcal{S} \vert$, and simultaneously minimizing distributional gap $\D_{\mathrm{KL}} \left( P_{\mathbf{S}} \, \Vert \, P_{\boldsymbol{\xi}} (\mathbf{S}) \right)$, the $\I_{\boldsymbol{\phi}, \boldsymbol{\xi}}  \left( \mathbf{S}; \mathbf{Z} \right) $ will approach zero, indicating that $\mathbf{Z}$ has minimal information about~$\mathbf{S}$.

\subsection{Information Utility Approximation}

In this subsection, we turn our focus on quantifying the utility of information. As with information leakage, we provide a careful decomposition of the $\I (\mathbf{X}; \mathbf{Z})$ and derive a parameterized variational approximation for information utility. These measures form the foundation of the Deep Variational PF framework and pave the way for practical and scalable privacy preservation in deep learning applications.
The end-to-end parameterized variational approximation associated to the information utility $\I (\mathbf{X}; \mathbf{Z})$ can be defined as:\vspace{-4pt}
\begin{subequations}\label{Eq:I_XZ_phi_theta}
\begin{align}
\!\!\! \I_{\boldsymbol{\phi}, \boldsymbol{\theta}}  \left( \mathbf{X}; \mathbf{Z} \right) 
& \!  \coloneqq  \!
\mathbb{E}_{P_{\mathsf{D}} (\mathbf{X})} \Big[ \mathbb{E}_{P_{\boldsymbol{\phi}} \left( \mathbf{Z} \mid \mathbf{X} \right) } \Big[ \log \frac{P_{\boldsymbol{\theta}}  \left( \mathbf{X} \! \mid \! \mathbf{Z} \right) }{P_{\mathsf{D}} (\mathbf{X})}  \Big] \Big]   \\
& =  \mathbb{E}_{P_{\mathsf{D}} (\mathbf{X})} \left[  \mathbb{E}_{P_{\boldsymbol{\phi}} \left( \mathbf{Z} \mid \mathbf{X} \right) } \left[ \log P_{\boldsymbol{\theta}}   \left( \mathbf{X} \! \mid \! \mathbf{Z} \right) \right] \right] \\
& - \D_{\mathrm{KL}} \left( P_{\mathsf{D}} (\mathbf{X}) \Vert P_{\boldsymbol{\theta}} (\mathbf{X}) \right)
+ \H \left( P_{\mathsf{D}} (\mathbf{X})  \Vert P_{\boldsymbol{\theta}} (\mathbf{X}) \right) 
  \nonumber \\
& \geq \!\!\!\!\!\!\!\!
  \underbrace{- \H_{\boldsymbol{\phi}, \boldsymbol{\theta}} \! \left( \mathbf{X} \! \mid  \! \mathbf{Z} \right)}_{\mathrm{Reconstruction~Fidelity}}
 \!\!\!\!\!  -  \,   \underbrace{\D_{\mathrm{KL}} \! \left( P_{\mathsf{D}} (\mathbf{X}) \Vert P_{\boldsymbol{\theta}} (\mathbf{X}) \right)}_{\mathrm{Distribution~Discrepancy}} \!\!\\
 & \eqqcolon \,\,\, \I_{\boldsymbol{\phi}, \boldsymbol{\theta}}^{\mathrm{L}}  \left( \mathbf{X}; \mathbf{Z} \right),
\end{align}
\end{subequations}
where $\H_{\boldsymbol{\phi}, \boldsymbol{\theta}} \left( \mathbf{X} \! \mid  \! \mathbf{Z} \right) \coloneqq \mathbb{E}_{P_{\mathsf{D}} (\mathbf{X})} \left[  \mathbb{E}_{P_{\boldsymbol{\phi}} \left( \mathbf{Z} \mid \mathbf{X} \right) } \left[ \log P_{\boldsymbol{\theta}} \left( \mathbf{X} \! \mid \! \mathbf{Z} \right) \right] \right]$.

\subsection{Deep Variational Privacy Funnel Objectives}

Considering \eqref{Eq:PF_LagrangianFunctional} and using the addressed parameterized approximations, one can obtain the $\mathsf{DisPF}$ and $\mathsf{GenPF}$ Lagrangian functionals. We recast the following maximization objectives:\vspace{-3pt}
\begin{align}\label{Eq:DVPF_P1_DisPF}
& (\textsf{P1})\!:\;   \mathcal{L}_{\mathsf{DisPF\text{-}MI}} \left( \boldsymbol{\phi}, \boldsymbol{\theta} , \boldsymbol{\xi}, \alpha \right)  \coloneqq \\
& \overbrace{ 
- \H_{\boldsymbol{\phi}, \boldsymbol{\theta}} \left( \mathbf{X} \! \mid  \! \mathbf{Z} \right) - \D_{\mathrm{KL}} \left( P_{\mathsf{D}} (\mathbf{X}) \, \Vert \, P_{\boldsymbol{\theta}} (\mathbf{X}) \right)
}^{\textcolor{violet}{\mathrm{Information~Utility:}~  \I_{\boldsymbol{\phi}, \boldsymbol{\theta}}^{\mathrm{L}} \left( \mathbf{X}; \mathbf{Z} \right) }} \nonumber 
\\
& - \alpha \underbrace{ \Big( \! - \H_{\boldsymbol{\phi}, \boldsymbol{\xi}} \left( \mathbf{S} \! \mid  \! \mathbf{Z} \right) 
+ \H \left( P_{\mathbf{S}} \, \Vert \, P_{\boldsymbol{\xi}} (\mathbf{S}) \right) - \D_{\mathrm{KL}} \! \left( P_{\mathbf{S}} \Vert P_{\boldsymbol{\xi}} (\mathbf{S}) \right) \! \Big) }_{\textcolor{red}{\mathrm{Information~Leakage:}~\I_{\boldsymbol{\phi}, \boldsymbol{\xi}}  \left( \mathbf{S} ; \mathbf{Z} \right) }} . \nonumber 
\end{align}

\vspace{-7pt}

\begin{multline}\label{Eq:DVPF_P2_GenPF}
\!\!\! (\textsf{P2})\! :  \; 
\mathcal{L}_{\mathsf{GenPF\text{-}MI}} \left( \boldsymbol{\phi}, \boldsymbol{\theta}, \boldsymbol{\psi} , \boldsymbol{\varphi}, \alpha \right)  \coloneqq \qquad \\
 \overbrace{ - \H_{\boldsymbol{\phi}, \boldsymbol{\theta}} \left( \mathbf{X} \! \mid  \! \mathbf{Z} \right) - \D_{\mathrm{KL}} \left( P_{\mathsf{D}} (\mathbf{X}) \, \Vert \, P_{\boldsymbol{\theta}} (\mathbf{X}) \right)}^{\textcolor{violet}{\mathrm{Information~Utility:}~\I_{\boldsymbol{\phi}, \boldsymbol{\theta}}^{\mathrm{L}}  \left( \mathbf{X}; \mathbf{Z} \right) }}  
\\  
 - \alpha  \underbrace{ \Big(  \I_{\boldsymbol{\phi}, \boldsymbol{\psi}} \left( \mathbf{X}; \mathbf{Z} \right) + \H_{\boldsymbol{\phi}, \boldsymbol{\varphi}}^{\mathrm{U}}   \left( \mathbf{X}  \mid \mathbf{S}, \mathbf{Z} \right) \Big) }_{\textcolor{red}{\mathrm{Information~Leakage}:~\I_{\boldsymbol{\phi}, \boldsymbol{\psi}, \boldsymbol{\varphi}}^{\mathrm{U}}   \left( \mathbf{S}; \mathbf{Z} \right)}}. 
\end{multline}

%
%
%
\subsection{Learning Framework}\label{Ssec:LearningModel}

\vspace{-4pt}

\begin{figure*}[htp]
    \centering
    ~~~\begin{subfigure}[h]{0.49\textwidth}
    \includegraphics[scale=0.37]{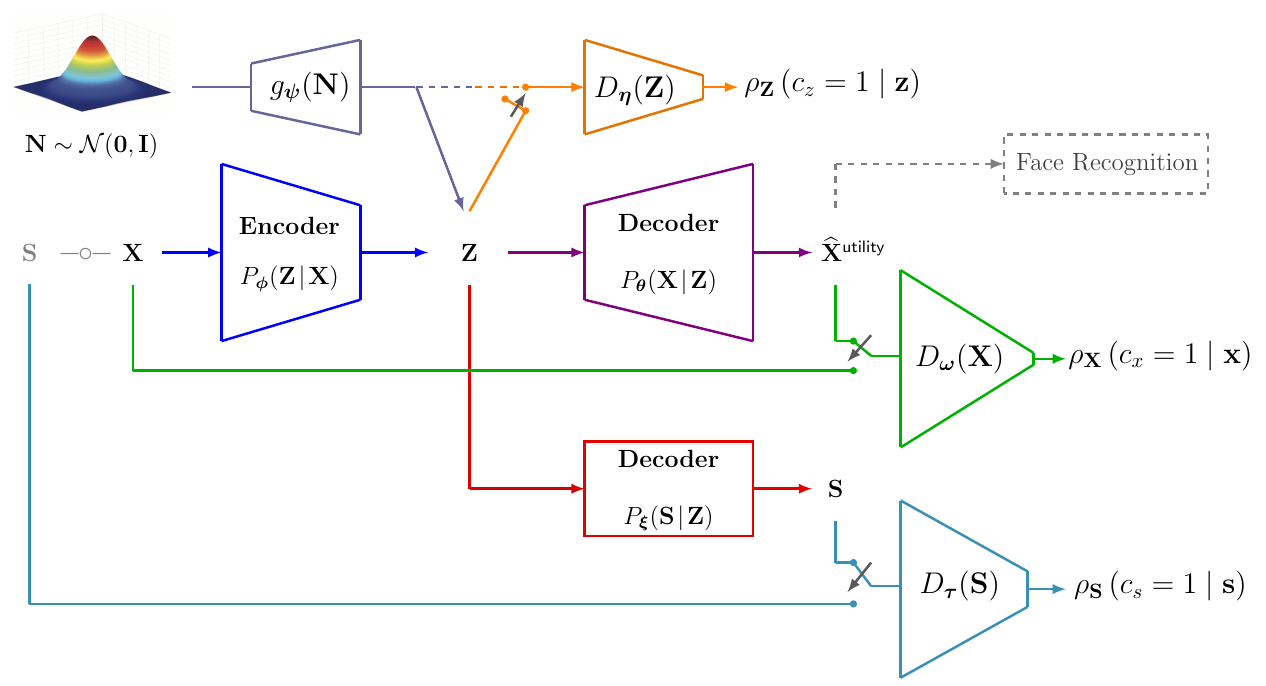}%
    \vspace{-5pt}
    \caption{}
    \label{Fig:DVPF_P1_FaceRecognition}
    \end{subfigure}%
%
%
%
    \begin{subfigure}[h]{0.49\textwidth}
    \includegraphics[scale=0.37]{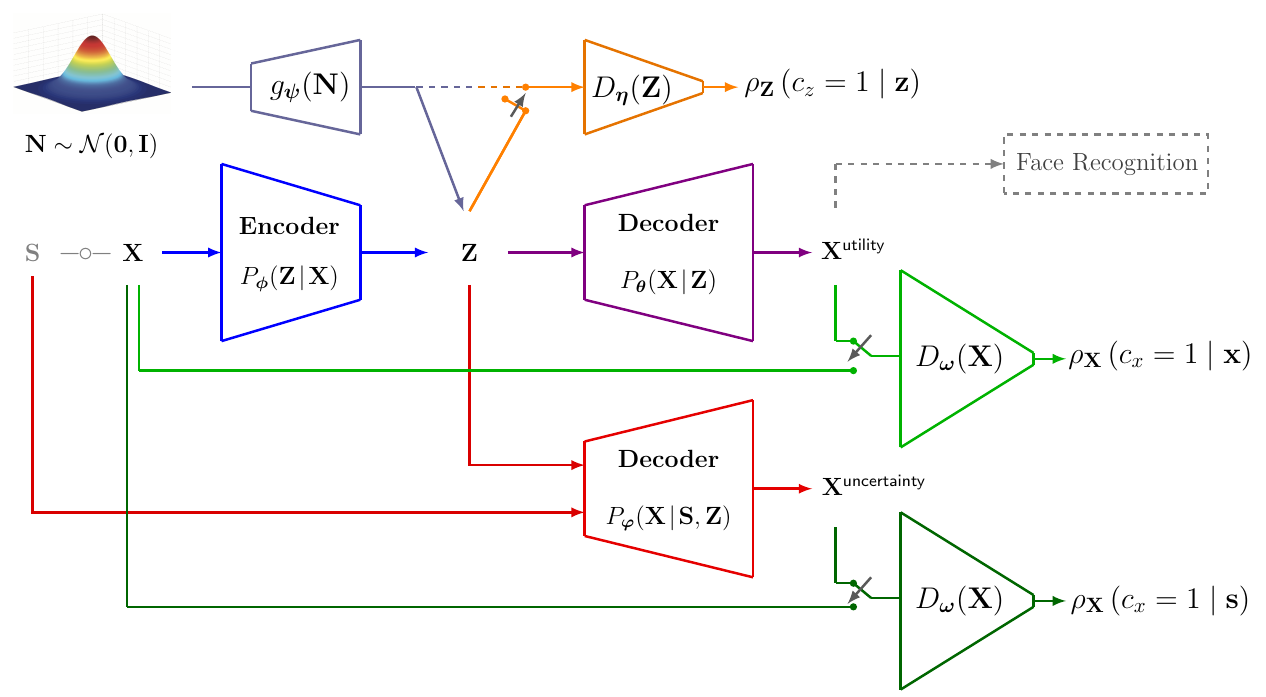}%
    \vspace{-5pt}
    \caption{}
    \label{Fig:GVPF_P2_FaceRecognition}
    \end{subfigure} 
    \vspace{-5pt} 
    \caption{The training architectures associated with: (a) deep variational $\mathsf{DisPF\text{-}MI}$\, (\textsf{P1}); (b) deep variational $\mathsf{GenPF\text{-}MI}$\, (\textsf{P2}).}
    \label{DVPF_P1_GVPF_P2_FaceRecognition}
\vspace{-5pt} 
\end{figure*}

\textbf{System~Designer:}
Consider a set of independent and identically distributed (i.i.d.) training samples
$\{(\mathbf{s}_n,\mathbf{x}_n)\}_{n=1}^{N} \subseteq \mathcal{S}\times\mathcal{X}$,
drawn from the joint distribution $P_{\mathbf{S},\mathbf{X}}$.
We optimize the deep neural networks (DNNs)
$f_{\boldsymbol{\phi}}$, $g_{\boldsymbol{\theta}}$, $g_{\boldsymbol{\xi}}$ (or $g_{\boldsymbol{\varphi}}$),
$D_{\boldsymbol{\eta}}$, $D_{\boldsymbol{\tau}}$, and $D_{\boldsymbol{\omega}}$
using stochastic-gradient-based updates.
The goal is to optimize a Monte Carlo estimate of the DVPF objective
with respect to the parameters
$\boldsymbol{\phi}$, $\boldsymbol{\theta}$, $\boldsymbol{\xi}$ (or $\boldsymbol{\varphi}$),
$\boldsymbol{\eta}$, $\boldsymbol{\tau}$, and $\boldsymbol{\omega}$,
as illustrated in Fig.~\ref{DVPF_P1_GVPF_P2_FaceRecognition}.
Since the objective depends on samples drawn from the stochastic encoder
$P_{\boldsymbol{\phi}}(\mathbf{Z}\mid \mathbf{X})$,
naive backpropagation through the sampled latent variable is not directly available.
To enable gradient-based optimization, we employ the reparameterization trick~\cite{kingma2014auto}.

We parameterize the encoder conditional distribution as a multivariate Gaussian with diagonal covariance.
Assuming $\mathcal{Z}=\mathbb{R}^{d}$, we write
$P_{\boldsymbol{\phi}}(\mathbf{Z}\mid \mathbf{x}) =\mathcal{N}\!\big(
\boldsymbol{\mu}_{\boldsymbol{\phi}}(\mathbf{x}), \mathsf{diag}(\boldsymbol{\sigma}_{\boldsymbol{\phi}}^{2}(\mathbf{x})) \big)$,
where $\boldsymbol{\mu}_{\boldsymbol{\phi}}(\mathbf{x})\in\mathbb{R}^{d}$ and $\boldsymbol{\sigma}_{\boldsymbol{\phi}}(\mathbf{x})\in\mathbb{R}_{+}^{d}$. Let $\boldsymbol{\varepsilon}\sim\mathcal{N}(\boldsymbol{0},\mathbf{I}_{d})$.
Then, for a given sample $\mathbf{x}\in\mathcal{X}$, a latent sample $\mathbf{z}$ can be expressed as $\mathbf{z} =\boldsymbol{\mu}_{\boldsymbol{\phi}}(\mathbf{x}) + \boldsymbol{\sigma}_{\boldsymbol{\phi}}(\mathbf{x})\odot \boldsymbol{\varepsilon}$,
where $\odot$ denotes the Hadamard (element-wise) product.

The prior distribution in the latent space is taken to be the standard isotropic Gaussian $Q_{\mathbf{Z}}=\mathcal{N}(\boldsymbol{0},\mathbf{I}_{d})$.
$\mathbb{E}_{P_{\boldsymbol{\phi}}(\mathbf{X},\mathbf{Z})}
\!\left[
\log \frac{P_{\boldsymbol{\phi}}(\mathbf{Z}\mid \mathbf{X})}{Q_{\mathbf{Z}}(\mathbf{Z})}
\right] = \mathbb{E}_{P_{\mathsf{D}}(\mathbf{X})} \!\left[ \D_{\mathrm{KL}} \!\left(
P_{\boldsymbol{\phi}}(\mathbf{Z}\mid \mathbf{X}) \,\Vert\, Q_{\mathbf{Z}} \right) \right]$.
Moreover, for each $\mathbf{x}\in\mathcal{X}$,
$\D_{\mathrm{KL}} \!\left( P_{\boldsymbol{\phi}}(\mathbf{Z}\mid \mathbf{X}=\mathbf{x}) \,\Vert\, Q_{\mathbf{Z}} \right) = \frac{1}{2} \sum_{i=1}^{d} \left( \mu_{\boldsymbol{\phi},i}(\mathbf{x})^{2}
+ \sigma_{\boldsymbol{\phi},i}(\mathbf{x})^{2} - 1 - \log \sigma_{\boldsymbol{\phi},i}(\mathbf{x})^{2} \right)$.

\vspace{1pt}

For the KL-divergence terms in \eqref{Eq:I_XZ_phi_psi}, \eqref{Eq:I_SZ_phi_xi_SecondDecomposition}, and \eqref{Eq:I_XZ_phi_theta}
that do not admit a tractable closed form, we employ the density-ratio trick \cite{nguyen2010estimating,sugiyama2012density}. This approach rewrites the density-ratio estimation problem as a binary classification task by introducing a label
$C\in\{0,1\}$ that indicates from which of the two distributions a sample was drawn.
A discriminator trained on this task provides an estimate of the log-density ratio, and hence of the corresponding KL divergence, without requiring explicit parametric models for the two densities.
By contrast, the KL term with respect to the Gaussian prior $Q_{\mathbf{Z}}$ above is computed analytically and does not require density-ratio estimation.

%

\textbf{Learning Procedure:}
The DVPF models $(\textsf{P1})$ \eqref{Eq:DVPF_P1_DisPF} and $(\textsf{P2})$ \eqref{Eq:DVPF_P2_GenPF} are trained via a six-step alternating block coordinate descent process. 
In this process, steps 1, 5, and 6 are specific for each model, while steps 2, 3, and 4 are identical for both $(\textsf{P1})$ and $(\textsf{P2})$.
The complete training algorithm of the deep variational $\mathsf{GenPF\text{-}MI}$ model is shown in the Algorithm~\ref{Algorithm:P2_GenPF} on page 10. The iterative alternating block coordinate descent algorithm associated with \eqref{Eq:DVPF_P1_DisPF} is provided in the supplemental materials. Fig.~\ref{DVPF_P1_GVPF_P2_FaceRecognition} illustrates the training architectures for $(\textsf{P1})$ \eqref{Eq:DVPF_P1_DisPF} and $(\textsf{P2})$ \eqref{Eq:DVPF_P2_GenPF}.

\noindent
(1) {\small\textbf{\textsf{Train the Encoder $\boldsymbol{\phi}$, Utility Decoder $\boldsymbol{\theta}$ and Uncertainty Decoder $\boldsymbol{\xi}$ for $(\textsf{P1})$ ($\boldsymbol{\varphi}$ for $(\textsf{P2})$)}}}.\vspace{-2pt}
\begin{multline}
    \!\!\!\! (\textsf{P1}): \mathop{\max}_{\boldsymbol{\phi}, \boldsymbol{\theta}, \boldsymbol{\xi}} \; \mathbb{E}_{P_{\mathsf{D}}(\mathbf{X})} \left[ \mathbb{E}_{P_{\boldsymbol{\phi}} (\mathbf{Z} \mid \mathbf{X})} \left[ \log P_{\boldsymbol{\theta}} (\mathbf{X} \! \mid \! \mathbf{Z} ) \right]  \right] \\ \qquad
    - \alpha \; \mathbb{E}_{P_{\mathbf{S}, \mathbf{X}}} \left[  \mathbb{E}_{P_{\boldsymbol{\phi}} \left( \mathbf{Z} \mid \mathbf{X} \right) } \left[ \log P_{\boldsymbol{\xi}}  \left( \mathbf{S} \! \mid \! \mathbf{Z} \right) \right] \right].
\end{multline}
\vspace{-16pt}
\begin{multline}
    \!\!\!\! (\textsf{P2}): \mathop{\max}_{\boldsymbol{\phi}, \boldsymbol{\theta}, \boldsymbol{\varphi}} \;   \mathbb{E}_{P_{\mathsf{D}}(\mathbf{X})} \left[ \mathbb{E}_{P_{\boldsymbol{\phi}} (\mathbf{Z} \mid \mathbf{X})} \left[ \log P_{\boldsymbol{\theta}} (\mathbf{X} \! \mid \! \mathbf{Z} ) \right]  \right] \\ \qquad\qquad
    -   \alpha  \; \; \D_{\mathrm{KL}} \left( P_{\boldsymbol{\phi}} (\mathbf{Z} \mid \mathbf{X}) \Vert Q_{\boldsymbol{\psi}} (\mathbf{Z}) \! \mid \! P_{\mathsf{D}} (\mathbf{X})\right)  \\ 
    \qquad \qquad
    -  \alpha \;\;  \mathbb{E}_{P_{\mathbf{S}, \mathbf{X}}} \left[ \mathbb{E}_{P_{\boldsymbol{\phi}}(\mathbf{Z} \mid \mathbf{X})} \left[ - \log P_{\boldsymbol{\varphi}} (\mathbf{X} \! \mid \! \mathbf{S}, \mathbf{Z})\right] \right]. \!\!\!\!
\end{multline}

\noindent
(2) {\small\textbf{\textsf{Train the Latent Space Discriminator}}~$ \boldsymbol{\eta}$}.\vspace{-5pt}
\begin{multline}
    \mathop{\min}_{\boldsymbol{\eta}}  \quad   \mathbb{E}_{P_{\mathsf{D}}(\mathbf{X})} \left[ \, \mathbb{E}_{P_{\boldsymbol{\phi}}(\mathbf{Z} \mid \mathbf{X})} \left[ - \log D_{\boldsymbol{\eta}} (\mathbf{Z}) \right] \, \right] \\+ \mathbb{E}_{Q_{\boldsymbol{\psi}} (\mathbf{Z})} \left[ \, - \log (1 - D_{\boldsymbol{\eta}} (\mathbf{Z})) \, \right]. 
\end{multline}

\noindent
(3) {\small\textbf{\textsf{Train the Encoder $\boldsymbol{\phi}$ and Prior Distribution Generator $ \boldsymbol{\psi}$ Adversarially}}}.\vspace{-5pt}
\begin{multline}
    \mathop{\max}_{\boldsymbol{\phi}, \boldsymbol{\psi}}  \quad   \mathbb{E}_{P_{\mathsf{D}}(\mathbf{X})} \left[ \, \mathbb{E}_{P_{\boldsymbol{\phi}}(\mathbf{Z} \mid \mathbf{X})} \left[ - \log D_{\boldsymbol{\eta}} (\mathbf{Z}) \right] \, \right] \\+ \mathbb{E}_{Q_{\boldsymbol{\psi}} (\mathbf{Z})} \left[ \, - \log (1 - D_{\boldsymbol{\eta}} (\mathbf{Z})) \, \right].
\end{multline}

\noindent
(4) {\small\textbf{\textsf{Train the Utility Output Space Discriminator}}~$\boldsymbol{\omega}$}.\vspace{-5pt}
\begin{multline}\label{Eq:TrainVisibleSpaceDiscriminator}
    \mathop{\min}_{\boldsymbol{\omega}}  \quad  \mathbb{E}_{P_{\mathsf{D}}(\mathbf{X})} \left[ \, - \log D_{\boldsymbol{\omega}} (\mathbf{X})  \, \right] \\ + \mathbb{E}_{Q_{\boldsymbol{\psi}} (\mathbf{Z})} \left[ \, - \log \left( 1 - D_{\boldsymbol{\omega}} ( \, g_{\boldsymbol{\theta}}(\mathbf{Z} ) \, ) \right) \, \right].
\end{multline}

\noindent
(5) {\small\textbf{\textsf{Train the Prior Distribution Generator $\boldsymbol{\psi}$, Utility Decoder $\boldsymbol{\theta}$, and Uncertainty Decoder $\boldsymbol{\xi}$ for $(\textsf{P1})$ ($\boldsymbol{\varphi}$ for $(\textsf{P2})$) Adversarially}}}.\vspace{-5pt}
\begin{multline}\label{TrainPriorDistributionGeneratorAndUtilityDecoderAdversarially}
    \!\!\!\! (\textsf{P1}): \mathop{\max}_{\boldsymbol{\psi}, \boldsymbol{\theta}, \boldsymbol{\xi}}  \quad  \mathbb{E}_{Q_{\boldsymbol{\psi}} (\mathbf{Z})} \left[ \, - \log \left( 1 - D_{\boldsymbol{\omega}} ( \, g_{\boldsymbol{\theta}}(\mathbf{Z} ) \, ) \right) \, \right] + \\
    \mathbb{E}_{Q_{\boldsymbol{\psi}} (\mathbf{Z})} \left[ \, - \log \left( 1 - D_{\boldsymbol{\omega}} ( \, g_{\boldsymbol{\xi}}(\mathbf{Z} ) \, ) \right) \, \right].
\end{multline}
\vspace{-18pt}
\begin{multline}\label{TrainPriorDistributionGeneratorAndUtilityDecoderAdversarially}
    \!\!\!\! (\textsf{P2}): \mathop{\max}_{\boldsymbol{\psi}, \boldsymbol{\theta}, \boldsymbol{\varphi}}  \quad  \mathbb{E}_{Q_{\boldsymbol{\psi}} (\mathbf{Z})} \left[ \, - \log \left( 1 - D_{\boldsymbol{\omega}} ( \, g_{\boldsymbol{\theta}}(\mathbf{Z} ) \, ) \right) \, \right] \\ \quad +
    \mathbb{E}_{Q_{\boldsymbol{\psi}} (\mathbf{Z})} \left[ \, - \log \left( 1 - D_{\boldsymbol{\omega}} ( \, g_{\boldsymbol{\varphi}}(\mathbf{S}, \mathbf{Z} ) \, ) \right) \, \right].\!\!\!\!
\end{multline}

\noindent
(6) {\small\textbf{\textsf{Train Uncertainty Output Space Discriminator $\boldsymbol{\tau}$ for $(\textsf{P1})$ ($\boldsymbol{\omega}$ for $(\textsf{P2})$)}}}.\vspace{-4pt}
\begin{multline}
    \!\!\!\! (\textsf{P1}): \mathop{\min}_{\boldsymbol{\tau}}  \,  \mathbb{E}_{P_{\mathbf{S}}} \left[ - \log D_{\boldsymbol{\tau}} (\mathbf{S}) \right] \\ + \mathbb{E}_{Q_{\boldsymbol{\psi}} (\mathbf{Z})} \left[ - \log \left( 1 - D_{\boldsymbol{\tau}} ( g_{\boldsymbol{\xi}}(\mathbf{Z} ) ) \right) \right].
\end{multline}

\vspace{-18pt}

\begin{multline}
    \!\!\!\! (\textsf{P2}): \mathop{\min}_{\boldsymbol{\omega}}  \quad  \mathbb{E}_{P_{\mathsf{D}}(\mathbf{X})} \left[ \, - \log D_{\boldsymbol{\omega}} (\mathbf{X})  \, \right] \\ + \mathbb{E}_{Q_{\boldsymbol{\psi}} (\mathbf{Z})} \left[ \, - \log \left( 1 - D_{\boldsymbol{\omega}} ( \, g_{\boldsymbol{\varphi}}(\mathbf{S}, \mathbf{Z} ) \, ) \right) \, \right].
\end{multline}

\vspace{-10pt}

%
%
%
\begin{center}
\centering
\begin{spacing}{1}
\begin{algorithm}
\small
    \setstretch{0.8}
    \begin{algorithmic}[1]
        \State \textbf{Input:} Training Dataset: $\{ \left( \mathbf{s}_n, \mathbf{x}_n  \right) \}_{n=1}^{N}$; Hyper-Parameter: $\alpha$
        \State $\boldsymbol{\phi}, \boldsymbol{\theta}, \boldsymbol{\psi} , \boldsymbol{\varphi}, \boldsymbol{\eta}, \boldsymbol{\omega}\; \gets$ Initialize Network Parameters
        
         \Repeat
        
         \hspace{-15pt}(1) {\small\textbf{\textsf{Train the Encoder $\boldsymbol{\phi}$, Utility Decoder $\boldsymbol{\theta}$, Uncertainty}}}

         {\small\textbf{\textsf{Decoder}} $\boldsymbol{\varphi}$}
         \State  ~~Sample a mini-batch $\{ \mathbf{x}_m, \mathbf{s}_m \}_{m=1}^{M} \sim P_{\mathsf{D}} (\mathbf{X}) P_{\mathbf{S} 
         \mid \mathbf{X}}$
         \State  ~~Compute encoder outputs $\boldsymbol{\mu}_m^{\mathsf{enc}}, \boldsymbol{\sigma}_m^{\mathsf{enc}} \! \! = \! \! f_{\boldsymbol{\phi}} (\mathbf{x}_m), \forall m \! \in \! [M]$
         \State  ~~Apply reparametrization trick:
         \begin{equation}
             \mathbf{z}_m^{\mathsf{enc}} = \boldsymbol{\mu}_m^{\mathsf{enc}} + \boldsymbol{\epsilon}_m  \odot \boldsymbol{\sigma}_m^{\mathsf{enc}}, \; \boldsymbol{\epsilon}_m \sim \mathcal{N}(0, \mathbf{I}) , \; \forall m \in [M]\nonumber
         \end{equation}
         \State  ~~Sample $\{ \mathbf{n}_m \}_{m=1}^{M} \sim \mathcal{N}(\boldsymbol{0}, \mathbf{I})$%
         \State  ~~Compute $\boldsymbol{\mu}_{m}^{\mathsf{prior}}, \boldsymbol{\sigma}_m^{\mathsf{prior}} = g_{\boldsymbol{\psi}} (\mathbf{n}_m), \forall m \in [M]$%
         \State  ~~Compute $\mathbf{z}_m^{\mathsf{prior}} \!\!=\! \boldsymbol{\mu}_m^{\mathsf{prior}} \! + \boldsymbol{\epsilon}_m'  \odot \boldsymbol{\sigma}_m^{\mathsf{prior}}, \boldsymbol{\epsilon}_m' \! \sim \! \mathcal{N}(0, \mathbf{I}), \forall m \! \in \! [M]\!$%
         \State  ~~Compute $\mathbf{\widehat{x}}_m =  g_{\boldsymbol{\theta}} ( \mathbf{z}_m^{\mathsf{enc}} ), \forall m \in [M]$%
         \State  ~~Compute $\mathbf{\widetilde{x}}_m =  g_{\boldsymbol{\varphi}} (\mathbf{z}_m^{\mathsf{enc}}, \mathbf{s}_m), \forall m \in [M]$
         \State  ~~Back-propagate loss:\vspace{-10pt}
                {\footnotesize \begin{multline}
                \quad \mathcal{L} \left( \boldsymbol{\phi},  \boldsymbol{\theta}, \boldsymbol{\varphi} \right) = \! - \frac{1}{M} \sum_{m=1}^{M} \! \Big( \mathsf{dis}( \mathbf{x}_m , \mathbf{\widehat{x}}_m ) \\
                 \qquad\qquad\qquad\qquad - \alpha  \, \D_{\mathrm{KL}} \! \left( P_{\boldsymbol{\phi}} (\mathbf{z}_m^{\mathsf{enc}} \! \mid \! \mathbf{x}_m ) \Vert  Q_{\boldsymbol{\psi}} (\mathbf{z}_m^{\mathsf{prior}}) \right) \\
                 +  \, \alpha \, \mathsf{dis}( \mathbf{x}_m , \mathbf{\widetilde{x}}_m )  \Big) \nonumber
                \end{multline}}
        
        \vspace{-3pt}
        
        \hspace{-15pt}(2) {\small\textbf{\textsf{Train the Latent Space Discriminator}} $ \boldsymbol{\eta} $}
        \State  ~~Sample $\{ \mathbf{x}_m \}_{m=1}^{M} \sim P_{\mathsf{D}} (\mathbf{X})$
        \State  ~~Sample $\{ \mathbf{n}_m \}_{m=1}^{M} \sim \mathcal{N}(\boldsymbol{0}, \mathbf{I})$
        \State  ~~Compute $\mathbf{z}_m^{\mathsf{enc}}\!$ from $\!f_{\! \boldsymbol{\phi}} (\mathbf{x}_m)\!$ with reparametrization, $\forall m$
        \State  ~~Compute $\mathbf{z}_m^{\mathsf{prior}}\!\!$ from $\! g_{\boldsymbol{\psi}} (\mathbf{n}_m \!)\!$ with reparametrization, $\! \forall m $
        \State  ~~Back-propagate loss:\vspace{-5pt}
                {\footnotesize \begin{equation}
                \;\;\; \mathcal{L} \left( \boldsymbol{\eta} \right) =  - \frac{\alpha}{M} \; \sum_{m=1}^{M}  \log D_{\boldsymbol{\eta}} (\mathbf{z}_m^{\mathsf{enc}}) + \log \big( 1- D_{\boldsymbol{\eta}} (\, \mathbf{z}_m^{\mathsf{prior}} \,) \big) \nonumber 
                \end{equation}}
          
        \vspace{-3pt}
        
        \hspace{-15pt}(3) {\small\textbf{\textsf{Train the Encoder $\boldsymbol{\phi}$ and Prior Distribution Generator}}} 

        {\small\textbf{\textsf{$\boldsymbol{\psi} $ Adversarially}}} 
        \State  ~~Sample $\{ \mathbf{x}_m \}_{m=1}^{M} \sim P_{\mathsf{D}} (\mathbf{X})$
        \State  ~~Compute $\mathbf{z}_m^{\mathsf{enc}}\!$ from $\!f_{\! \boldsymbol{\phi}} (\mathbf{x}_m)\!$ with reparametrization, $\forall m $
        \State  ~~Sample $\{ \mathbf{n}_m \}_{m=1}^{M} \sim \mathcal{N}(\boldsymbol{0}, \mathbf{I})$
        \State  ~~Compute $\mathbf{z}_m^{\mathsf{prior}}\!\!$ from $\! g_{ \boldsymbol{\psi}} (\mathbf{n}_m \!)\!$ with reparametrization, $\! \forall m $
        \State  ~~Back-propagate loss:\vspace{-5pt}
                {\footnotesize \begin{equation}
                \;\;\; \mathcal{L} \left( \boldsymbol{\phi}, \boldsymbol{\psi} \right) = \frac{\alpha}{M} \;   \sum_{m=1}^{M}  \log D_{\boldsymbol{\eta}} (\mathbf{z}_m^{\mathsf{enc}})  + \log \big( 1- D_{\boldsymbol{\eta}} (\, \mathbf{z}_m^{\mathsf{prior}} \,) \big) \nonumber 
                \end{equation}}
        
        \vspace{-3pt}  
        
        \hspace{-15pt}(4) {\small\textbf{\textsf{Train the Utility Output Space Discriminator}} $ \boldsymbol{\omega} $}
        \State  ~~Sample $\{ \mathbf{x}_m \}_{m=1}^{M} \sim P_{\mathsf{D}} (\mathbf{X})$
        \State  ~~Sample $\{ \mathbf{n}_m \}_{m=1}^{M} \sim \mathcal{N} \! \left( \boldsymbol{0}, \mathbf{I}\right)$
        \State ~~Compute $\mathbf{z}_m^{\mathsf{prior}}\!\!$ from $\! g_{ \boldsymbol{\psi}} (\mathbf{n}_m \!)\!$ with reparametrization, $\! \forall m $
        \State  ~~Compute $\mathbf{\widehat{x}}_m = g_{\boldsymbol{\theta}} (\mathbf{z}_m^{\mathsf{prior}} ), \forall m \in [M]$
        \State  ~~Back-propagate loss:\vspace{-5pt}
                {\footnotesize \begin{equation}
                \mathcal{L} \left( \boldsymbol{\omega} \right) = -  \frac{1}{M} \sum_{m=1}^{M} \log D_{\boldsymbol{\omega}} (\mathbf{x}_m)  +  \log \left( 1- D_{\boldsymbol{\omega}} (\, \mathbf{\widehat{x}}_m \,) \right) \nonumber
                \end{equation}}
        
        \vspace{-3pt} 
        
        \hspace{-15pt}(5) {\small\textbf{\textsf{Train the Prior Distribution Generator $\boldsymbol{\psi}$, Utility}}}

        {\small\textbf{\textsf{Decoder $\boldsymbol{\theta}$, and Uncertainty Decoder $\boldsymbol{\varphi}$ Adversarially}}}
        \State  ~~Sample $\{ \mathbf{n}_m \}_{m=1}^{M} \sim \mathcal{N} \! \left( \boldsymbol{0}, \mathbf{I}\right)$
        \State ~~Compute $\mathbf{z}_m^{\mathsf{prior}}\!\!$ from $\! g_{ \boldsymbol{\psi}} (\mathbf{n}_m \!)\!$ with reparametrization, $\! \forall m $
        \State  ~~Compute $\mathbf{\widehat{x}}_m \sim g_{\boldsymbol{\theta}} \left( \mathbf{z}_m^{\mathsf{prior}} \right), \forall m \in [M]$
        \State  ~~Compute $\mathbf{\widetilde{x}}_m \sim g_{\boldsymbol{\varphi}} \left( \mathbf{z}_m^{\mathsf{prior}} , \mathbf{s}_m \right), \forall m \in [M]$
        \State  ~~Back-propagate loss:\vspace{-5pt}
                {\footnotesize \begin{equation}
                \;\;\; \;\mathcal{L} \left( \boldsymbol{\psi}, \boldsymbol{\theta}, \boldsymbol{\varphi} \right)\! = \! \frac{1}{M} \!\! \sum_{m=1}^{M} \!\! \log \left( 1 \! - \! D_{\boldsymbol{\omega}} (\, \mathbf{\widehat{x}}_m \,) \right) +  \log \left( 1 \! - \! D_{\boldsymbol{\omega}} (\, \mathbf{\widetilde{x}}_m \,) \right)\!\!\!\! \nonumber
                \end{equation}}
        
        \vspace{-3pt}

        \hspace{-15pt}(6) {\small\textbf{\textsf{Train Uncertainty Output Space Discriminator $\boldsymbol{\omega}$}}}
        \State  ~~Sample a mini-batch $\{ \mathbf{s}_m, \mathbf{x}_m \}_{m=1}^{M} \sim P_{\mathsf{D}} (\mathbf{X}) P_{\mathbf{S} \mid \mathbf{X}}$
        \State  ~~Sample $\{ \mathbf{n}_m \}_{m=1}^{M} \sim \mathcal{N}(\boldsymbol{0}, \mathbf{I})$
        \State  ~~Compute $\mathbf{z}_m^{\mathsf{prior}}\!\!$ from $\! g_{ \boldsymbol{\psi}} (\mathbf{n}_m \!)\!$ with reparametrization, $\! \forall m $
        \State  ~~Compute $\mathbf{\widetilde{x}}_m \sim g_{\boldsymbol{\varphi}} \left( \mathbf{z}_m^{\mathsf{prior}} , \mathbf{s}_m \right), \forall m \in [M]$
        \State  ~~Back-propagate loss:\vspace{-5pt}
                {\footnotesize \begin{equation}
                \mathcal{L} \left( \boldsymbol{\omega} \right) =  \frac{1}{M} \sum_{m=1}^{M} \log D_{\boldsymbol{\omega}} (\, \mathbf{x}_m \,)  +  \log \left( 1- D_{\boldsymbol{\omega}} (\, \mathbf{\widetilde{x}}_m \,) \right) \nonumber
                \end{equation}}
        
        \Until{Convergence}
        \State \textbf{return} $\boldsymbol{\phi}, \boldsymbol{\theta}, \boldsymbol{\psi}, \boldsymbol{\varphi}, \boldsymbol{\eta}, \boldsymbol{\omega}$
    \end{algorithmic}
    \caption{$\textsf{GenPF\text{-}MI}\; (\textsf{P2})$ Training Algorithm}
    \label{Algorithm:P2_GenPF}
\end{algorithm}
\end{spacing}
\end{center}

\vspace{-18pt}

\subsection{Role of Information Complexity in Privacy Leakage}
\label{SubSec:RoleInformationComplexity}

A standard assumption in the PF model is that the sensitive attribute of interest is specified \emph{a priori}. In other words, the defender is assumed to know in advance which feature or variable of the underlying data the adversary seeks to infer. Accordingly, the data-release mechanism can be designed to minimize the information leaked about that specific random variable.
In practice, however, this assumption may be too restrictive. The attribute regarded as sensitive by the defender need not coincide with the attribute that is actually of interest to the adversary. For example, in a given utility-preserving release mechanism, the defender may attempt to suppress inference of gender, whereas an adversary may instead seek to infer identity or facial expression. Motivated by \cite{issa2019operational}, one may therefore consider a more general setting in which the adversary is interested in an attribute that is \emph{not known a priori} to the system designer.
Following \cite{atashin2021variational}, let $\mathbf{S}$ denote an attribute of the data $\mathbf{X}$ whose conditional law $P_{\mathbf{S}\mid \mathbf{X}}$ is unknown to the defender. Since $\mathbf{S}$ is generated from $\mathbf{X}$, the released representation $\mathbf{Z}$ satisfies the Markov chain $\mathbf{S} \markov \mathbf{X} \markov \mathbf{Z}$.
Therefore, by the data-processing inequality, $\I\!\left(\mathbf{S};\mathbf{Z}\right)\le \I\!\left(\mathbf{X};\mathbf{Z}\right)$. 
This shows that the information complexity of the representation, measured by $\I\!\left(\mathbf{X};\mathbf{Z}\right)$, provides a universal upper bound on the leakage about any latent sensitive attribute $\mathbf{S}$ of $\mathbf{X}$.

\vspace{-6pt}

%
\vspace{-2pt}
\section{Face Recognition Experiments}
\label{Sec:Experiments}

\subsubsection{Leading Models and their Core Mechanisms}

Modern face recognition (FR) systems have evolved through a sequence of influential models, including DeepFace \cite{Taigman2014DeepFaceCT}, FaceNet \cite{schroff2015facenet}, OpenFace \cite{amos2016openface}, SphereFace \cite{liu2017sphereface}, CosFace \cite{wang2018cosface}, ArcFace \cite{arcface2019}, and AdaFace \cite{kim2022adaface}. DeepFace combined explicit 3D alignment with a large deep network to improve robustness to pose variation. FaceNet introduced an embedding-based formulation trained with triplet loss, enabling face verification and clustering through distances in the embedding space. SphereFace, CosFace, and ArcFace subsequently shifted the emphasis toward angular- and margin-based objectives on the hypersphere, leading to more discriminative face embeddings. In particular, ArcFace employs an additive angular margin with a clear geometric interpretation, while AdaFace further adapts the margin to image quality in order to improve robustness under quality variation.
In this work, we focus primarily on ArcFace and AdaFace, since they provide strong and well-established margin-based formulations for modern FR systems. This choice also allows us to evaluate privacy-preserving mechanisms on top of competitive and widely used recognition pipelines without introducing unnecessary architectural variability.

%
\begin{figure*}[!t]
    \centering
    \includegraphics[width=0.73\textwidth]{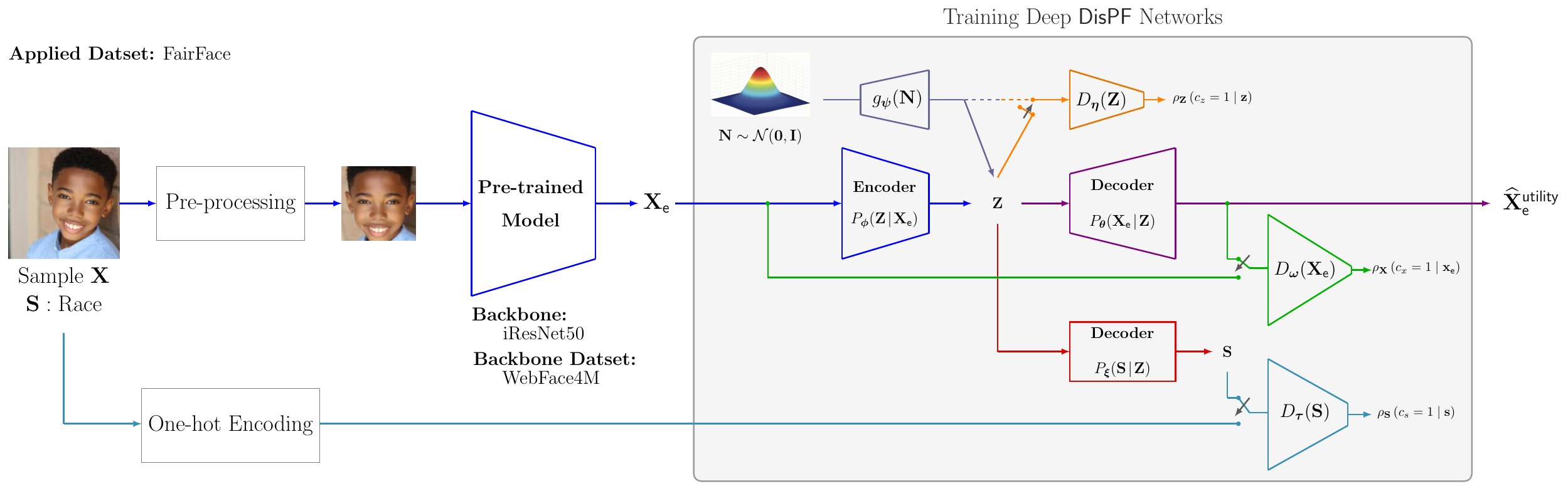}%
    \vspace{-5pt} 
    \caption{Training the deep variational $\mathsf{DisPF}$ model for face recognition experiments.}
    \label{TrainDVPF_FaceRecognition_EmbeddingBased}
\vspace{-5pt} 
\end{figure*} 

%
\begin{figure*}[t!]
    \centering
    \includegraphics[width=0.73\textwidth]{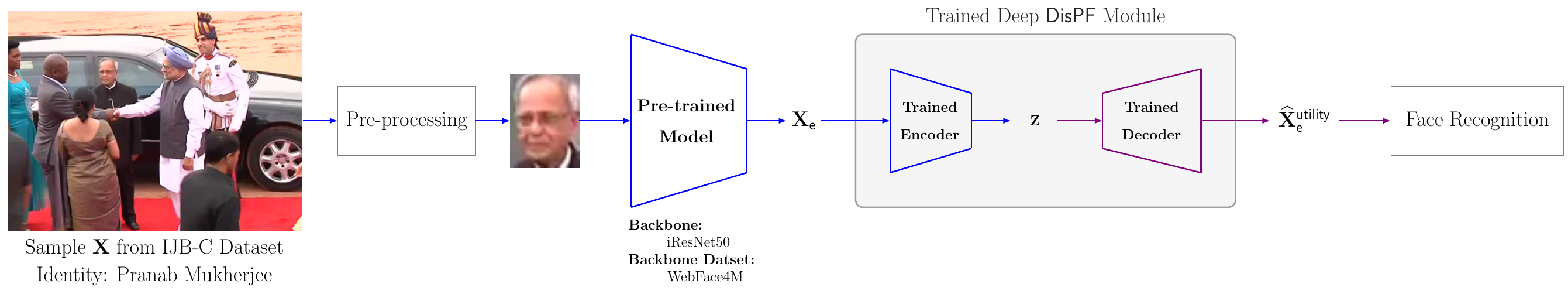}%
    \vspace{-5pt} 
    \caption{Evaluating the performance of the deep variational $\mathsf{DisPF}$ model, which was trained on the FairFace dataset, when applied to the IJB-C test dataset (cross-dataset evaluation). This evaluation highlights the use of $\mathsf{DisPF}$ as a plug-and-play module within the information flow of state-of-the-art face recognition models.}
    \label{InferenceDVPF_FaceRecognition_EmbeddingBased}
\vspace{-5pt} 
\end{figure*} 

\subsubsection{Backbone Architectures for Feature Extraction}

The backbone network plays a central role in FR by mapping raw face images into discriminative feature representations. In our experiments, we use the Improved ResNet (iResNet) architecture \cite{duta2021improved} as the backbone for feature extraction. iResNet is an enhanced residual architecture that modifies several components of the standard ResNet design \cite{resnet2016}, including the information-flow path, the residual building block, and the projection shortcut. These modifications improve optimization and allow deeper networks to be trained more reliably while preserving computational practicality.
The use of iResNet is motivated by its strong empirical performance and its compatibility with margin-based FR losses such as ArcFace and AdaFace. This makes it a suitable and stable backbone for studying the effect of the proposed privacy mechanism on face representations.

\subsubsection{Datasets for Training and Evaluation}

The performance of FR systems depends strongly on the choice of training and evaluation data. Large-scale web-collected datasets such as MS-Celeb-1M \cite{deng2019lightweight_ms1mv3} and WebFace \cite{zhu2021webface260m} have played a central role in training modern FR models, since they provide broad identity coverage and substantial variation in pose, expression, and imaging conditions. In contrast, datasets such as Morph \cite{morph1} and FairFace \cite{karkkainenfairface} are particularly useful when the analysis involves age-related variation and demographic balance, respectively. In particular, FairFace is designed to provide more balanced coverage across race, gender, and age attributes, which is important in studies involving fairness and sensitive-attribute leakage.

For evaluation, unconstrained benchmarks such as Labeled Faces in the Wild (LFW) \cite{huang2008labeled} and IARPA Janus Benchmark-C (IJB-C) \cite{ijbc} remain important testbeds for real-world FR performance. LFW captures substantial variability in pose, illumination, expression, and occlusion under unconstrained conditions, while IJB-C provides a more challenging benchmark for template-based verification and identification. In our experiments, these datasets serve complementary roles: large-scale datasets are used for training, whereas Morph, FairFace, LFW, and IJB-C are used to assess utility preservation, demographic behavior, and generalization under realistic FR conditions.

\vspace{-7pt}

\subsection{Experimental Setup}

\vspace{-2pt}

We consider three iResNet-based FR backbones \cite{resnet2016,arcface2019}, namely iResNet100, iResNet50, and iResNet18. These backbone models were pre-trained on either the MS1MV3 \cite{deng2019lightweight_ms1mv3} or WebFace4M/12M \cite{zhu2021webface260m} datasets. The corresponding FR training losses are ArcFace \cite{arcface2019} and AdaFace \cite{kim2022adaface}.

In the experimental pipeline, we use the above pre-trained FR models as fixed feature extractors. All input images undergo the standard pre-processing steps required by the corresponding pre-trained models, including alignment, resizing, and normalization. On top of these backbones, we train the proposed DVPF frameworks in \eqref{Eq:DVPF_P1_DisPF} and \eqref{Eq:DVPF_P2_GenPF} using the Morph dataset \cite{morph1} and FairFace \cite{karkkainenfairface}. The experiments consider different sensitive-attribute configurations, including demographic groupings based on race and gender.

\autoref{TrainDVPF_FaceRecognition_EmbeddingBased} and \autoref{InferenceDVPF_FaceRecognition_EmbeddingBased} illustrate the framework during the \textit{training} and \textit{inference} phases, respectively, for one representative setup that is described later. During inference, we conduct both same-dataset evaluations, in which the models are tested on unseen portions of the dataset used for training, and cross-dataset evaluations, in which the models are tested on different datasets in order to assess generalization to previously unseen data.

Additional details are provided in Appendix~\ref{AppxSec:TrainingDetails}, Appendix~\ref{AppxSec:DenPF}, and Appendix~\ref{AppxSec:Experiments}.

\begin{table*}[h]
    \caption{Evaluation of facial recognition models using various backbones and loss functions. Metrics include entropy, mutual information between embeddings and labels (gender and race), and recognition accuracy of the sensitive attribute $\mathbf{S}$ on the `Morph' and `FairFace' datasets.}
    \vspace{-16pt}
    \label{Table:PerformanceMetrics_FacialRecognitionModels}
    \centering
    \resizebox{0.92\linewidth}{!}{%
    \begin{tabular}{cccc|cccccccccccc}
    \cline{5-16} 
\multicolumn{4}{c||}{}                                                                                           & \multicolumn{6}{c|}{$\mathbf{S}$: Gender}  & \multicolumn{6}{c|}{$\mathbf{S}$: Race}  \\ \cline{5-16}
\multicolumn{4}{c||}{}   & \multicolumn{2}{c|}{$\H(\mathbf{S})$} &\multicolumn{2}{c|}{$\I(\mathbf{X}; \mathbf{S})$} & \multicolumn{2}{c|}{Acc} & \multicolumn{2}{c|}{$\H(\mathbf{S})$} &\multicolumn{2}{c|}{$\I(\mathbf{X}; \mathbf{S})$} & \multicolumn{2}{c|}{Acc} \\ \hline
\rowcolor{headercolor}
\multicolumn{1}{|c|}{\begin{tabular}[c]{@{}c@{}}Backbone Dataset\end{tabular}} & \multicolumn{1}{c|}{Backbone}  & \multicolumn{1}{c|}{Loss Function} & \multicolumn{1}{c||}{Applied Dataset} & \multicolumn{1}{c|}{Train} & \multicolumn{1}{c|}{Test} & \multicolumn{1}{c|}{Train} & \multicolumn{1}{c|}{Test} & \multicolumn{1}{c|}{Train} & \multicolumn{1}{c|}{Test} & \multicolumn{1}{c|}{Train} & \multicolumn{1}{c|}{Test} & \multicolumn{1}{c|}{Train} & \multicolumn{1}{c|}{Test}  & \multicolumn{1}{c|}{Train} & \multicolumn{1}{c|}{Test}  \\ \hline \cline{1-16} 
\multicolumn{1}{|c|}{WebFace\textbf{4M}} & \multicolumn{1}{c|}{iResNet\textbf{18}} & \multicolumn{1}{c|}{AdaFace} & \multicolumn{1}{c||}{Morph} & \multicolumn{1}{c|}{\multirow{5}{*}{0.619}} &  \multicolumn{1}{c|}{\multirow{5}{*}{0.621}} & \multicolumn{1}{c|}{0.610}   & \multicolumn{1}{c|}{0.620}  & \multicolumn{1}{c|}{0.999}   & \multicolumn{1}{c|}{0.996}  & \multicolumn{1}{c|}{\multirow{5}{*}{0.924}} & \multicolumn{1}{c|}{\multirow{5}{*}{0.933}} & \multicolumn{1}{c|}{0.878}   & \multicolumn{1}{c|}{0.924}    & \multicolumn{1}{c|}{0.998}   & \multicolumn{1}{c|}{0.993} \\ \cline{1-4} \cline{7-10}  \cline{13-16} 
\rowcolor{lightgray}
\multicolumn{1}{|c|}{WebFace\textbf{4M}} & \multicolumn{1}{c|}{iResNet\textbf{50}} & \multicolumn{1}{c|}{AdaFace} & \multicolumn{1}{c||}{Morph} & \multicolumn{1}{c|}{\cellcolor{white}} & \multicolumn{1}{c|}{\cellcolor{white}} & \multicolumn{1}{c|}{0.610}   & \multicolumn{1}{c|}{0.620}    & \multicolumn{1}{c|}{0.999}   & \multicolumn{1}{c|}{0.996} & \multicolumn{1}{c|}{\cellcolor{white}} & \multicolumn{1}{c|}{\cellcolor{white}} & \multicolumn{1}{c|}{0.873}   & \multicolumn{1}{c|}{0.930}    & \multicolumn{1}{c|}{0.998}   & \multicolumn{1}{c|}{0.992}  \\ \cline{1-4} \cline{7-10}  \cline{13-16} 
\multicolumn{1}{|c|}{WebFace\textbf{12M}} & \multicolumn{1}{c|}{iResNet\textbf{101}} & \multicolumn{1}{c|}{AdaFace}  & \multicolumn{1}{c||}{Morph} & \multicolumn{1}{c|}{} & \multicolumn{1}{c|}{} & \multicolumn{1}{c|}{0.605}   & \multicolumn{1}{c|}{0.622}    & \multicolumn{1}{c|}{0.999}   & \multicolumn{1}{c|}{0.996} & \multicolumn{1}{c|}{} & \multicolumn{1}{c|}{} & \multicolumn{1}{c|}{0.873}   & \multicolumn{1}{c|}{0.911}    & \multicolumn{1}{c|}{0.998}   & \multicolumn{1}{c|}{0.992} \\ \cline{1-4} \cline{7-10}  \cline{13-16}
\rowcolor{lightgray}
\multicolumn{1}{|c|}{MS1M-RetinaFace} & \multicolumn{1}{c|}{iResNet\textbf{50}} & \multicolumn{1}{c|}{ArcFace} & \multicolumn{1}{c||}{Morph} & \multicolumn{1}{c|}{\cellcolor{white}} & \multicolumn{1}{c|}{\cellcolor{white}} & \multicolumn{1}{c|}{0.600}   & \multicolumn{1}{c|}{0.620}    & \multicolumn{1}{c|}{0.999}   & \multicolumn{1}{c|}{0.996} & \multicolumn{1}{c|}{\cellcolor{white}} & \multicolumn{1}{c|}{\cellcolor{white}} & \multicolumn{1}{c|}{0.865}   & \multicolumn{1}{c|}{0.910}    & \multicolumn{1}{c|}{0.997}   & \multicolumn{1}{c|}{0.993}  \\ \cline{1-4} \cline{7-10}  \cline{13-16}
\multicolumn{1}{|c|}{MS1M-RetinaFace} & \multicolumn{1}{c|}{iResNet\textbf{100}} & \multicolumn{1}{c|}{ArcFace} & \multicolumn{1}{c||}{Morph} & \multicolumn{1}{c|}{\cellcolor{white}} & \multicolumn{1}{c|}{\cellcolor{white}} & \multicolumn{1}{c|}{0.597}   & \multicolumn{1}{c|}{0.618}    & \multicolumn{1}{c|}{0.999}   & \multicolumn{1}{c|}{0.997} & \multicolumn{1}{c|}{\cellcolor{white}} & \multicolumn{1}{c|}{\cellcolor{white}} & \multicolumn{1}{c|}{0.868}   & \multicolumn{1}{c|}{0.905}    & \multicolumn{1}{c|}{0.997}   & \multicolumn{1}{c|}{0.993}  \\ \cline{1-16} 
\rowcolor{lightgray}
\multicolumn{1}{|c|}{WebFace\textbf{4M}} & \multicolumn{1}{c|}{iResNet\textbf{18}} & \multicolumn{1}{c|}{AdaFace} & \multicolumn{1}{c||}{FairFace} & \multicolumn{1}{c|}{\cellcolor{white}\multirow{5}{*}{0.999}} &  \multicolumn{1}{c|}{\cellcolor{white}\multirow{5}{*}{0.999}} & \multicolumn{1}{c|}{0.930}   & \multicolumn{1}{c|}{0.968}  & \multicolumn{1}{c|}{0.953}   & \multicolumn{1}{c|}{0.923}  & \multicolumn{1}{c|}{\cellcolor{white}\multirow{5}{*}{2.517}} & \multicolumn{1}{c|}{\cellcolor{white}\multirow{5}{*}{2.515}} & \multicolumn{1}{c|}{2.099}   & \multicolumn{1}{c|}{2.405}    & \multicolumn{1}{c|}{0.882}   & \multicolumn{1}{c|}{0.763} \\ \cline{1-4} \cline{7-10}  \cline{13-16} 
\multicolumn{1}{|c|}{WebFace\textbf{4M}} & \multicolumn{1}{c|}{iResNet\textbf{50}} & \multicolumn{1}{c|}{AdaFace} & \multicolumn{1}{c||}{FairFace} & \multicolumn{1}{c|}{} & \multicolumn{1}{c|}{} & \multicolumn{1}{c|}{0.932}   & \multicolumn{1}{c|}{0.968}    & \multicolumn{1}{c|}{0.954}   & \multicolumn{1}{c|}{0.931} & \multicolumn{1}{c|}{} & \multicolumn{1}{c|}{} & \multicolumn{1}{c|}{2.113}   & \multicolumn{1}{c|}{2.409}    & \multicolumn{1}{c|}{0.883}   & \multicolumn{1}{c|}{0.769}  \\ \cline{1-4} \cline{7-10}  \cline{13-16} 
\multicolumn{1}{|c|}{\cellcolor{lightgray}WebFace\textbf{12M}} & \multicolumn{1}{c|}{\cellcolor{lightgray}iResNet\textbf{101}} & \multicolumn{1}{c|}{\cellcolor{lightgray}AdaFace}  & \multicolumn{1}{c||}{\cellcolor{lightgray}FairFace} & \multicolumn{1}{c|}{} & \multicolumn{1}{c|}{} & \multicolumn{1}{c|}{\cellcolor{lightgray}0.934}   & \multicolumn{1}{c|}{\cellcolor{lightgray}0.969}    & \multicolumn{1}{c|}{\cellcolor{lightgray}0.957}   & \multicolumn{1}{c|}{\cellcolor{lightgray}0.930} & \multicolumn{1}{c|}{} & \multicolumn{1}{c|}{} & \multicolumn{1}{c|}{\cellcolor{lightgray}2.151}   & \multicolumn{1}{c|}{\cellcolor{lightgray}2.417}    & \multicolumn{1}{c|}{\cellcolor{lightgray}0.892}   & \multicolumn{1}{c|}{\cellcolor{lightgray}0.765}    \\ \cline{1-4} \cline{7-10}  \cline{13-16} 
\multicolumn{1}{|c|}{MS1M-RetinaFace} & \multicolumn{1}{c|}{iResNet\textbf{50}} & \multicolumn{1}{c|}{ArcFace} & \multicolumn{1}{c||}{FairFace} & \multicolumn{1}{c|}{\cellcolor{white}} & \multicolumn{1}{c|}{\cellcolor{white}} & \multicolumn{1}{c|}{0.892}   & \multicolumn{1}{c|}{0.962}    & \multicolumn{1}{c|}{0.950}   & \multicolumn{1}{c|}{0.927} & \multicolumn{1}{c|}{\cellcolor{white}} & \multicolumn{1}{c|}{\cellcolor{white}} & \multicolumn{1}{c|}{1.952}   & \multicolumn{1}{c|}{2.355}    & \multicolumn{1}{c|}{0.872}   & \multicolumn{1}{c|}{0.753}  \\ \cline{1-4} \cline{7-10}  \cline{13-16}
\rowcolor{lightgray}
\multicolumn{1}{|c|}{MS1M-RetinaFace} & \multicolumn{1}{c|}{iResNet\textbf{100}} & \multicolumn{1}{c|}{ArcFace} & \multicolumn{1}{c||}{FairFace} & \multicolumn{1}{c|}{\cellcolor{white}} & \multicolumn{1}{c|}{\cellcolor{white}} & \multicolumn{1}{c|}{0.889}   & \multicolumn{1}{c|}{0.954}    & \multicolumn{1}{c|}{0.951}   & \multicolumn{1}{c|}{0.927} & \multicolumn{1}{c|}{\cellcolor{white}} & \multicolumn{1}{c|}{\cellcolor{white}} & \multicolumn{1}{c|}{1.949}   & \multicolumn{1}{c|}{2.348}    & \multicolumn{1}{c|}{0.875}   & \multicolumn{1}{c|}{0.765}  \\ \cline{1-16} 
\end{tabular}}
\vspace{-7pt}
\end{table*}

\vspace{-7pt}

\subsection{Experimental Results}

\vspace{-3pt}

\subsubsection{Evaluation of Morph and FairFace Datasets Before Applying DVPF}

\autoref{Table:PerformanceMetrics_FacialRecognitionModels} reports the Shannon entropy, the estimated MI (see Appendix~\ref{AppxSec:MINE}) between the extracted embeddings $\mathbf{X} \in \mathbb{R}^{512}$ and the sensitive attributes $\mathbf{S}$, and the classification accuracy of $\mathbf{S}$, for both the training and test sets, before applying the proposed DVPF model. A close proximity between $\I(\mathbf{X};\mathbf{S})$ and $\H(\mathbf{S})$ indicates that the embeddings substantially reduce the uncertainty about $\mathbf{S}$. Since $\mathbf{S}$ is discrete, we have $\I(\mathbf{X};\mathbf{S}) = \H(\mathbf{S}) - \H(\mathbf{S}\mid \mathbf{X})$, so MI directly quantifies how much information the embeddings reveal about the sensitive attribute. In particular, $\I(\mathbf{X};\mathbf{S}) \leq \H(\mathbf{S})$.
For the Morph and FairFace datasets, the entropy of the sensitive attributes (gender or race) is determined by the corresponding label distribution and therefore remains nearly unchanged across the train/test splits and across different FR embeddings. This reflects the use of the same underlying dataset labels throughout the experiments. For both Morph and FairFace, the gender attribute has two labels (`male' and `female'), so its maximum possible entropy is $\log_2(2)=1$. For race, the maximum possible entropy is $\log_2(4)=2$ for Morph, which has four race labels, and $\log_2(6)=2.585$ for FairFace, which has six race labels.
For Morph, the MI for gender is close to the corresponding entropy, indicating that gender remains highly predictable from the embeddings. For race, the MI values are approximately $0.92\text{-}0.93$, which are also close to the corresponding empirical entropy values. This indicates that the embeddings preserve a substantial amount of race information, while the fact that these entropy values remain well below the theoretical maximum of $\log_2(4)=2$ reflects the imbalance of the race-label distribution in Morph.
In contrast, FairFace exhibits near-maximal empirical entropies for both race ($\sim 2.517$, compared to the maximum possible value $2.585$) and gender ($\sim 0.999$, compared to the maximum possible value $1$), which is consistent with its relatively balanced demographic composition. The corresponding MI and classification results show that these sensitive attributes are also strongly represented in the extracted embeddings prior to applying DVPF.

\begin{table*}[h]
    \caption{Analysis of the obfuscation--utility trade-off in face recognition models based on the iResNet-50 architecture under (\textsf{P1}) and (\textsf{P2}). Performance is reported for different values of the privacy-weight parameter $\alpha$, showing clear differences between $\alpha = 0.1$ and $\alpha = 10$. The sensitive attributes are `Gender' and `Race'. The results are shown for latent dimensionalities $d_{\mathbf{z}} = 512$ (\texttt{top}), $d_{\mathbf{z}} = 256$ (\texttt{middle}), and $d_{\mathbf{z}} = 128$ (\texttt{bottom}). Here, ``WF4M'' denotes ``WebFace4M'', ``MS1M-RF'' denotes ``MS1M-RetinaFace'', and ``TMR'' denotes the true match rate at $\mathrm{FMR}=10^{-1}$ on IJB-C.}
    \vspace{-5pt}
    \label{Table:PerformanceMetrics_FacialRecognitionModels_AfterPrivacyFunnel}
    \centering
    \resizebox{\linewidth}{!}{%
    \begin{tabular}{c|cccccccccccccccccc}
    \cline{2-19} 
    \multicolumn{1}{c||}{\large{(\textsf{P1})}} & \multicolumn{9}{c|}{$\mathbf{S}$: Gender} & \multicolumn{9}{c|}{$\mathbf{S}$: Race} \\ \cline{2-19} 
\multicolumn{1}{c||}{\large{(\textcolor{black}{$d_{\mathbf{z}} = 512$})}}  & 
\multicolumn{3}{c|}{$\alpha = 0.1$}  & \multicolumn{3}{c|}{$\alpha = 1$} & \multicolumn{3}{c|}{$\alpha = 10$}
& \multicolumn{3}{c|}{$\alpha = 0.1$} & \multicolumn{3}{c|}{$\alpha = 1$} & \multicolumn{3}{c|}{$\alpha = 10$}\\ \cline{1-19}
\rowcolor{headercolor}
\multicolumn{1}{|c||}{Face Recognition Model}
& \multicolumn{1}{c|}{TMR} & \multicolumn{1}{c|}{$\I(\mathbf{Z}; \mathbf{S})$} & \multicolumn{1}{c|}{Acc on $\mathbf{S}$} 
& \multicolumn{1}{c|}{TMR} & \multicolumn{1}{c|}{$\I(\mathbf{Z}; \mathbf{S})$} & \multicolumn{1}{c|}{Acc on $\mathbf{S}$} 
& \multicolumn{1}{c|}{TMR} & \multicolumn{1}{c|}{$\I(\mathbf{Z}; \mathbf{S})$} & \multicolumn{1}{c|}{Acc on $\mathbf{S}$} 
& \multicolumn{1}{c|}{TMR} & \multicolumn{1}{c|}{$\I(\mathbf{Z}; \mathbf{S})$} & \multicolumn{1}{c|}{Acc on $\mathbf{S}$}
& \multicolumn{1}{c|}{TMR} & \multicolumn{1}{c|}{$\I(\mathbf{Z}; \mathbf{S})$} & \multicolumn{1}{c|}{Acc on $\mathbf{S}$} 
& \multicolumn{1}{c|}{TMR} & \multicolumn{1}{c|}{$\I(\mathbf{Z}; \mathbf{S})$} & \multicolumn{1}{c|}{Acc on $\mathbf{S}$}\\ \hline \cline{1-19}
%
%
\multicolumn{1}{|c||}{WF4M-i50-Ada-Morph} 
& \multicolumn{1}{c|}{87.31} & \multicolumn{1}{c|}{0.486} & \multicolumn{1}{c|}{0.985} 
& \multicolumn{1}{c|}{67.55} & \multicolumn{1}{c|}{0.484} & \multicolumn{1}{c|}{0.946} 
& \multicolumn{1}{c|}{34.42} & \multicolumn{1}{c|}{0.410} & \multicolumn{1}{c|}{0.847}
& \multicolumn{1}{c|}{87.13} & \multicolumn{1}{c|}{0.658} & \multicolumn{1}{c|}{0.997} 
& \multicolumn{1}{c|}{63.51} & \multicolumn{1}{c|}{0.656} & \multicolumn{1}{c|}{0.997} 
& \multicolumn{1}{c|}{32.58} & \multicolumn{1}{c|}{0.558} & \multicolumn{1}{c|}{0.997}
\\ \hline \cline{1-7} 
\multicolumn{1}{|c||}{MS1M-RF-i50-Arc-Morph} 
& \multicolumn{1}{c|}{95.60} & \multicolumn{1}{c|}{0.473} & \multicolumn{1}{c|}{0.991} 
& \multicolumn{1}{c|}{83.42} & \multicolumn{1}{c|}{0.468} & \multicolumn{1}{c|}{0.970} 
& \multicolumn{1}{c|}{60.49} & \multicolumn{1}{c|}{0.416} & \multicolumn{1}{c|}{0.846}
& \multicolumn{1}{c|}{95.64} & \multicolumn{1}{c|}{0.573} & \multicolumn{1}{c|}{0.997} 
& \multicolumn{1}{c|}{83.34} & \multicolumn{1}{c|}{0.566} & \multicolumn{1}{c|}{0.997} 
& \multicolumn{1}{c|}{60.10} & \multicolumn{1}{c|}{0.554} & \multicolumn{1}{c|}{0.997}
\\ \hline \cline{1-7} 
\multicolumn{1}{|c||}{WF4M-i50-Ada-FairFace} 
& \multicolumn{1}{c|}{84.00} & \multicolumn{1}{c|}{0.736} & \multicolumn{1}{c|}{0.916} 
& \multicolumn{1}{c|}{65.66} & \multicolumn{1}{c|}{0.650} & \multicolumn{1}{c|}{0.807} 
& \multicolumn{1}{c|}{42.97} & \multicolumn{1}{c|}{0.524} & \multicolumn{1}{c|}{0.582}
& \multicolumn{1}{c|}{84.30} & \multicolumn{1}{c|}{1.306} & \multicolumn{1}{c|}{0.942} 
& \multicolumn{1}{c|}{65.51} & \multicolumn{1}{c|}{1.129} & \multicolumn{1}{c|}{0.893} 
& \multicolumn{1}{c|}{43.18} & \multicolumn{1}{c|}{0.858} & \multicolumn{1}{c|}{0.756}
\\ \hline \cline{1-7} 
\multicolumn{1}{|c||}{MS1M-RF-i50-Arc-FairFace}
& \multicolumn{1}{c|}{93.78} & \multicolumn{1}{c|}{0.680} & \multicolumn{1}{c|}{0.917} 
& \multicolumn{1}{c|}{83.99} & \multicolumn{1}{c|}{0.677} & \multicolumn{1}{c|}{0.859} 
& \multicolumn{1}{c|}{61.03} & \multicolumn{1}{c|}{0.586} & \multicolumn{1}{c|}{0.605}
& \multicolumn{1}{c|}{93.81} & \multicolumn{1}{c|}{1.090} & \multicolumn{1}{c|}{0.945} 
& \multicolumn{1}{c|}{84.03} & \multicolumn{1}{c|}{1.005} & \multicolumn{1}{c|}{0.914} 
& \multicolumn{1}{c|}{61.44} & \multicolumn{1}{c|}{0.830} & \multicolumn{1}{c|}{0.762}
\\ \hline \cline{1-7} 
\end{tabular}}
\bigskip

\vspace{-4pt}
    \resizebox{\linewidth}{!}{%
    \begin{tabular}{c|cccccccccccccccccc}
    \cline{2-19} 
    \multicolumn{1}{c||}{\large{(\textsf{P1})}} & \multicolumn{9}{c|}{$\mathbf{S}$: Gender} & \multicolumn{9}{c|}{$\mathbf{S}$: Race} \\ \cline{2-19} 
\multicolumn{1}{c||}{\large{(\textcolor{black}{$d_{\mathbf{z}} = 256$})}}  & 
\multicolumn{3}{c|}{$\alpha = 0.1$}  & \multicolumn{3}{c|}{$\alpha = 1$} & \multicolumn{3}{c|}{$\alpha = 10$}
& \multicolumn{3}{c|}{$\alpha = 0.1$} & \multicolumn{3}{c|}{$\alpha = 1$} & \multicolumn{3}{c|}{$\alpha = 10$}\\ \cline{1-19}
\rowcolor{headercolor}
\multicolumn{1}{|c||}{Face Recognition Model}
& \multicolumn{1}{c|}{TMR} & \multicolumn{1}{c|}{$\I(\mathbf{Z}; \mathbf{S})$} & \multicolumn{1}{c|}{Acc on $\mathbf{S}$} 
& \multicolumn{1}{c|}{TMR} & \multicolumn{1}{c|}{$\I(\mathbf{Z}; \mathbf{S})$} & \multicolumn{1}{c|}{Acc on $\mathbf{S}$} 
& \multicolumn{1}{c|}{TMR} & \multicolumn{1}{c|}{$\I(\mathbf{Z}; \mathbf{S})$} & \multicolumn{1}{c|}{Acc on $\mathbf{S}$} 
& \multicolumn{1}{c|}{TMR} & \multicolumn{1}{c|}{$\I(\mathbf{Z}; \mathbf{S})$} & \multicolumn{1}{c|}{Acc on $\mathbf{S}$}
& \multicolumn{1}{c|}{TMR} & \multicolumn{1}{c|}{$\I(\mathbf{Z}; \mathbf{S})$} & \multicolumn{1}{c|}{Acc on $\mathbf{S}$} 
& \multicolumn{1}{c|}{TMR} & \multicolumn{1}{c|}{$\I(\mathbf{Z}; \mathbf{S})$} & \multicolumn{1}{c|}{Acc on $\mathbf{S}$}\\ \hline \cline{1-19}
%
%
\multicolumn{1}{|c||}{WF4M-i50-Ada-Morph} 
& \multicolumn{1}{c|}{91.99} & \multicolumn{1}{c|}{0.464} & \multicolumn{1}{c|}{0.992} 
& \multicolumn{1}{c|}{46.98} & \multicolumn{1}{c|}{0.444} & \multicolumn{1}{c|}{0.949} 
& \multicolumn{1}{c|}{29.56} & \multicolumn{1}{c|}{0.388} & \multicolumn{1}{c|}{0.843}
& \multicolumn{1}{c|}{91.86} & \multicolumn{1}{c|}{0.628} & \multicolumn{1}{c|}{0.997} 
& \multicolumn{1}{c|}{47.42} & \multicolumn{1}{c|}{0.705} & \multicolumn{1}{c|}{0.997} 
& \multicolumn{1}{c|}{30.99} & \multicolumn{1}{c|}{0.550} & \multicolumn{1}{c|}{0.857}
\\ \hline \cline{1-7} 
\multicolumn{1}{|c||}{MS1M-RF-i50-Arc-Morph} 
& \multicolumn{1}{c|}{93.30} & \multicolumn{1}{c|}{0.485} & \multicolumn{1}{c|}{0.992} 
& \multicolumn{1}{c|}{84.08} & \multicolumn{1}{c|}{0.492} & \multicolumn{1}{c|}{0.971} 
& \multicolumn{1}{c|}{58.62} & \multicolumn{1}{c|}{0.335} & \multicolumn{1}{c|}{0.846}
& \multicolumn{1}{c|}{94.01} & \multicolumn{1}{c|}{0.635} & \multicolumn{1}{c|}{0.997} 
& \multicolumn{1}{c|}{84.10} & \multicolumn{1}{c|}{0.707} & \multicolumn{1}{c|}{0.997} 
& \multicolumn{1}{c|}{58.24} & \multicolumn{1}{c|}{0.558} & \multicolumn{1}{c|}{0.868}
\\ \hline \cline{1-7} 
\multicolumn{1}{|c||}{WF4M-i50-Ada-FairFace} 
& \multicolumn{1}{c|}{92.34} & \multicolumn{1}{c|}{0.638} & \multicolumn{1}{c|}{0.925} 
& \multicolumn{1}{c|}{63.12} & \multicolumn{1}{c|}{0.653} & \multicolumn{1}{c|}{0.815} 
& \multicolumn{1}{c|}{39.75} & \multicolumn{1}{c|}{0.367} & \multicolumn{1}{c|}{0.576}
& \multicolumn{1}{c|}{92.41} & \multicolumn{1}{c|}{0.866} & \multicolumn{1}{c|}{0.946} 
& \multicolumn{1}{c|}{58.67} & \multicolumn{1}{c|}{0.950} & \multicolumn{1}{c|}{0.893} 
& \multicolumn{1}{c|}{38.80} & \multicolumn{1}{c|}{0.595} & \multicolumn{1}{c|}{0.756}
\\ \hline \cline{1-7} 
\multicolumn{1}{|c||}{MS1M-RF-i50-Arc-FairFace} 
& \multicolumn{1}{c|}{90.87} & \multicolumn{1}{c|}{0.636} & \multicolumn{1}{c|}{0.915} 
& \multicolumn{1}{c|}{82.01} & \multicolumn{1}{c|}{0.652} & \multicolumn{1}{c|}{0.860} 
& \multicolumn{1}{c|}{59.62} & \multicolumn{1}{c|}{0.388} & \multicolumn{1}{c|}{0.598}
& \multicolumn{1}{c|}{90.86} & \multicolumn{1}{c|}{0.899} & \multicolumn{1}{c|}{0.947} 
& \multicolumn{1}{c|}{81.98} & \multicolumn{1}{c|}{0.873} & \multicolumn{1}{c|}{0.919} 
& \multicolumn{1}{c|}{60.33} & \multicolumn{1}{c|}{0.608} & \multicolumn{1}{c|}{0.766}
\\ \hline \cline{1-7} 
\end{tabular}}

\bigskip

\vspace{-4pt}
    \resizebox{\linewidth}{!}{%
    \begin{tabular}{c|cccccccccccccccccc}
    \cline{2-19} 
    \multicolumn{1}{c||}{\large{(\textsf{P1})}} & \multicolumn{9}{c|}{$\mathbf{S}$: Gender} & \multicolumn{9}{c|}{$\mathbf{S}$: Race} \\ \cline{2-19} 
\multicolumn{1}{c||}{\large{(\textcolor{black}{$d_{\mathbf{z}} = 128$})}}  & 
\multicolumn{3}{c|}{$\alpha = 0.1$}  & \multicolumn{3}{c|}{$\alpha = 1$} & \multicolumn{3}{c|}{$\alpha = 10$}
& \multicolumn{3}{c|}{$\alpha = 0.1$} & \multicolumn{3}{c|}{$\alpha = 1$} & \multicolumn{3}{c|}{$\alpha = 10$}\\ \cline{1-19}
\rowcolor{headercolor}
\multicolumn{1}{|c||}{Face Recognition Model}
& \multicolumn{1}{c|}{TMR} & \multicolumn{1}{c|}{$\I(\mathbf{Z}; \mathbf{S})$} & \multicolumn{1}{c|}{Acc on $\mathbf{S}$} 
& \multicolumn{1}{c|}{TMR} & \multicolumn{1}{c|}{$\I(\mathbf{Z}; \mathbf{S})$} & \multicolumn{1}{c|}{Acc on $\mathbf{S}$} 
& \multicolumn{1}{c|}{TMR} & \multicolumn{1}{c|}{$\I(\mathbf{Z}; \mathbf{S})$} & \multicolumn{1}{c|}{Acc on $\mathbf{S}$} 
& \multicolumn{1}{c|}{TMR} & \multicolumn{1}{c|}{$\I(\mathbf{Z}; \mathbf{S})$} & \multicolumn{1}{c|}{Acc on $\mathbf{S}$}
& \multicolumn{1}{c|}{TMR} & \multicolumn{1}{c|}{$\I(\mathbf{Z}; \mathbf{S})$} & \multicolumn{1}{c|}{Acc on $\mathbf{S}$} 
& \multicolumn{1}{c|}{TMR} & \multicolumn{1}{c|}{$\I(\mathbf{Z}; \mathbf{S})$} & \multicolumn{1}{c|}{Acc on $\mathbf{S}$}\\ \hline \cline{1-19}
%
%
\multicolumn{1}{|c||}{WF4M-i50-Ada-Morph} 
& \multicolumn{1}{c|}{88.20} & \multicolumn{1}{c|}{0.392} & \multicolumn{1}{c|}{0.988} 
& \multicolumn{1}{c|}{67.55} & \multicolumn{1}{c|}{0.387} & \multicolumn{1}{c|}{0.952} 
& \multicolumn{1}{c|}{21.76} & \multicolumn{1}{c|}{0.205} & \multicolumn{1}{c|}{0.845}
& \multicolumn{1}{c|}{87.70} & \multicolumn{1}{c|}{0.563} & \multicolumn{1}{c|}{0.998} 
& \multicolumn{1}{c|}{67.50} & \multicolumn{1}{c|}{0.632} & \multicolumn{1}{c|}{0.997} 
& \multicolumn{1}{c|}{20.85} & \multicolumn{1}{c|}{0.375} & \multicolumn{1}{c|}{0.997}
\\ \hline \cline{1-7} 
\multicolumn{1}{|c||}{MS1M-RF-i50-Arc-Morph} 
& \multicolumn{1}{c|}{97.60} & \multicolumn{1}{c|}{0.358} & \multicolumn{1}{c|}{0.988} 
& \multicolumn{1}{c|}{85.91} & \multicolumn{1}{c|}{0.320} & \multicolumn{1}{c|}{0.974} 
& \multicolumn{1}{c|}{62.97} & \multicolumn{1}{c|}{0.278} & \multicolumn{1}{c|}{0.848}
& \multicolumn{1}{c|}{97.61} & \multicolumn{1}{c|}{0.574} & \multicolumn{1}{c|}{0.998} 
& \multicolumn{1}{c|}{86.01} & \multicolumn{1}{c|}{0.603} & \multicolumn{1}{c|}{0.997} 
& \multicolumn{1}{c|}{62.41} & \multicolumn{1}{c|}{0.421} & \multicolumn{1}{c|}{0.996}
\\ \hline \cline{1-7} 
\multicolumn{1}{|c||}{WF4M-i50-Ada-FairFace} 
& \multicolumn{1}{c|}{94.38} & \multicolumn{1}{c|}{0.437} & \multicolumn{1}{c|}{0.892} 
& \multicolumn{1}{c|}{68.70} & \multicolumn{1}{c|}{0.420} & \multicolumn{1}{c|}{0.809} 
& \multicolumn{1}{c|}{21.47} & \multicolumn{1}{c|}{0.198} & \multicolumn{1}{c|}{0.546}
& \multicolumn{1}{c|}{94.49} & \multicolumn{1}{c|}{0.716} & \multicolumn{1}{c|}{0.937} 
& \multicolumn{1}{c|}{68.49} & \multicolumn{1}{c|}{0.665} & \multicolumn{1}{c|}{0.892} 
& \multicolumn{1}{c|}{21.36} & \multicolumn{1}{c|}{0.291} & \multicolumn{1}{c|}{0.733}
\\ \hline \cline{1-7} 
\multicolumn{1}{|c||}{MS1M-RF-i50-Arc-FairFace} 
& \multicolumn{1}{c|}{98.03} & \multicolumn{1}{c|}{0.425} & \multicolumn{1}{c|}{0.890} 
& \multicolumn{1}{c|}{86.07} & \multicolumn{1}{c|}{0.412} & \multicolumn{1}{c|}{0.860} 
& \multicolumn{1}{c|}{61.11} & \multicolumn{1}{c|}{0.284} & \multicolumn{1}{c|}{0.637}
& \multicolumn{1}{c|}{97.77} & \multicolumn{1}{c|}{0.631} & \multicolumn{1}{c|}{0.933} 
& \multicolumn{1}{c|}{86.07} & \multicolumn{1}{c|}{0.657} & \multicolumn{1}{c|}{0.919} 
& \multicolumn{1}{c|}{61.25} & \multicolumn{1}{c|}{0.551} & \multicolumn{1}{c|}{0.783}
\\ \hline \cline{1-7} 
\end{tabular}}

\bigskip

\vspace{-4pt}

    \resizebox{\linewidth}{!}{%
    \begin{tabular}{c|cccccccccccccccccccccccc}
    \cline{2-25} 
    \multicolumn{1}{c||}{\large{(\textsf{P2})}} & \multicolumn{12}{c|}{$\mathbf{S}$: Gender} & \multicolumn{12}{c|}{$\mathbf{S}$: Race} \\ \cline{2-25} 
\multicolumn{1}{c||}{\large{(\textcolor{black}{$d_{\mathbf{z}} = 512$})}}  & 
\multicolumn{3}{c|}{$\alpha = 0.1$} & \multicolumn{3}{c|}{$\alpha = 0.5$} & \multicolumn{3}{c|}{$\alpha = 1$} & \multicolumn{3}{c|}{$\alpha = 10$}
& \multicolumn{3}{c|}{$\alpha = 0.1$} & \multicolumn{3}{c|}{$\alpha = 0.5$} & \multicolumn{3}{c|}{$\alpha = 1$} & \multicolumn{3}{c|}{$\alpha = 10$}\\ \cline{1-25}
\rowcolor{headercolor}
\multicolumn{1}{|c||}{Face Recognition Model}
& \multicolumn{1}{c|}{TMR} & \multicolumn{1}{c|}{$\I(\mathbf{Z}; \mathbf{S})$} & \multicolumn{1}{c|}{Acc on $\mathbf{S}$} 
& \multicolumn{1}{c|}{TMR} & \multicolumn{1}{c|}{$\I(\mathbf{Z}; \mathbf{S})$} & \multicolumn{1}{c|}{Acc on $\mathbf{S}$} 
& \multicolumn{1}{c|}{TMR} & \multicolumn{1}{c|}{$\I(\mathbf{Z}; \mathbf{S})$} & \multicolumn{1}{c|}{Acc on $\mathbf{S}$} 
& \multicolumn{1}{c|}{TMR} & \multicolumn{1}{c|}{$\I(\mathbf{Z}; \mathbf{S})$} & \multicolumn{1}{c|}{Acc on $\mathbf{S}$}
& \multicolumn{1}{c|}{TMR} & \multicolumn{1}{c|}{$\I(\mathbf{Z}; \mathbf{S})$} & \multicolumn{1}{c|}{Acc on $\mathbf{S}$} 
& \multicolumn{1}{c|}{TMR} & \multicolumn{1}{c|}{$\I(\mathbf{Z}; \mathbf{S})$} & \multicolumn{1}{c|}{Acc on $\mathbf{S}$}
& \multicolumn{1}{c|}{TMR} & \multicolumn{1}{c|}{$\I(\mathbf{Z}; \mathbf{S})$} & \multicolumn{1}{c|}{Acc on $\mathbf{S}$}
& \multicolumn{1}{c|}{TMR} & \multicolumn{1}{c|}{$\I(\mathbf{Z}; \mathbf{S})$} & \multicolumn{1}{c|}{Acc on $\mathbf{S}$}\\ \hline \cline{1-19}
%
%
\multicolumn{1}{|c||}{WF4M-i50-Ada-Morph} 
& \multicolumn{1}{c|}{81.68} & \multicolumn{1}{c|}{0.559} & \multicolumn{1}{c|}{0.986} 
& \multicolumn{1}{c|}{60.90} & \multicolumn{1}{c|}{0.570} & \multicolumn{1}{c|}{0.966} 
& \multicolumn{1}{c|}{51.86} & \multicolumn{1}{c|}{0.564} & \multicolumn{1}{c|}{0.945} 
& \multicolumn{1}{c|}{38.20} & \multicolumn{1}{c|}{0.529} & \multicolumn{1}{c|}{0.853}
& \multicolumn{1}{c|}{82.22} & \multicolumn{1}{c|}{0.788} & \multicolumn{1}{c|}{0.998} 
& \multicolumn{1}{c|}{61.07} & \multicolumn{1}{c|}{0.803} & \multicolumn{1}{c|}{0.997} 
& \multicolumn{1}{c|}{52.18} & \multicolumn{1}{c|}{0.791} & \multicolumn{1}{c|}{0.997} 
& \multicolumn{1}{c|}{36.26} & \multicolumn{1}{c|}{0.737} & \multicolumn{1}{c|}{0.996}
\\ \hline \cline{1-7} 
\multicolumn{1}{|c||}{MS1M-RF-i50-Arc-Morph} 
& \multicolumn{1}{c|}{91.18} & \multicolumn{1}{c|}{0.552} & \multicolumn{1}{c|}{0.991} 
& \multicolumn{1}{c|}{77.86} & \multicolumn{1}{c|}{0.572} & \multicolumn{1}{c|}{0.978} 
& \multicolumn{1}{c|}{73.82} & \multicolumn{1}{c|}{0.562} & \multicolumn{1}{c|}{0.962} 
& \multicolumn{1}{c|}{67.40} & \multicolumn{1}{c|}{0.524} & \multicolumn{1}{c|}{0.876}
& \multicolumn{1}{c|}{91.37} & \multicolumn{1}{c|}{0.765} & \multicolumn{1}{c|}{0.998} 
& \multicolumn{1}{c|}{77.76} & \multicolumn{1}{c|}{0.796} & \multicolumn{1}{c|}{0.977} 
& \multicolumn{1}{c|}{73.56} & \multicolumn{1}{c|}{0.794} & \multicolumn{1}{c|}{0.997} 
& \multicolumn{1}{c|}{67.82} & \multicolumn{1}{c|}{0.751} & \multicolumn{1}{c|}{0.996}
\\ \hline \cline{1-7} 
\multicolumn{1}{|c||}{WF4M-i50-Ada-FairFace} 
& \multicolumn{1}{c|}{85.56} & \multicolumn{1}{c|}{0.850} & \multicolumn{1}{c|}{0.918} 
& \multicolumn{1}{c|}{63.75} & \multicolumn{1}{c|}{0.868} & \multicolumn{1}{c|}{0.885} 
& \multicolumn{1}{c|}{54.94} & \multicolumn{1}{c|}{0.859} & \multicolumn{1}{c|}{0.853} 
& \multicolumn{1}{c|}{40.42} & \multicolumn{1}{c|}{0.809} & \multicolumn{1}{c|}{0.759}
& \multicolumn{1}{c|}{85.43} & \multicolumn{1}{c|}{1.719} & \multicolumn{1}{c|}{0.944} 
& \multicolumn{1}{c|}{63.89} & \multicolumn{1}{c|}{1.810} & \multicolumn{1}{c|}{0.926} 
& \multicolumn{1}{c|}{54.38} & \multicolumn{1}{c|}{1.794} & \multicolumn{1}{c|}{0.908} 
& \multicolumn{1}{c|}{39.47} & \multicolumn{1}{c|}{1.699} & \multicolumn{1}{c|}{0.839}
\\ \hline \cline{1-25} 
\multicolumn{1}{|c||}{MS1M-RF-i50-Arc-FairFace} 
& \multicolumn{1}{c|}{92.20} & \multicolumn{1}{c|}{0.819} & \multicolumn{1}{c|}{0.914} 
& \multicolumn{1}{c|}{78.34} & \multicolumn{1}{c|}{0.869} & \multicolumn{1}{c|}{0.891} 
& \multicolumn{1}{c|}{74.08} & \multicolumn{1}{c|}{0.863} & \multicolumn{1}{c|}{0.868} 
& \multicolumn{1}{c|}{68.00} & \multicolumn{1}{c|}{0.827} & \multicolumn{1}{c|}{0.795}
& \multicolumn{1}{c|}{92.15} & \multicolumn{1}{c|}{1.547} & \multicolumn{1}{c|}{0.944} 
& \multicolumn{1}{c|}{78.26} & \multicolumn{1}{c|}{1.796} & \multicolumn{1}{c|}{0.932} 
& \multicolumn{1}{c|}{73.36} & \multicolumn{1}{c|}{1.745} & \multicolumn{1}{c|}{0.920} 
& \multicolumn{1}{c|}{67.65} & \multicolumn{1}{c|}{1.708} & \multicolumn{1}{c|}{0.872}
\\ \hline \cline{1-25} 
\end{tabular}}
\bigskip

\vspace{-4pt}
    \resizebox{\linewidth}{!}{%
    \begin{tabular}{c|cccccccccccccccccccccccc}
    \cline{2-25} 
    \multicolumn{1}{c||}{\large{(\textsf{P2})}} & \multicolumn{12}{c|}{$\mathbf{S}$: Gender} & \multicolumn{12}{c|}{$\mathbf{S}$: Race} \\ \cline{2-25} 
\multicolumn{1}{c||}{\large{(\textcolor{black}{$d_{\mathbf{z}} = 256$})}}  & 
\multicolumn{3}{c|}{$\alpha = 0.1$} & \multicolumn{3}{c|}{$\alpha = 0.5$} & \multicolumn{3}{c|}{$\alpha = 1$} & \multicolumn{3}{c|}{$\alpha = 10$}
& \multicolumn{3}{c|}{$\alpha = 0.1$} & \multicolumn{3}{c|}{$\alpha = 0.5$} & \multicolumn{3}{c|}{$\alpha = 1$} & \multicolumn{3}{c|}{$\alpha = 10$}\\ \cline{1-25}
\rowcolor{headercolor}
\multicolumn{1}{|c||}{Face Recognition Model}
& \multicolumn{1}{c|}{TMR} & \multicolumn{1}{c|}{$\I(\mathbf{Z}; \mathbf{S})$} & \multicolumn{1}{c|}{Acc on $\mathbf{S}$} 
& \multicolumn{1}{c|}{TMR} & \multicolumn{1}{c|}{$\I(\mathbf{Z}; \mathbf{S})$} & \multicolumn{1}{c|}{Acc on $\mathbf{S}$} 
& \multicolumn{1}{c|}{TMR} & \multicolumn{1}{c|}{$\I(\mathbf{Z}; \mathbf{S})$} & \multicolumn{1}{c|}{Acc on $\mathbf{S}$} 
& \multicolumn{1}{c|}{TMR} & \multicolumn{1}{c|}{$\I(\mathbf{Z}; \mathbf{S})$} & \multicolumn{1}{c|}{Acc on $\mathbf{S}$}
& \multicolumn{1}{c|}{TMR} & \multicolumn{1}{c|}{$\I(\mathbf{Z}; \mathbf{S})$} & \multicolumn{1}{c|}{Acc on $\mathbf{S}$} 
& \multicolumn{1}{c|}{TMR} & \multicolumn{1}{c|}{$\I(\mathbf{Z}; \mathbf{S})$} & \multicolumn{1}{c|}{Acc on $\mathbf{S}$}
& \multicolumn{1}{c|}{TMR} & \multicolumn{1}{c|}{$\I(\mathbf{Z}; \mathbf{S})$} & \multicolumn{1}{c|}{Acc on $\mathbf{S}$}
& \multicolumn{1}{c|}{TMR} & \multicolumn{1}{c|}{$\I(\mathbf{Z}; \mathbf{S})$} & \multicolumn{1}{c|}{Acc on $\mathbf{S}$}\\ \hline \cline{1-19}
%
%
\multicolumn{1}{|c||}{WF4M-i50-Ada-Morph} 
& \multicolumn{1}{c|}{81.88} & \multicolumn{1}{c|}{0.585} & \multicolumn{1}{c|}{0.987}
& \multicolumn{1}{c|}{60.65} & \multicolumn{1}{c|}{0.586} & \multicolumn{1}{c|}{0.971}
& \multicolumn{1}{c|}{50.92} & \multicolumn{1}{c|}{0.569} & \multicolumn{1}{c|}{0.953}
& \multicolumn{1}{c|}{37.57} & \multicolumn{1}{c|}{0.539} & \multicolumn{1}{c|}{0.873}
& \multicolumn{1}{c|}{81.90} & \multicolumn{1}{c|}{0.773} & \multicolumn{1}{c|}{0.998} 
& \multicolumn{1}{c|}{60.66} & \multicolumn{1}{c|}{0.812} & \multicolumn{1}{c|}{0.997} 
& \multicolumn{1}{c|}{51.51} & \multicolumn{1}{c|}{0.816} & \multicolumn{1}{c|}{0.997} 
& \multicolumn{1}{c|}{38.08} & \multicolumn{1}{c|}{0.765} & \multicolumn{1}{c|}{0.996}
\\ \hline \cline{1-7} 
\multicolumn{1}{|c||}{MS1M-RF-i50-Arc-Morph} 
& \multicolumn{1}{c|}{91.58} & \multicolumn{1}{c|}{0.539} & \multicolumn{1}{c|}{0.991} 
& \multicolumn{1}{c|}{77.60} & \multicolumn{1}{c|}{0.575} & \multicolumn{1}{c|}{0.981} 
& \multicolumn{1}{c|}{72.96} & \multicolumn{1}{c|}{0.580} & \multicolumn{1}{c|}{0.968} 
& \multicolumn{1}{c|}{67.06} & \multicolumn{1}{c|}{0.549} & \multicolumn{1}{c|}{0.899}
& \multicolumn{1}{c|}{91.74} & \multicolumn{1}{c|}{0.792} & \multicolumn{1}{c|}{0.998} 
& \multicolumn{1}{c|}{77.59} & \multicolumn{1}{c|}{0.812} & \multicolumn{1}{c|}{0.997} 
& \multicolumn{1}{c|}{73.03} & \multicolumn{1}{c|}{0.812} & \multicolumn{1}{c|}{0.997} 
& \multicolumn{1}{c|}{67.31} & \multicolumn{1}{c|}{0.776} & \multicolumn{1}{c|}{0.996}
\\ \hline \cline{1-7} 
\multicolumn{1}{|c||}{WF4M-i50-Ada-FairFace} 
& \multicolumn{1}{c|}{86.67} & \multicolumn{1}{c|}{0.844} & \multicolumn{1}{c|}{0.916} 
& \multicolumn{1}{c|}{63.64} & \multicolumn{1}{c|}{0.865} & \multicolumn{1}{c|}{0.892} 
& \multicolumn{1}{c|}{54.41} & \multicolumn{1}{c|}{0.830} & \multicolumn{1}{c|}{0.865} 
& \multicolumn{1}{c|}{40.61} & \multicolumn{1}{c|}{0.771} & \multicolumn{1}{c|}{0.762}
& \multicolumn{1}{c|}{86.61} & \multicolumn{1}{c|}{1.611} & \multicolumn{1}{c|}{0.944} 
& \multicolumn{1}{c|}{63.62} & \multicolumn{1}{c|}{1.699} & \multicolumn{1}{c|}{0.930} 
& \multicolumn{1}{c|}{54.43} & \multicolumn{1}{c|}{1.653} & \multicolumn{1}{c|}{0.916} 
& \multicolumn{1}{c|}{39.75} & \multicolumn{1}{c|}{1.503} & \multicolumn{1}{c|}{0.855}
\\ \hline \cline{1-25} 
\multicolumn{1}{|c||}{MS1M-RF-i50-Arc-FairFace} 
& \multicolumn{1}{c|}{92.34} & \multicolumn{1}{c|}{0.845} & \multicolumn{1}{c|}{0.915} 
& \multicolumn{1}{c|}{77.51} & \multicolumn{1}{c|}{0.863} & \multicolumn{1}{c|}{0.901} 
& \multicolumn{1}{c|}{73.00} & \multicolumn{1}{c|}{0.853} & \multicolumn{1}{c|}{0.882} 
& \multicolumn{1}{c|}{67.51} & \multicolumn{1}{c|}{0.779} & \multicolumn{1}{c|}{0.803}
& \multicolumn{1}{c|}{92.35} & \multicolumn{1}{c|}{1.528} & \multicolumn{1}{c|}{0.943} 
& \multicolumn{1}{c|}{77.48} & \multicolumn{1}{c|}{1.701} & \multicolumn{1}{c|}{0.936} 
& \multicolumn{1}{c|}{72.76} & \multicolumn{1}{c|}{1.678} & \multicolumn{1}{c|}{0.926} 
& \multicolumn{1}{c|}{66.90} & \multicolumn{1}{c|}{1.571} & \multicolumn{1}{c|}{0.882}
\\ \hline \cline{1-25} 
\end{tabular}}

\bigskip

\vspace{-4pt}
    \resizebox{\linewidth}{!}{%
    \begin{tabular}{c|cccccccccccccccccccccccc}
    \cline{2-25} 
    \multicolumn{1}{c||}{\large{(\textsf{P2})}} & \multicolumn{12}{c|}{$\mathbf{S}$: Gender} & \multicolumn{12}{c|}{$\mathbf{S}$: Race} \\ \cline{2-25} 
\multicolumn{1}{c||}{\large{(\textcolor{black}{$d_{\mathbf{z}} = 128$})}}  & 
\multicolumn{3}{c|}{$\alpha = 0.1$} & \multicolumn{3}{c|}{$\alpha = 0.5$} & \multicolumn{3}{c|}{$\alpha = 1$} & \multicolumn{3}{c|}{$\alpha = 10$}
& \multicolumn{3}{c|}{$\alpha = 0.1$} & \multicolumn{3}{c|}{$\alpha = 0.5$} & \multicolumn{3}{c|}{$\alpha = 1$} & \multicolumn{3}{c|}{$\alpha = 10$}\\ \cline{1-25}
\rowcolor{headercolor}
\multicolumn{1}{|c||}{Face Recognition Model}
& \multicolumn{1}{c|}{TMR} & \multicolumn{1}{c|}{$\I(\mathbf{Z}; \mathbf{S})$} & \multicolumn{1}{c|}{Acc on $\mathbf{S}$} 
& \multicolumn{1}{c|}{TMR} & \multicolumn{1}{c|}{$\I(\mathbf{Z}; \mathbf{S})$} & \multicolumn{1}{c|}{Acc on $\mathbf{S}$} 
& \multicolumn{1}{c|}{TMR} & \multicolumn{1}{c|}{$\I(\mathbf{Z}; \mathbf{S})$} & \multicolumn{1}{c|}{Acc on $\mathbf{S}$} 
& \multicolumn{1}{c|}{TMR} & \multicolumn{1}{c|}{$\I(\mathbf{Z}; \mathbf{S})$} & \multicolumn{1}{c|}{Acc on $\mathbf{S}$}
& \multicolumn{1}{c|}{TMR} & \multicolumn{1}{c|}{$\I(\mathbf{Z}; \mathbf{S})$} & \multicolumn{1}{c|}{Acc on $\mathbf{S}$} 
& \multicolumn{1}{c|}{TMR} & \multicolumn{1}{c|}{$\I(\mathbf{Z}; \mathbf{S})$} & \multicolumn{1}{c|}{Acc on $\mathbf{S}$}
& \multicolumn{1}{c|}{TMR} & \multicolumn{1}{c|}{$\I(\mathbf{Z}; \mathbf{S})$} & \multicolumn{1}{c|}{Acc on $\mathbf{S}$}
& \multicolumn{1}{c|}{TMR} & \multicolumn{1}{c|}{$\I(\mathbf{Z}; \mathbf{S})$} & \multicolumn{1}{c|}{Acc on $\mathbf{S}$}\\ \hline \cline{1-19}
%
%
\multicolumn{1}{|c||}{WF4M-i50-Ada-Morph} 
& \multicolumn{1}{c|}{84.06} & \multicolumn{1}{c|}{0.556} & \multicolumn{1}{c|}{0.984} 
& \multicolumn{1}{c|}{62.62} & \multicolumn{1}{c|}{0.575} & \multicolumn{1}{c|}{0.973} 
& \multicolumn{1}{c|}{52.05} & \multicolumn{1}{c|}{0.572} & \multicolumn{1}{c|}{0.963} 
& \multicolumn{1}{c|}{36.76} & \multicolumn{1}{c|}{0.531} & \multicolumn{1}{c|}{0.906}
& \multicolumn{1}{c|}{84.14} & \multicolumn{1}{c|}{0.810} & \multicolumn{1}{c|}{0.998} 
& \multicolumn{1}{c|}{62.94} & \multicolumn{1}{c|}{0.827} & \multicolumn{1}{c|}{0.997} 
& \multicolumn{1}{c|}{52.34} & \multicolumn{1}{c|}{0.820} & \multicolumn{1}{c|}{0.997} 
& \multicolumn{1}{c|}{36.26} & \multicolumn{1}{c|}{0.789} & \multicolumn{1}{c|}{0.996}
\\ \hline \cline{1-7} 
\multicolumn{1}{|c||}{MS1M-RF-i50-Arc-Morph} 
& \multicolumn{1}{c|}{93.00} & \multicolumn{1}{c|}{0.541} & \multicolumn{1}{c|}{0.987} 
& \multicolumn{1}{c|}{79.40} & \multicolumn{1}{c|}{0.573} & \multicolumn{1}{c|}{0.981} 
& \multicolumn{1}{c|}{73.50} & \multicolumn{1}{c|}{0.572} & \multicolumn{1}{c|}{0.974} 
& \multicolumn{1}{c|}{66.28} & \multicolumn{1}{c|}{0.535} & \multicolumn{1}{c|}{0.927}
& \multicolumn{1}{c|}{93.10} & \multicolumn{1}{c|}{0.800} & \multicolumn{1}{c|}{0.998} 
& \multicolumn{1}{c|}{79.99} & \multicolumn{1}{c|}{0.828} & \multicolumn{1}{c|}{0.997} 
& \multicolumn{1}{c|}{74.04} & \multicolumn{1}{c|}{0.825} & \multicolumn{1}{c|}{0.997} 
& \multicolumn{1}{c|}{65.94} & \multicolumn{1}{c|}{0.793} & \multicolumn{1}{c|}{0.996}
\\ \hline \cline{1-7} 
\multicolumn{1}{|c||}{WF4M-i50-Ada-FairFace} 
& \multicolumn{1}{c|}{88.51} & \multicolumn{1}{c|}{0.724} & \multicolumn{1}{c|}{0.893} 
& \multicolumn{1}{c|}{67.43} & \multicolumn{1}{c|}{0.738} & \multicolumn{1}{c|}{0.870} 
& \multicolumn{1}{c|}{57.44} & \multicolumn{1}{c|}{0.729} & \multicolumn{1}{c|}{0.854} 
& \multicolumn{1}{c|}{39.27} & \multicolumn{1}{c|}{0.676} & \multicolumn{1}{c|}{0.800}
& \multicolumn{1}{c|}{88.55} & \multicolumn{1}{c|}{1.375} & \multicolumn{1}{c|}{0.938} 
& \multicolumn{1}{c|}{67.34} & \multicolumn{1}{c|}{1.503} & \multicolumn{1}{c|}{0.926} 
& \multicolumn{1}{c|}{57.28} & \multicolumn{1}{c|}{1.479} & \multicolumn{1}{c|}{0.916} 
& \multicolumn{1}{c|}{39.60} & \multicolumn{1}{c|}{1.359} & \multicolumn{1}{c|}{0.877}
\\ \hline \cline{1-25} 
\multicolumn{1}{|c||}{MS1M-RF-i50-Arc-FairFace} 
& \multicolumn{1}{c|}{94.14} & \multicolumn{1}{c|}{0.700} & \multicolumn{1}{c|}{0.890} 
& \multicolumn{1}{c|}{81.81} & \multicolumn{1}{c|}{0.749} & \multicolumn{1}{c|}{0.878} 
& \multicolumn{1}{c|}{75.95} & \multicolumn{1}{c|}{0.743} & \multicolumn{1}{c|}{0.869} 
& \multicolumn{1}{c|}{67.16} & \multicolumn{1}{c|}{0.719} & \multicolumn{1}{c|}{0.836}
& \multicolumn{1}{c|}{94.23} & \multicolumn{1}{c|}{1.136} & \multicolumn{1}{c|}{0.934} 
& \multicolumn{1}{c|}{81.67} & \multicolumn{1}{c|}{1.381} & \multicolumn{1}{c|}{0.927} 
& \multicolumn{1}{c|}{75.96} & \multicolumn{1}{c|}{1.404} & \multicolumn{1}{c|}{0.922} 
& \multicolumn{1}{c|}{67.16} & \multicolumn{1}{c|}{1.368} & \multicolumn{1}{c|}{0.903}
\\ \hline \cline{1-25} 
\end{tabular}}
\vspace{-8pt}
\end{table*}

\vspace{1pt}

\subsubsection{Evaluation of Morph and FairFace Datasets After Applying DVPF}

We applied our deep variational $\mathsf{DisPF}$ \eqref{Eq:DVPF_P1_DisPF} and $\mathsf{GenPF}$ \eqref{Eq:DVPF_P2_GenPF} models to the embeddings obtained from the FR models referenced in \autoref{Table:PerformanceMetrics_FacialRecognitionModels}. The assessment was initiated with the pre-trained backbones, followed by our $\mathsf{DisPF}$ or $\mathsf{GenPF}$ model, which was developed using embeddings from these pre-trained structures. 
\autoref{TrainDVPF_FaceRecognition_EmbeddingBased} represents our training framework for the deep $\mathsf{DisPF}$ problem, using iResNet50 as the backbone, WebFace4M as the backbone dataset, and ArcFace for the FR loss. The applied dataset is FairFace, with race as the sensitive attribute. We considered a similar embedding-based learning framework for the deep $\mathsf{GenPF}$ problem. Given the consistent accuracy for sensitive attribute $\mathbf{S}$ and similar information leakage $\I \left( \mathbf{X}; \mathbf{S} \right)$ observed across various iResNet architectures, we present results specific to iResNet50.

In \autoref{Table:PerformanceMetrics_FacialRecognitionModels_AfterPrivacyFunnel}, we quantify the disclosed information leakage, represented as $\I(\mathbf{S}; \mathbf{Z})$. Additionally, we provide a detailed account of the accuracy achieved in recognizing sensitive attributes from the disclosed representation $\mathbf{Z} \in \mathbb{R}^{256}$, utilizing a support vector classifier optimization. These evaluations are based on test sets derived from either the Morph or FairFace datasets.
Consistent with our expectations, as $\alpha$ increases towards infinity ($\alpha \rightarrow \infty$), the information leakage $\I (\mathbf{S}; \mathbf{Z})$ decreases to zero. At the same time, the recognition accuracy for the sensitive attribute $\mathbf{S}$ approaches $0.5$, indicative of \textit{random guessing}.


\subsubsection{TMR Benchmark on IJB-C in FairFace Experiments}

\vspace{-1pt}

\begin{figure}[t!]
    \centering
    \begin{subfigure}[h]{0.5\linewidth}
        \includegraphics[width=\linewidth]{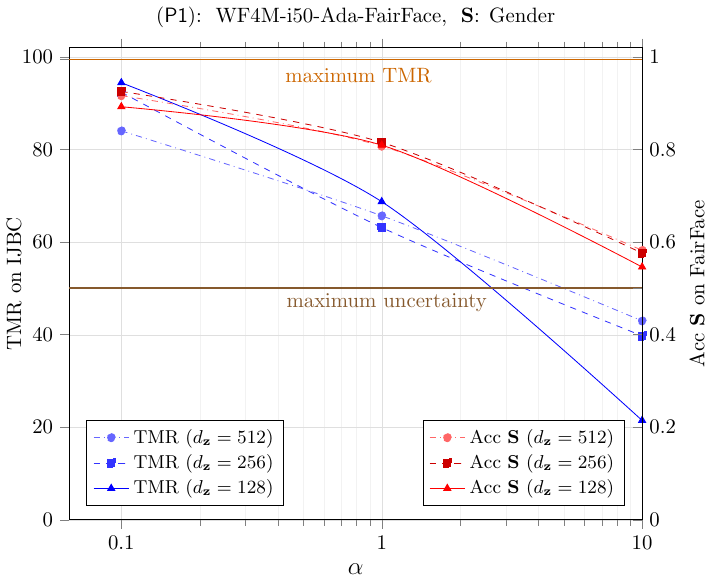}%
           \vspace{-6pt}
        \caption{}
        \label{fig:P1_WF4M-i50-Ada-FairFace_S-gender}
    \end{subfigure}%
~
    \begin{subfigure}[h]{0.5\linewidth}
        \includegraphics[width=\textwidth]{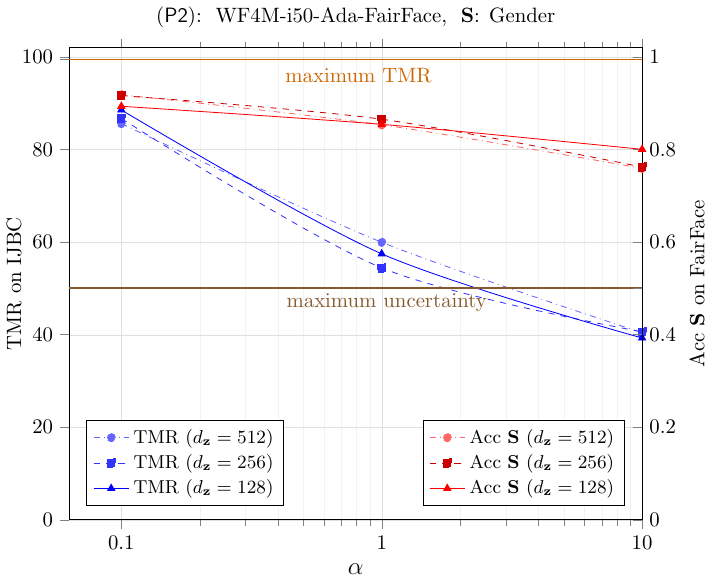}%
        \vspace{-6pt}
        \caption{}
        \label{fig:P2_WF4M-i50-Ada-FairFace_S-gender}
    \end{subfigure}%

    \begin{subfigure}[h]{0.5\linewidth}
        \includegraphics[width=\textwidth]{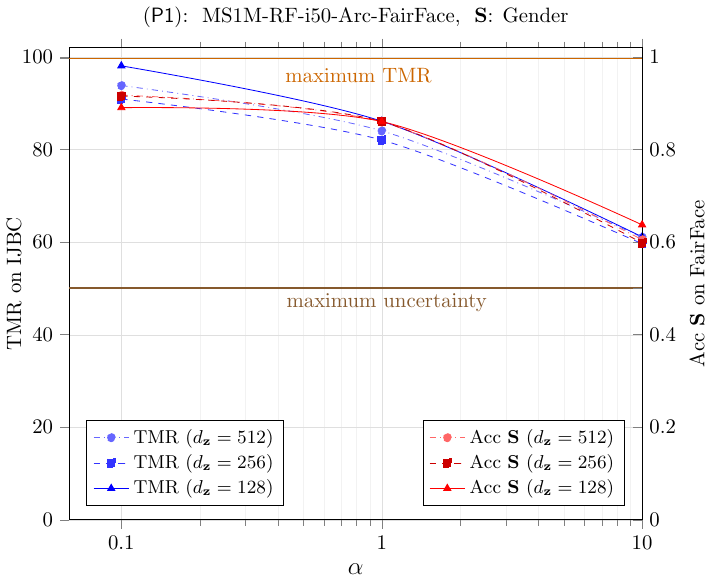}%
        \vspace{-6pt}
        \caption{}
        \label{fig:P1_MS1M-RF-i50-Arc-FairFace_S-gender}
    \end{subfigure}%
~
    \begin{subfigure}[h]{0.5\linewidth}
        \includegraphics[width=\textwidth]{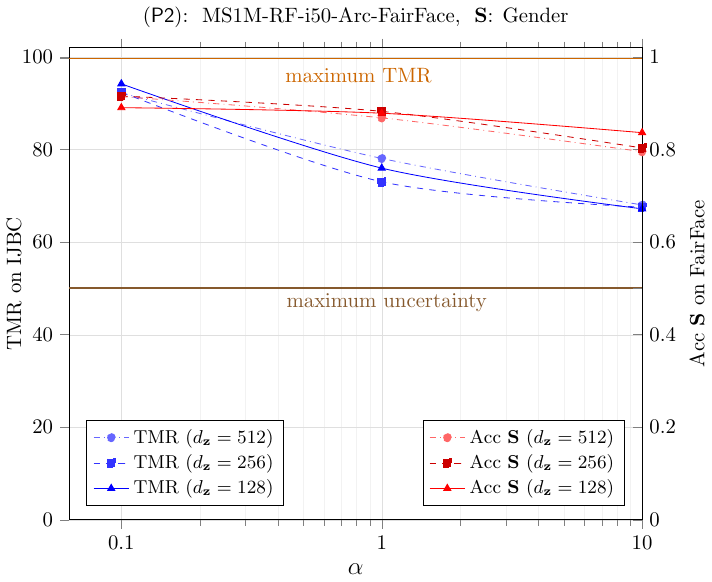}%
        \vspace{-6pt}
        \caption{}
        \label{fig:P2_MS1M-RF-i50-Arc-FairFace_S-gender}
    \end{subfigure}
    \vspace{-3pt}
    \caption{Trade-off between information utility and privacy leakage using DVPF models for gender attribute: comparing classification accuracy on FairFace and TMR on IJB-C.}
    \label{Fig:P1_P2_WF4M-MS1M_i50-Arc_Ada-FairFace_S-gender}
   \vspace{-10pt}
\end{figure}

\begin{figure}[t!]
    \centering
    \begin{subfigure}[h]{0.5\linewidth}
        \includegraphics[width=\linewidth]{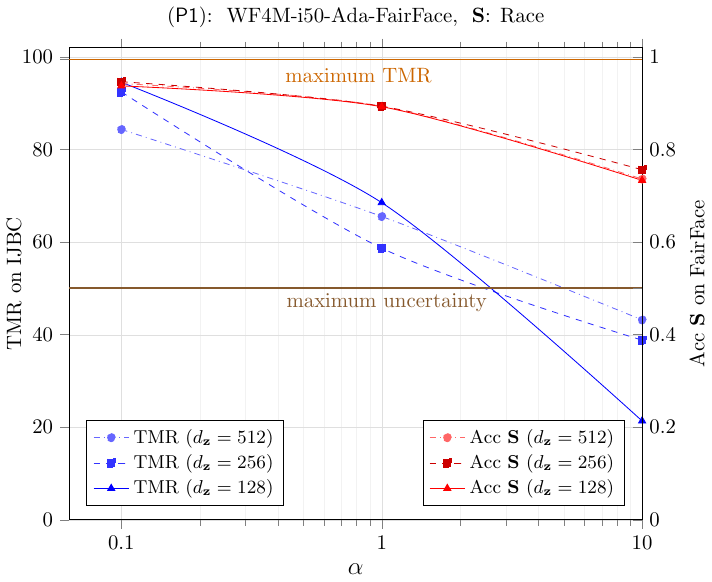}%
          \vspace{-6pt}
        \caption{}
        \label{fig:P1_WF4M-i50-Ada-FairFace_S-race}
    \end{subfigure}%
~
    \begin{subfigure}[h]{0.5\linewidth}
        \includegraphics[width=\linewidth]{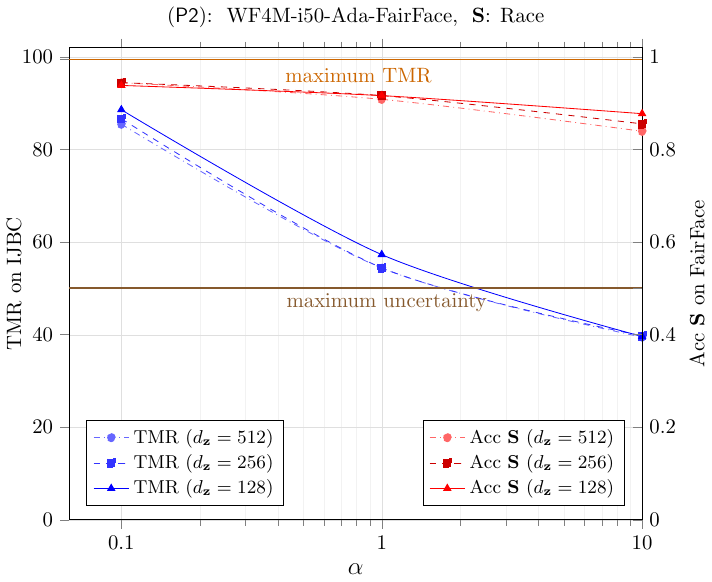}%
        \vspace{-6pt}
        \caption{}
        \label{fig:P2_WF4M-i50-Ada-FairFace_S-race}
    \end{subfigure}%
\\
    \begin{subfigure}[h]{0.5\linewidth}
        \includegraphics[width=\textwidth]{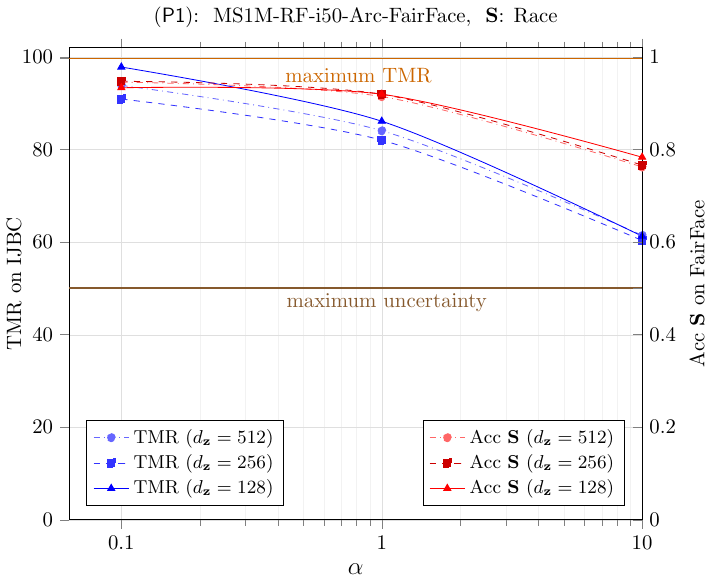}%
        \vspace{-6pt}
        \caption{}
        \label{fig:P1_MS1M-RF-i50-Arc-FairFace_S-race}
    \end{subfigure}%
~
    \begin{subfigure}[h]{0.5\linewidth}
        \includegraphics[width=\linewidth]{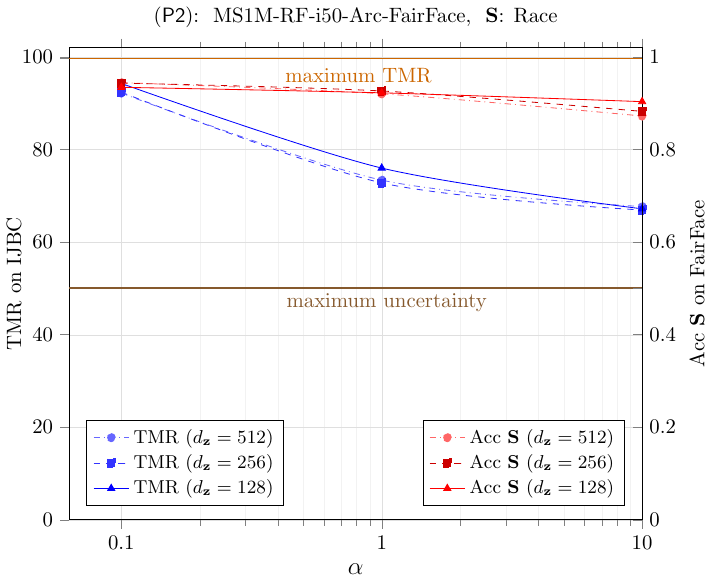}%
        \vspace{-6pt}
        \caption{}
        \label{fig:P2_MS1M-RF-i50-Arc-FairFace_S-race}
    \end{subfigure}
   \vspace{-3pt}
    \caption{Trade-off between information utility and privacy leakage using DVPF models for race attribute: comparing classification accuracy on FairFace and TMR on IJB-C.}
    \label{Fig:P1_P2_WF4M-MS1M_i50-Arc_Ada-FairFace_S-race}
  \vspace{-8pt}
\end{figure}

To evaluate the generalization of our mechanisms in terms of FR accuracy, we utilized the challenging IJB-C test dataset \cite{ijbc} as a challenging benchmark. \autoref{InferenceDVPF_FaceRecognition_EmbeddingBased} 
depicts our inference framework, which incorporates the $\mathsf{DisPF}$ trained module. We employ a similar inference framework for the $\mathsf{GenPF}$ trained module.
We detail the $\mathsf{TMR}$ of our models in \autoref{Table:PerformanceMetrics_FacialRecognitionModels_AfterPrivacyFunnel}. It's imperative to note that all these evaluations are systematically benchmarked against a predetermined False Match Rate ($\mathsf{FMR}$) of $10^{-1}$.%
When subjecting the `WF4M-i50-Ada' model to evaluation against the IJB-C dataset---prior to the DVPF model's integration---a $\mathsf{TMR}$ of $\mathsf{99.40\%}$ at $\mathsf{FMR=10e-1}$ was observed. Similarly, for the `MS1M-RF-i50-Arc' configuration, a $\mathsf{TMR}$ of $\mathsf{99.58\%}$ was observed on the IJB-C dataset before the integration of the DVPF model, with measurements anchored to the same $\mathsf{FMR}$.
In \autoref{Fig:P1_P2_WF4M-MS1M_i50-Arc_Ada-FairFace_S-gender} and \autoref{Fig:P1_P2_WF4M-MS1M_i50-Arc_Ada-FairFace_S-race}, we demonstrate the interplay between information utility and privacy leakage across varying information leakage weights $\alpha$. The right y-axis quantifies the classification accuracy of the sensitive attribute $\mathbf{S}$, as evaluated on the FairFace dataset. In contrast, the left y-axis depicts the $\mathsf{TMR}$ on the IJB-C test dataset. This measurement is derived from the performance of trained Deep Variational Privacy Filtering (DVPF) models $(\textsf{P1})$ and $(\textsf{P2})$, initially trained on the FairFace dataset and subsequently tested on the IJB-C dataset.

\vspace{-2pt}

\autoref{Fig:P1_P2_WF4M-MS1M_i50-Arc_Ada-FairFace_S-gender} focuses on the results obtained using the WF4M-i50-Ada-FairFace configuration (where `Backbone Dataset' is WebFace4M, `Backbone Architecture' is iResNet50, `Loss Function' is AdaFace, and `Applied Dataset' for training is FairFace, `Dataset for Testing Utility' ($\mathsf{TMR}$) being IJB-C) and MS1M-RF-i50-Arc-FairFace configuration (with `Backbone Dataset' as MS1M-RetinaFace, `Backbone Architecture' as iResNet50, `Loss Function' as ArcFace, and `Applied Dataset' for training as FairFace; `Dataset for Testing Utility' ($\mathsf{TMR}$) being IJB-C) when the sensitive attribute under consideration for training is gender. \autoref{Fig:P1_P2_WF4M-MS1M_i50-Arc_Ada-FairFace_S-race} presents analogous results, but for cases where the sensitive attribute for training is race.

%

%

%
\begin{figure*}[t!]
    \centering
    \begin{subfigure}[h]{0.2\linewidth}
        \includegraphics[width=\linewidth,clip]{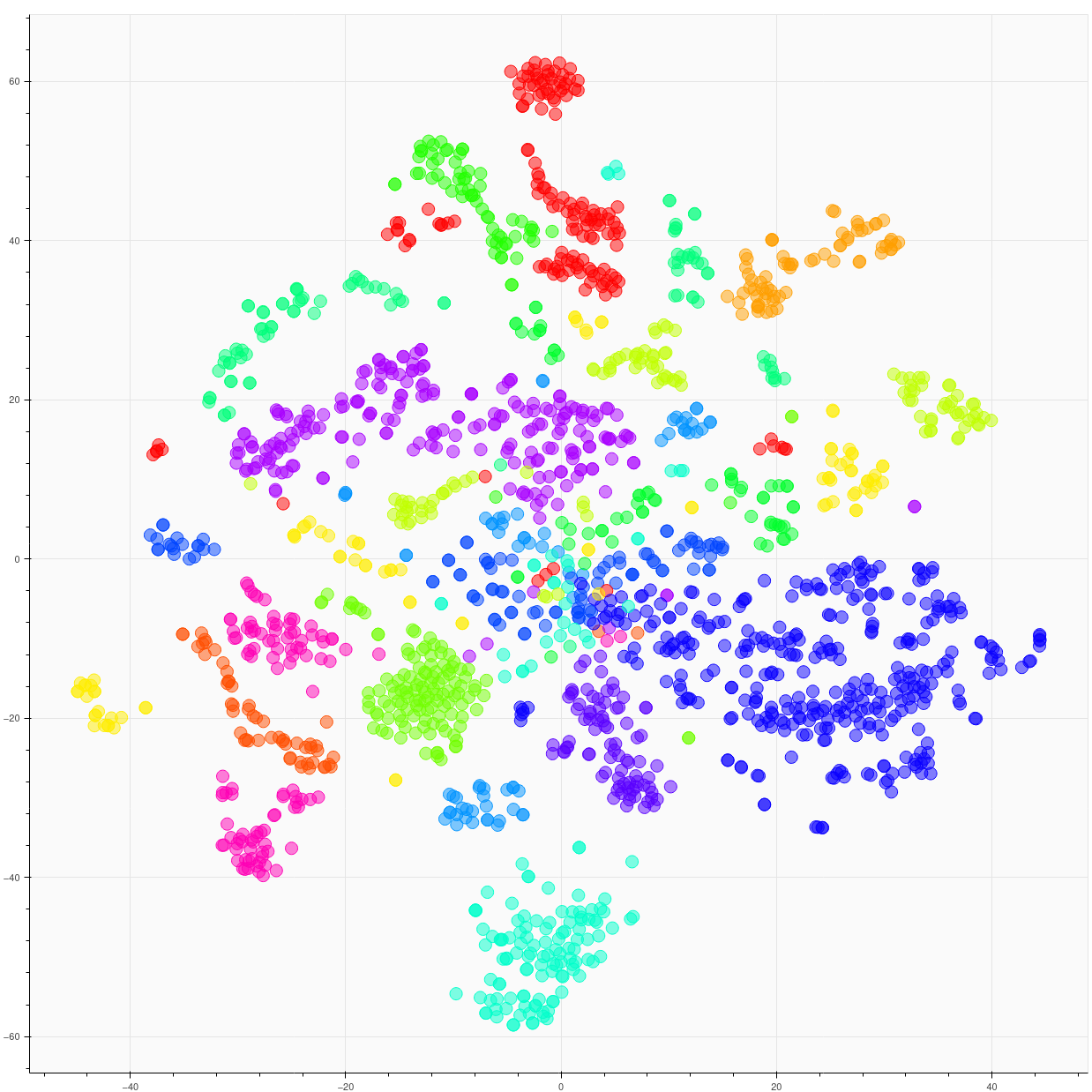}%
           \vspace{-6pt}
        \caption{}
        \label{fig:tSNE_arcface_r50_SUBJECT_ID}
    \end{subfigure}%
    \begin{subfigure}[h]{0.2\linewidth}
        \includegraphics[width=\textwidth,clip]{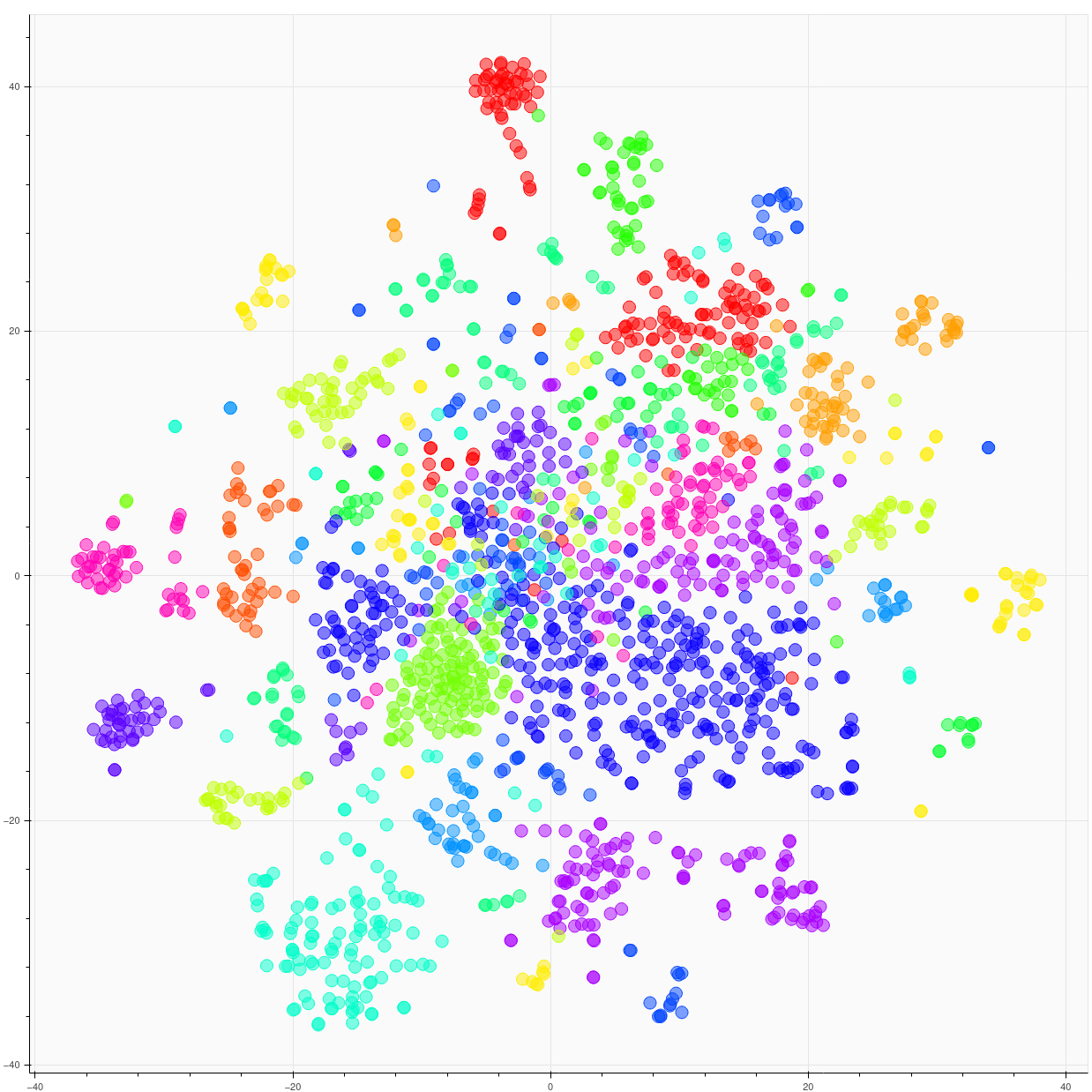}%
        \vspace{-6pt}
        \caption{}
        \label{fig:tSNE_dvpf_arcface_r50_fairface_race_0.1_LB_128_arcface_r50_MS1Mv3_SUBJECT_ID}
    \end{subfigure}%
    ~~~~
    \begin{subfigure}[h]{0.2\linewidth}
        \includegraphics[width=\textwidth,clip]{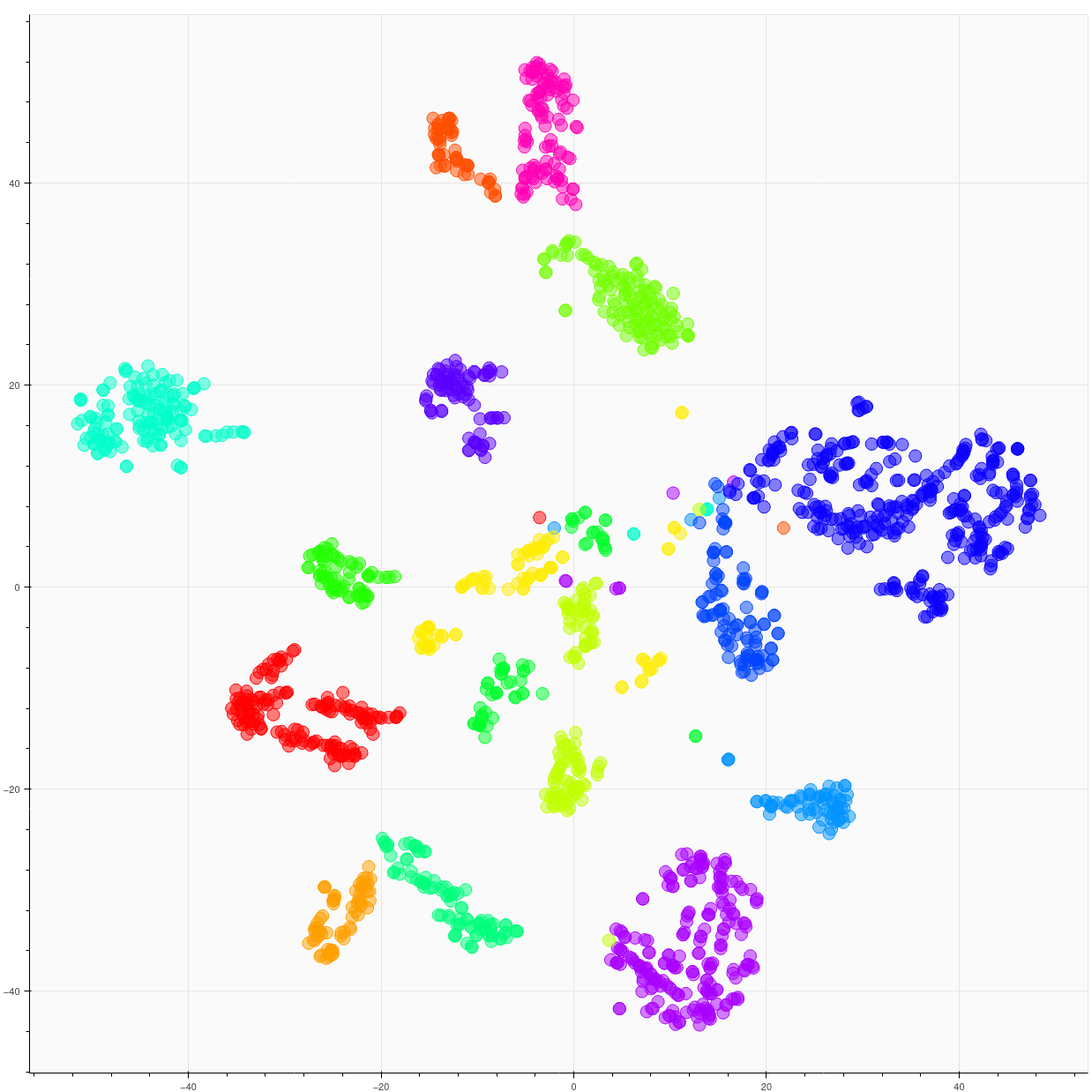}%
        \vspace{-6pt}
        \caption{}
        \label{fig:tSNE_adaface_ir50_webface4m_SUBJECT_ID}
    \end{subfigure}%
    \begin{subfigure}[h]{0.205\linewidth}
        \includegraphics[width=\textwidth,clip]{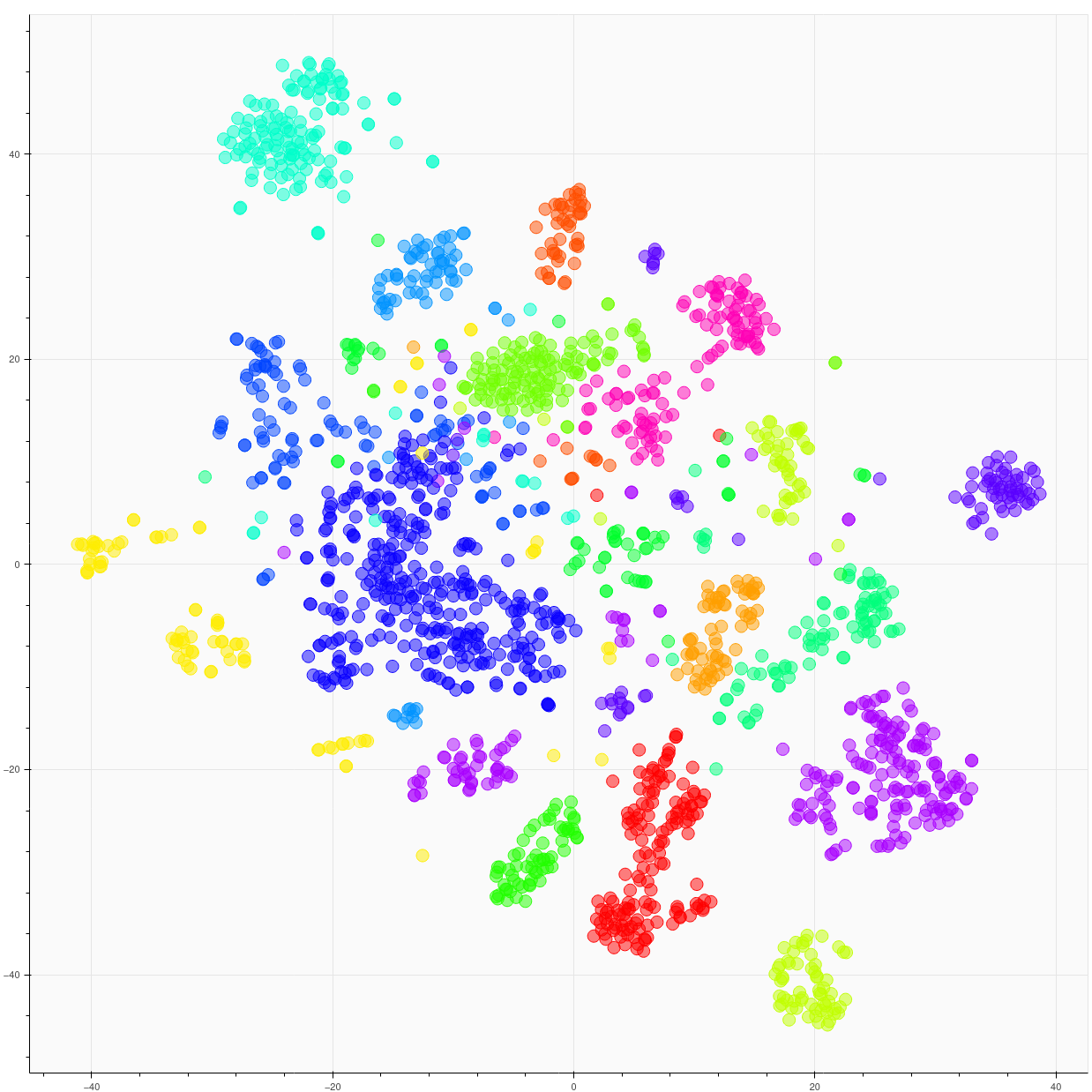}%
        \vspace{-6pt}
        \caption{}
    \label{fig:tSNE_dvpf_adaface_ir50_webface4m_fairface_race_0.1_LB_128_adaface_ir50_webface4m_SUBJECT_ID}
    \end{subfigure}%
   \vspace{-6pt}
    \caption{t-SNE visualizations of 16 randomly selected identities on the IJB-C dataset: (a) ArcFace, (b) ArcFace with DVPF, (c) AdaFace, (d) AdaFace with DVPF.}
    \vspace{-5pt}
    \label{Fig:tSNE_on_IJBC_identity}
\end{figure*}

%
\begin{figure*}[t!]
    \centering
    \begin{subfigure}[h]{0.2\linewidth}
        \includegraphics[width=\linewidth]{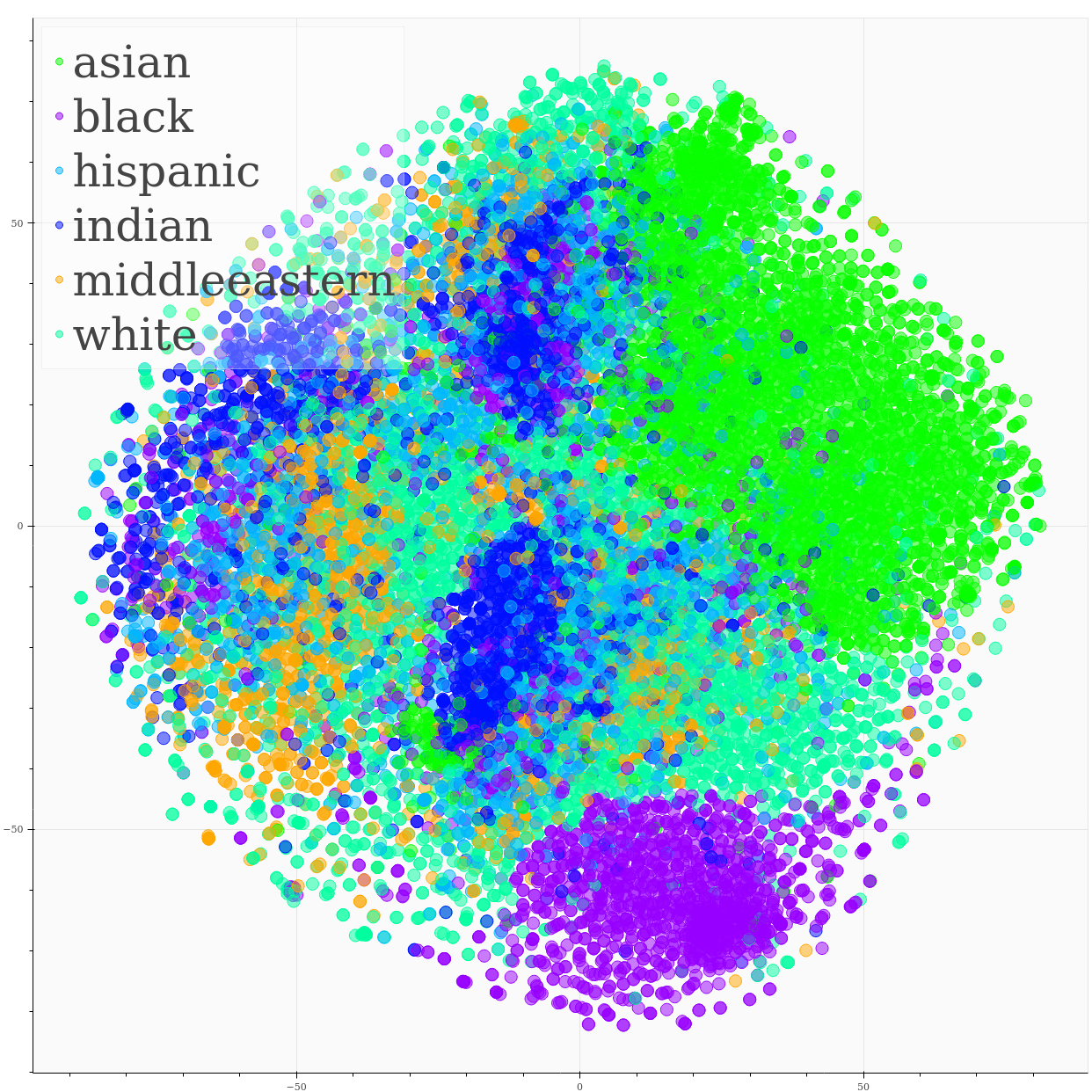}%
           \vspace{-6pt}
        \caption{}
        \label{fig:tsne_adaface_ir50_webface4m_race}
    \end{subfigure}%
    \begin{subfigure}[h]{0.2\linewidth}
        \includegraphics[width=\textwidth]{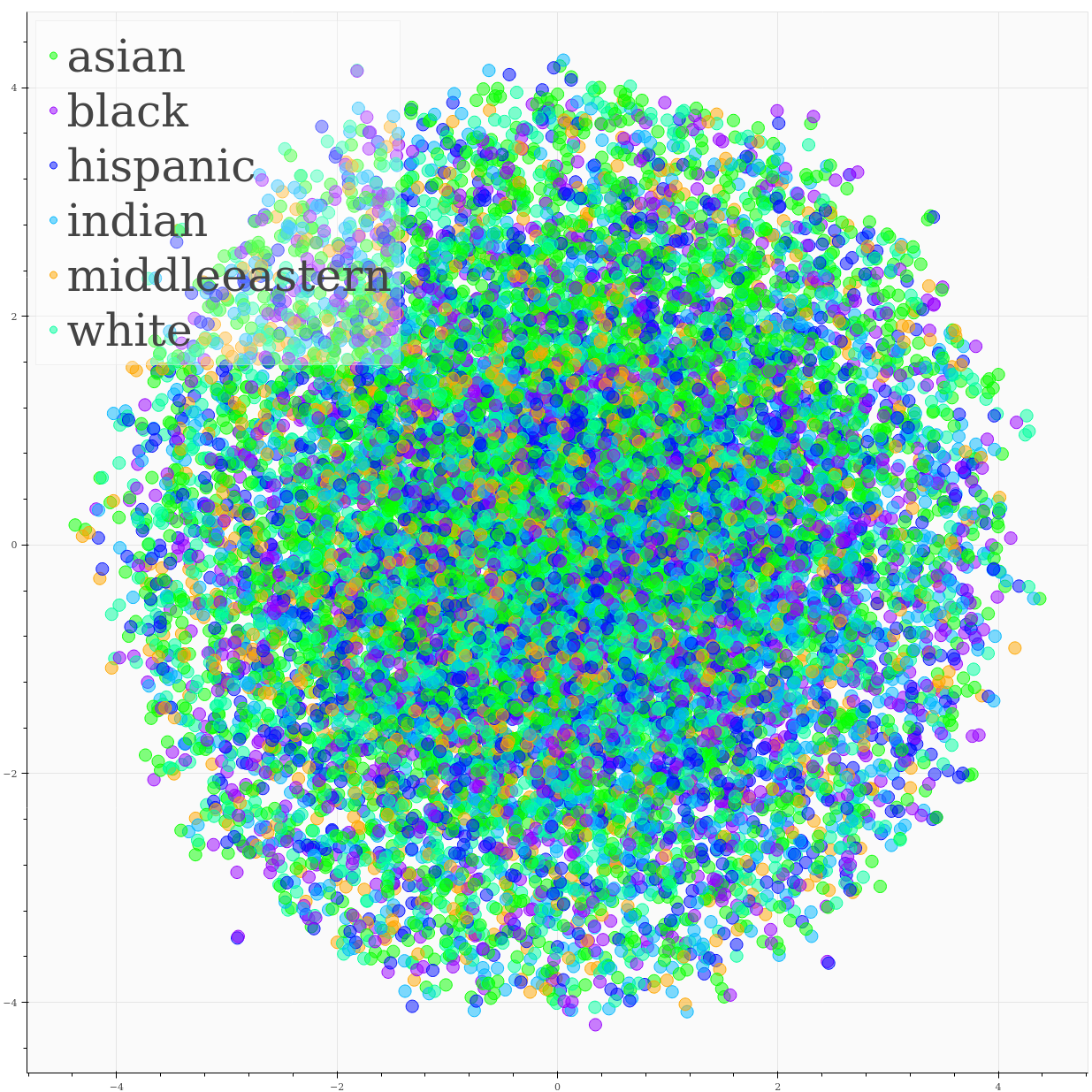}%
        \vspace{-6pt}
        \caption{}
        \label{fig:tsne_dvpf_adaface_ir50_webface4m_fairface_race_10_UB_128_adaface_ir50_webface4m_race}
    \end{subfigure}%
    ~~~~
    \begin{subfigure}[h]{0.2\linewidth}
        \includegraphics[width=\textwidth]{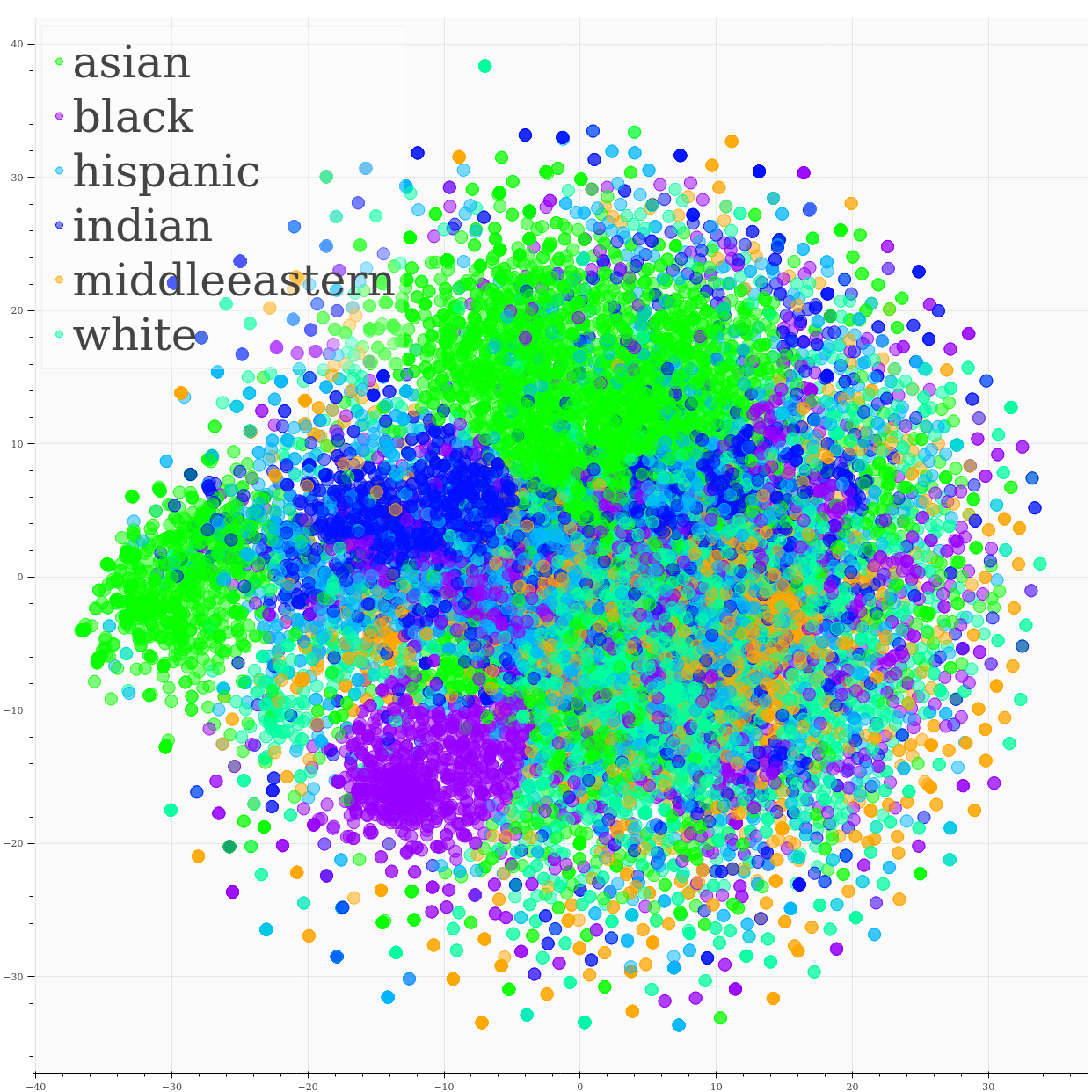}%
        \vspace{-6pt}
        \caption{}
        \label{fig:tsne_arcface_r50_race}
    \end{subfigure}%
    \begin{subfigure}[h]{0.2\linewidth}
        \includegraphics[width=\textwidth]{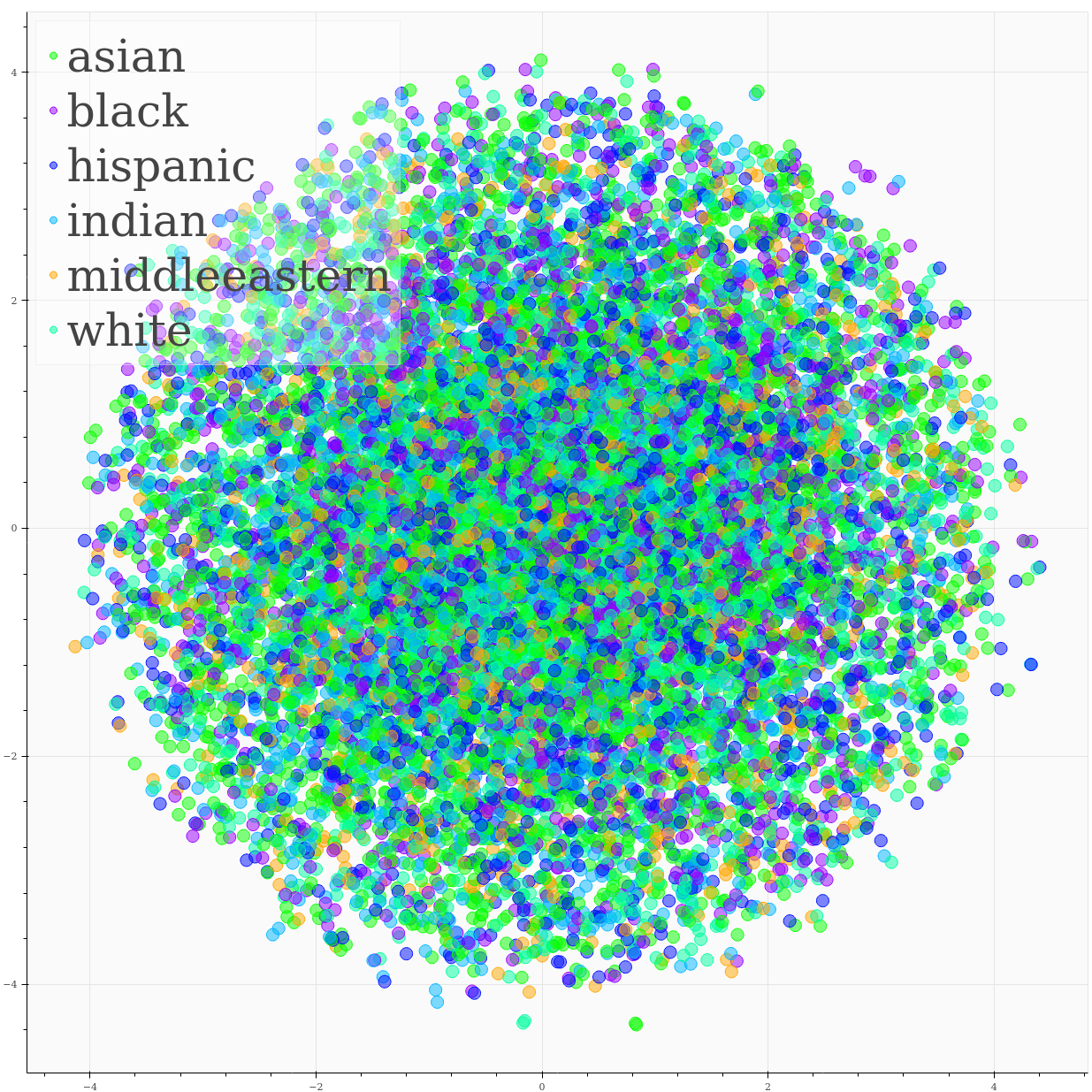}%
        \vspace{-6pt}
        \caption{}
        \label{fig:tsne_dvpf_arcface_r50_fairface_race_10_UB_128_arcface_r50_MS1Mv3_race}
    \end{subfigure}%
    ~~~~
    \vspace{-6pt}
    \caption{t-SNE visualizations of the FairFace dataset with $\mathbf{S}$ representing `\textit{race}', using the (\textsf{P2}) model, setting $\alpha = 10$ and $d_{\mathbf{z}} = 128$. The visualizations include: 
    (a) AdaFace original (clean) embeddings, 
    (b) Post-DVPF AdaFace embeddings,
    (c) ArcFace original (clean) embeddings, 
    and 
    (d) Post-DVPF ArcFace embeddings.}
    \vspace{-12pt}
    \label{Fig:tSNE_on_sensitive_attribute_race}
\end{figure*}

%
\begin{figure}[t!]
    \centering 
    \includegraphics[width=0.99\linewidth]{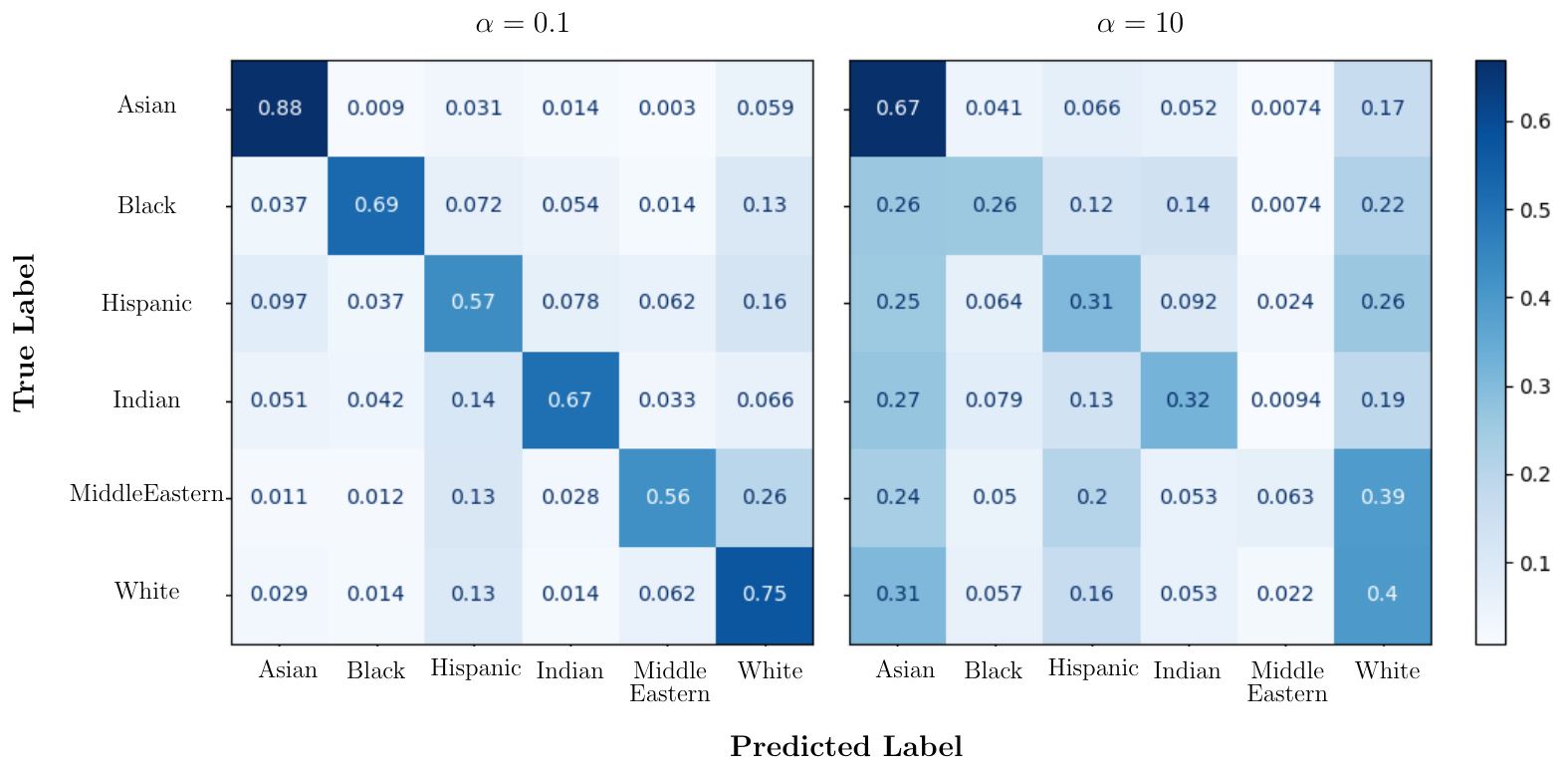}%
    \vspace{-6pt}
    \caption{Normalized confusion matrices for the FairFace dataset, considering $\mathbf{S}$ as race, with $\alpha$ values of $0.1$ and~$10$.}
    \vspace{-15pt}
    \label{Fig:NormalizedConfusionMatrices_FairFace_race}
\end{figure}


\subsubsection{Visualizing DVPF Effects on FairFace and IJB-C Data with t-SNE}

\vspace{-1pt}

\autoref{Fig:tSNE_on_IJBC_identity} presents a qualitative visualization of FR utility performance on the IJB-C dataset. For this visualization, we utilized t-distributed stochastic neighbor embedding (t-SNE) \cite{maaten2008visualizing} to project the underlying space into 2D. 
This figure illustrates visualizations for 10 randomly selected identities from the IJB-C dataset: (a) and (c) show the original (clean) embeddings from ArcFace and AdaFace, respectively, while (b) and (d) depict the obfuscated embeddings of the corresponding FR models using the DVPF (\textsf{P1}) mechanism with $\alpha=0.1$. Notably, increasing the information leakage weight $\alpha$ results in more overlapping regions among identities in this illustrative 2D visualization method.

\autoref{Fig:tSNE_on_sensitive_attribute_race} provides a qualitative visualization of the leakage in sensitive attribute classification on the FairFace database, both before and after applying the DVPF model, with $\mathbf{S}$ set as race. 
As illustrated, distinct regions associated with six racial classes (Asian, Black, Hispanic, Indian, Middle-Eastern, White) are evident in the clean embedding. However, after applying the DVPF (\textsf{P1}) mechanism with $\alpha=10$, the sensitive labels become almost uniformly distributed across the space. This distribution aligns with our interpretation of random guessing performance on the adversary's side. 
This behavior is consistent for both ArcFace and AdaFace protected embeddings, and for both gender and race as sensitive attributes. However, for brevity, we present only one example.
\autoref{Fig:NormalizedConfusionMatrices_FairFace_race} depicts the normalized confusion matrices for the FairFace dataset, obtained after applying the DVPF (\textsf{P1}) mechanism. In these matrices, $\mathbf{S}$ is considered as race, and the configuration is MS1M-RF-i50-Arc-FairFace, with $\alpha$ values set at $0.1$ and $10$. Notably, as $\alpha$ increases, the diagonal dominance in the matrices becomes less pronounced, indicating a higher probability of misclassification of the sensitive attribute.

\vspace{-3pt}

\subsection{Discussions and Future Directions}
\label{Sec:Discussions_FutureDirections}

\vspace{-1pt}

%
%

\subsubsection{Potential Contribution of $\mathsf{GenPF}$ to Bias Mitigation}
\label{SubSec:GenPF_for_BiasMitigation}

\vspace{-3pt}

The $\mathsf{GenPF}$ model may also contribute to bias mitigation through two conceptually distinct mechanisms:

\noindent
\paragraph*{a) Generation of Unbiased Synthetic Datasets for Utility Services Training and Evaluation}
Assume that the conditional generator $g_{\boldsymbol{\varphi}}$ can synthesize data of sufficient fidelity and utility conditioned on a discrete sensitive variable $\mathbf{S}$ supported on $\mathcal{S}$. Then the system designer can generate a synthetic dataset with a controlled marginal distribution over $\mathbf{S}$, including, for example, a balanced distribution over the values of $\mathcal{S}$. In the discrete case, this corresponds to choosing $P_{\mathbf{\widetilde{S}}}$ so that $\mathrm{H}(P_{\mathbf{\widetilde{S}}}) = \log_2 |\mathcal{S}|$, which yields a uniform distribution over the states of $\mathbf{S}$. This can help reduce dataset imbalance with respect to $\mathbf{S}$, although it does not by itself guarantee fairness of a downstream utility model.

\noindent
\paragraph*{b) Learning Invariant Representations with Respect to $\mathbf{S}$}
The privacy term in the $\mathsf{GenPF}$ objective encourages the learned representation $\mathbf{Z}$ to carry less information about the sensitive variable $\mathbf{S}$. In this sense, it promotes representations that are less predictive of $\mathbf{S}$, which is closely related to the objective of in-processing bias-mitigation methods that seek to reduce undesirable dependence on sensitive attributes during training. This perspective is also related to classical invariance objectives in computer vision, where representations are encouraged to be less sensitive to nuisance factors such as translation, scaling, or rotation \cite{lowe1999object}. A related example is the Fader Network \cite{lample2017fader}, in which the encoder is adversarially trained to learn feature representations that are less informative about selected facial attributes.

\subsubsection{Future Directions}
\label{SSSec:FutureDirections}

An important direction for future work is to extend the generative formulation to realistic privacy-preserving image synthesis. In the present paper, the main face-recognition validation is conducted in the embedding-based setting, while the raw-image examples serve only as proof-of-concept illustrations. A broader study should therefore evaluate high-fidelity private generation on realistic datasets.

A second direction is to combine the proposed privacy-funnel (`\textit{context-aware}') framework with prior-independent mechanisms such as differential privacy (`\textit{context-free}'). This would enable the joint study of complementary privacy protections under different threat models.

Finally, the general framework can be instantiated with alternative architectures in both the discriminative and generative components, including diffusion-based generators and transformer-based encoders.

\vspace{-4pt}

\section{Conclusion}
\label{Sec:Conclusion}

\vspace{-1pt}

In this work, we studied privacy-preserving representation learning for face recognition using the information-theoretic Privacy Funnel model. We introduced the Generative Privacy Funnel ($\mathsf{GenPF}$) and Discriminative Privacy Funnel ($\mathsf{DisPF}$) formulations, and developed the Deep Variational Privacy Funnel (DVPF) framework to make the corresponding objectives tractable in deep models. The proposed framework quantifies the privacy--utility trade-off and is compatible with recent face recognition architectures such as ArcFace and AdaFace.
Experiments with recent face recognition architectures, including ArcFace and AdaFace, on Morph and FairFace show the trade-off between utility and privacy leakage induced by the proposed framework. In particular, increasing the leakage weight \(\alpha\) reduces information leakage about sensitive attributes, but this typically comes at the cost of lower face-recognition utility, especially in high-privacy regimes. We further evaluated the trained models on the challenging IJB-C benchmark to assess generalization beyond the training distribution. A reproducible software package is also provided to facilitate further work in privacy-preserving face recognition.

\section*{Acknowledgement}

\vspace{-1pt}

This research is supported by the Swiss Center for Biometrics Research and Testing at the Idiap Research Institute. It is also conducted as part of the SAFER project, which received support from the Hasler Foundation under the Responsible AI program.

\putbib

\end{bibunit}


\appendices
\clearpage

\AppendixOnlyTOC


\begin{bibunit}




\counterwithin{table}{section}
\counterwithin{figure}{section}

\renewcommand{\thesubsection}{A.\arabic{subsection}}

\pagebreak
\appsection{Navigating the Data Privacy Paradigm}
\label{AppxSec:Navigating_DataPrivacy_Paradigm}

\vspace{10pt}

The domain of data privacy is evolving at a fast pace, especially because personal and sensitive data is increasingly being generated and shared through digital channels. Data privacy refers to guidelines and rules governing the collection, use, storage, and sharing of personal and sensitive data, with the aim of safeguarding such data against exposure, unauthorized access, or misuse. Data privacy employs various measures, such as encryption, access control, as well as privacy-enhancing technologies (PETs), in order to prevent unauthorized access to personal and sensitive data and minimize unnecessary sharing of such data.

One of the key challenges in data privacy is managing the balance between protecting personal and sensitive information and enabling its use for legitimate purposes. This trade-off becomes especially difficult in light of rapid technological change and the growing demand for data-driven services. Another challenge is the lack of harmonized global standards and regulations for protecting personal and sensitive information. Although many countries have established their own data privacy laws, significant variation across these legal frameworks complicates the consistent protection of personal and sensitive data across borders. Despite these challenges, the field of data privacy continues to develop through new technologies and approaches aimed at improving the protection of personal and sensitive information.

A central challenge in the era of big data is balancing the use of data-driven machine learning algorithms with the protection of individual privacy. The increasing volume of data collected and used to train machine learning models raises concerns about misuse, re-identification, and other privacy risks. This situation presents several open problems, including how to de-identify or anonymize data effectively so as to reduce the risk of identifying individuals in training data, and how to develop reliable methods for safeguarding personal information. Furthermore, there is a pressing need to establish ethical and regulatory frameworks for data use in machine learning that protect individuals' rights.

\vspace{-5pt}

\appsubsection{Lunch with Turing and Shannon}

Alan Turing visited Bell Labs in 1943, during the peak of World War II, to examine the X-system, a secret voice scrambler for private telephone communications between the authorities in London and Washington\footnote{This section is inspired by the insightful work of \cite{calmon2015thesis, hsu2021survey} and adapted from \cite{razeghi2023thesisCLUB}.
}. While there, he met Claude Shannon, who was also working on cryptography. 
In a 28 July 1982 interview with Robert Price in Winchester, MA \cite{price1982claude}, Shannon reminisced about their regular lunch meetings where they discussed computing machines and the human brain instead of cryptography \cite{guizzo2003essential}.
Shannon shared with Turing his ideas for what would eventually become known as information theory, but according to Shannon, Turing did not believe these ideas were heading in the right direction and provided negative feedback. Despite this, Shannon's ideas went on to be influential in the development of information theory, which has had a significant impact on the fields of computer science and telecommunications.

Protecting information from unauthorized access has been a central concern in the fields of information theory and computer science since their early development. The interaction between Shannon and Turing foreshadows some of the different approaches that later emerged in the two communities for addressing the problem of preventing unauthorized access to information contained in disclosed data. These approaches often involve different models and distinct mathematical techniques. It is important to note that these approaches have evolved over time as technology and threats to privacy have changed, and they continue to be active areas of research and development in both fields.

In the 1970s, two influential papers on secrecy appeared, and they made clear how differently information theory and computer science were approaching the problem. One of them, written by Aaron Wyner at Bell Labs, introduced the wiretap channel: a setting in which data is sent over a channel that can also be observed by an eavesdropper through a second, noisier channel. Wyner showed that, under suitable conditions, one can design codes so that the intended receiver can decode the message while the eavesdropper learns essentially nothing from what they observe. This line of work does not rest on assumptions about what the eavesdropper can or cannot compute, and it later became foundational in information-theoretic secrecy.

In November 1976, Diffie and Hellman published a paper that introduced the concept of public-key cryptography and described how it could be used to achieve secure communication without the need for a shared secret key \cite{hellman1976new}. This approach to cryptography relies on computational assumptions: its security depends on the practical difficulty of recovering private information without access to the private key. As a result, public-key cryptographic systems made key distribution much more practical than approaches that rely on information-theoretic secrecy, which do not make assumptions about an adversary’s computational capabilities. The paper also discussed public-key distribution systems and verifiable digital signatures, both of which became central tools in modern cryptography.

After the publication of these works in the 1970s, public key cryptography, which assumes that adversaries are computationally constrained, became mainstream. Many applications ranging from banking to health care and public services use public key cryptography. It is estimated that public key cryptography is used billions of times a day in systems ranging from digital rights management to cryptocurrencies. Information-theoretic approaches to secrecy, on the other hand, seek security without making assumptions about the computational power of adversaries, but they typically require stronger assumptions on the communication setting or system model. This leads to a class of security schemes with very strong guarantees, under more constrained assumptions, resulting in a mathematically elegant theory whose practical deployment is often limited.

The intersection of information theory and computer science approaches to privacy continues to be relevant in today's world, where the collection of individual-level data has increased significantly. This development has brought both challenges and opportunities for both fields, as the widespread collection of data has brought significant economic benefits, such as personalized services and innovative business models, but also poses new privacy threats. For example, social media posts may be used for undesirable political targeting \cite{effing2011social, o2018social}, machine learning models may reveal sensitive information about the data used for training \cite{abadi2016deep}, and public databases may be deanonymized with only a few queries \cite{narayanan2008robust, su2017anonymizing}. Both fields have faced new challenges and opportunities in addressing these issues.

\vspace{-10pt}

\appsubsection{Identification, Quantification, and Mitigation of Privacy Risks}

Protecting privacy requires attention at every stage of personal data handling, including \textbf{(i)} \textit{\textbf{collection}}, \textbf{(ii)} \textit{\textbf{storage}}, \textbf{(iii)} \textit{\textbf{processing}}, and \textbf{(iv)} \textit{\textbf{sharing} (dissemination)}. Taking all of these stages into account makes it possible to think about privacy in a more complete way, across settings that range from traditional data management to more advanced machine learning systems. Research on privacy risk management is often organized around three basic questions: how privacy risks can be \textit{identified}, how they can be \textit{measured}, and how they can be \textit{mitigated}.


\begin{itemize}[leftmargin=1.9em]
    \item[\textbf{(a)}]
    \textbf{Identification:} How can we identify the risks of data leakage and potential privacy attacks across the entire data lifecycle, from collection through to processing and sharing?
    \item[\textbf{(b)}] 
    \textbf{Quantification:} Following the identification of privacy risks, what \textit{metrics}\footnote{In this document, `metric' is used not in the traditional mathematical sense of a distance function, but as a quantifier for assessing privacy risk.} can be developed and applied to precisely quantify these risks and monitor the effectiveness of implemented privacy protection strategies?
    \item[\textbf{(c)}] 
    \textbf{Mitigation:} Given an understanding of the identified privacy risks, what strategies can be formulated and implemented to mitigate those risks, while ensuring an appropriate balance between operational objectives and privacy, in line with legal and ethical standards?
\end{itemize}

\vspace{5pt}

The following discussion will provide a brief exploration of these pivotal questions.

\vspace{10pt}
 
\subsubsection{Identification of Privacy Risks}

\vspace{-2pt}

Identifying and understanding privacy risks is a critical first step in safeguarding privacy across the entire data lifecycle, including \textit{collection}, \textit{storage}, \textit{processing}, and \textit{dissemination} \cite{solove2002conceptualizing, solove2005taxonomy}. This task becomes increasingly vital and, at times, complex within the context of both traditional data management practices and the utilization of machine learning algorithms \cite{solove2010understanding, solove2024artificial}. The identification process requires a detailed understanding of potential vulnerabilities that could lead to data leakage and privacy attacks, alongside the development of systematic approaches to \textit{detect} and \textit{assess} these risks \cite{solove2010understanding, smith2011information, orekondy2017towards, milne2017information, beigi2020survey}. We briefly explore several key methodologies that are essential for the comprehensive identification of privacy risks in these areas.

\vspace{3pt}

\paragraph{Data Sensitivity Analysis}
Identifying privacy risks inherent in various types of data is a significant challenge in conventional database systems as well as in big data analytics. This requires a careful examination of the data to identify personally identifiable information or sensitive personal information. Using attribute-based risk assessment and principles of privacy-preserving data mining, an organization can identify the sensitive data elements that need protection. Identifying these privacy risks is essential for privacy risk management and is the first step toward protecting privacy-sensitive data.

\vspace{4pt}

\paragraph{Vulnerability Assessment Across Data Lifecycle}
Protecting privacy requires careful examination of weaknesses that could lead to breaches. This means looking at every stage of the data lifecycle, from collection to storage, processing, and dissemination. In machine learning contexts, this requires careful examination of model design as well as data processing procedures in order to identify points at which data might leak. Tools that automatically check for privacy risks can greatly assist these assessments, helping to identify and address problems before they result in privacy harms.

\vspace{4pt}

\paragraph{Simulated Privacy Attack Scenarios}
There is a growing body of studies that simulate privacy attacks to identify potential vulnerabilities in privacy-preserving measures for general data processing systems and ML models. In this context, these studies propose attacks using adversarial modeling and synthetic data to examine how easily a model can be attacked or a data record can be re-identified. Such attacks are becoming more common in the ML context and are particularly associated with model inversion attacks and membership inference attacks. These simulated attacks help evaluate the effectiveness of adopted privacy-preserving measures and strengthen them by identifying weaknesses that require countermeasures.

\vspace{10pt}

\subsubsection{Quantification of Privacy Risks}

Following the identification of privacy risks and the determination of applicable privacy regulations and standards, the next step is to establish and apply metrics. In this step, the identified privacy risks have to be quantified and the progress of their mitigation has to be monitored.
Today, data is processed in highly diverse applications, so determining the privacy risk of data processing in these applications requires metrics that measure the risk at various points in the data life cycle, such as collection, storage, processing, and dissemination of data. In addition, it is important to consider the applicability and relevance of the privacy metrics based on the stage of the data life cycle and the application context \cite{duchi2013local, duchi2014privacy, mendes2017privacy, duchi2018minimax, wagner2018technical, bhowmick2018protection, liao2019tunable, hsu2020obfuscation, bloch2021overview, saeidian2021quantifying}. Thus, knowing the operational interpretation of the privacy metrics \cite{issa2019operational, kurri2023operational} is important.
Metrics applied to the personal data processed by each data processing system serve as an indicator of the privacy level of such systems, which enables data controllers to better manage their privacy risks. We discuss one such metric in Sec.~\ref{AppxSec:Preliminaries}.

\vspace{10pt}

\subsubsection{Mitigation of Privacy Risks}

The data privacy risk cannot be mitigated with a single control and therefore requires a multi-faceted approach based on a range of techniques and methodologies. A subset of these techniques is known as Privacy-Enhancing Technologies (PETs) and deals with the privacy protection of data at all stages in its life cycle.
PETs are privacy-protecting tools and techniques that directly address privacy threats affecting personal data over its whole lifecycle, i.e., collection, storage, processing, and transmission. In short, PETs aim to achieve privacy by design.
Simple PETs include pseudonymization \cite{chaum1981untraceable, chaum1985security}, anonymization \cite{sweeney2000simple, sweeney2002k}, and encryption \cite{shannon1949communication, diffie1976new, hellman1977extension}. They deal with privacy issues directly by helping ensure that sensitive data, especially personal data, is kept confidential, cannot be readily identified, and cannot be modified without authorization.
We review PETs in Sec.~\ref{ssec:PETs_and_Metrics_PriorDependentIndependent}.

\vspace{-5pt}

%
%
\appsubsection{Privacy-Enhancing Technologies}
\label{ssec:PETs_and_Metrics_PriorDependentIndependent}

\vspace{-2pt}

PETs protect personal privacy directly by tackling privacy threats. As attackers continually improve their attack methods, the need for PETs to protect personal data from unauthorized access and use remains constant. There is a wide range of PETs that cover many aspects of privacy and data protection.

\vspace{5pt}

\subsubsection{Encryption, Anonymization, Obfuscation, and Information-Theoretic Technologies}

Cryptographic techniques for modern PETs have evolved to deal with the challenge of securing the data we store (at rest), send (in transit), or use (in use). Examples of such techniques include symmetric and asymmetric encryption, as well as homomorphic encryption. Data pseudonymization and anonymization are other relevant techniques. They are employed to transform sensitive data so that identifying attributes are no longer directly visible, in such a way that it becomes very hard to link the anonymized data to the corresponding individuals. Finally, differential privacy provides a probabilistic form of protection for sensitive information in datasets and outputs, introduced via carefully computed and tuned noise additions to the dataset, to the output of a query or analytics, or even to the model in the context of machine learning, so as to limit what can be inferred about any individual’s personal data from statistics and/or ML models.
So, in addition to those we have already mentioned, there are information-theoretic privacy approaches that give a more fundamental perspective on data protection, based on the information that an attacker can potentially gain from a dataset, regardless of the attacker’s computational resources. In fact, the information-theoretic approach analyzes the information leakage from the data, and thus the uncertainty associated with what can be inferred from it, with the purpose of establishing a bound on the information that can be derived from the data, and therefore guaranteeing a level of privacy without relying on assumptions about the computational resources of the attacker.
In Sec.~\ref{ssec:PriorDependent_PriorIndependent_Mechanisms}, we review these techniques from the standpoint of the prior knowledge we have regarding the data distribution.

\vspace{5pt}

\subsubsection{Privacy-Preserving Computation Technologies}

Secure computation techniques are essential for maintaining privacy during data processing \cite{yao1982protocols, micali1992secure, mohassel2018aby3, juvekar2018gazelle, keller2020mp, knott2021crypten, neel2021descent}. Confidential computing \cite{mohassel2018aby3, mo2022sok, vaswani2023confidential}, which employs Trusted Execution Environments (TEEs) \cite{sabt2015trusted}, is an important tool, isolating computation to protect data in use from both internal and external threats. Additionally, Secure Multi-party Computation (SMPC) \cite{goldreich1998secure, du2001secure, cramer2015secure, knott2021crypten} facilitates collaborative computation over data distributed among multiple parties without revealing the data itself, enabling joint data analysis or model training while preserving the privacy of each party's data. Zero-Knowledge Proofs (ZKPs) \cite{fiege1987zero, kilian1992note, goldreich1994definitions} offer another layer of security, allowing one party to prove the truth of a statement to another party without revealing any information beyond the validity of the statement itself, essential for scenarios requiring validation of data authenticity or integrity without exposing the data.

\vspace{5pt}

\subsubsection{Decentralized Privacy Technologies}

Decentralized privacy-preserving technologies, which support \textit{collaborative} and/or \textit{federated} data analysis and model building using distributed data, have drawn considerable interest in recent years from a variety of disciplines \cite{shokri2015privacy, mcmahan2017communication, dwivedi2019decentralized, wei2020federated, kaissis2020secure, kairouz2021advances, shiri2023multi}. They enable the training of machine learning models on decentralized data while helping to protect privacy. Of these technologies, federated learning has become particularly popular, as it enables the training of machine learning models across distributed devices or servers. In contrast to conventional solutions that transfer raw data to a central server for analysis, federated learning methods transfer model updates computed locally from the decentralized data. These updates then contribute to the overall model while reducing the need to centralize sensitive data.

\vspace{-3pt}

\appsubsection{Prior-Dependent vs. Prior-Independent Mechanisms in PETs}
\label{ssec:PriorDependent_PriorIndependent_Mechanisms}

There are two main types of privacy-enhancing mechanisms: `\textit{\textbf{prior-independent}}' and `\textit{\textbf{prior-dependent}}' \cite{hsu2021survey, razeghi2023thesisCLUB}. Prior-independent mechanisms make minimal assumptions about the data distribution and the information held by an adversary and are designed to protect privacy regardless of the specific characteristics of the data being protected or the motivations and capabilities of any potential adversaries. Prior-dependent mechanisms, on the other hand, make use of knowledge about the probability distribution of private data and the abilities of adversaries in order to design privacy-preserving mechanisms. These mechanisms may be more effective in certain scenarios where the characteristics of the data and the adversary are known or can be reasonably estimated but may be less robust in situations where such information is uncertain or changes over time.

Data anonymization \cite{sweeney2000simple} techniques, such as $k$-anonymity \cite{sweeney2002k}, $\ell$-diversity \cite{machanavajjhala2006diversity}, $t$-closeness \cite{li2007t}, differential privacy (DP) \cite{dwork2006calibrating}, and pufferfish \cite{kifer2012rigorous}, aim to preserve the privacy of data through various forms of data perturbation. These techniques focus on queries, inference algorithms, and probability measures, with DP being the most popular context-free privacy notion based on the distinguishability of ``\textit{neighboring}'' databases. However, DP does not provide any guarantee on the average or maximum information leakage \cite{du2012privacy}, and pufferfish, while able to capture data correlation, does not prioritize preserving data utility.

DP is a privacy metric that measures the impact of small perturbations at the input of a privacy mechanism on the probability distribution of the output. A mechanism is said to be $\epsilon$-differentially private if the probability of any output event does not change by more than a multiplicative factor $e^{\epsilon}$ for any two neighboring inputs, where the definition of ``neighboring'' inputs depends on the chosen metric of the input space. DP is prior-independent and often used in statistical queries to ensure the result remains approximately the same regardless of whether an individual's record is included in the dataset. The privacy guarantee of DP can typically be achieved through the use of additive noise mechanisms, such as adding a small perturbation or random noise from a Gaussian, Laplacian, or exponential distribution \cite{dwork2014algorithmic}.

Since its introduction, DP has been extended in several ways. These include approximate differential privacy, which introduces a small additional parameter $\delta$ \cite{dwork2006our}; local differential privacy, which requires the privacy guarantee to hold for every pair of possible input values of an individual \cite{duchi2013local_minimax}; and R{\'e}nyi differential privacy, which uses R{\'e}nyi divergence to measure the difference between output distributions induced by neighboring inputs \cite{mironov2017renyi}. DP has two key properties that make it especially useful for privacy protection: \textbf{(i)} it is composable \cite{dwork2014algorithmic, abadi2016deep}, meaning that the privacy loss from multiple applications of DP mechanisms can be tracked and bounded in a controlled way; and \textbf{(ii)} it is robust to post-processing \cite{dwork2014algorithmic}, meaning that further processing of the output cannot weaken the privacy guarantee. Together, these properties support the modular design and analysis of privacy mechanisms under a specified privacy budget.


Information-theoretic (IT) privacy is the study of designing mechanisms and metrics that preserve privacy when the statistical properties or probability distribution of data can be estimated or partially known. IT privacy approaches \cite{reed1973information,yamamoto1983source, evfimievski2003limiting, rebollo2009t, du2012privacy, sankar2013utility, calmon2013bounds, makhdoumi2013privacy, asoodeh2014notes, calmon2015fundamental, salamatian2015managing, basciftci2016privacy, asoodeh2016information, kalantari2017information, rassouli2018latent, asoodeh2018estimation, rassouli2018perfect, liao2018privacy, osia2018deep, tripathy2019privacy, Hsu2019watchdogs, liao2019tunable,  sreekumar2019optimal, xiao2019maximal, diaz2019robustness, rassouli2019data, rassouli2019optimal, razeghi2020perfectobfuscation, zarrabian2023lift, zamani2023privacy, saeidian2023pointwise} model and analyze the trade-off between privacy and utility using IT metrics, which quantify how much information an adversary can gain about private features of data from access to disclosed data. These metrics are often formulated in terms of divergences between probability distributions, such as f-divergences and R{\'e}nyi divergence. IT privacy metrics can be operationalized in terms of an adversary's ability to infer sensitive data and can be used to balance the trade-off between allowing useful information to be drawn from disclosed data and preserving privacy. By using prior knowledge about the statistical properties of data and assumptions about the adversary's inference capabilities, IT privacy can help to understand the fundamental limits of privacy and how to balance privacy and utility.

The IT privacy framework is based on the presence of a private variable and a correlated non-private variable, and the goal is to design a privacy-assuring mapping that transforms these variables into a new representation that achieves a specific target utility while minimizing the information inferred about the private variable. IT privacy approaches provide a context-aware notion of privacy that can explicitly model the capabilities of data users and adversaries, but they require statistical knowledge of data, also known as priors. This framework is inspired by Shannon's information-theoretic notion of secrecy \cite{shannon1949communication}, where security is measured through the equivocation rate at the eavesdropper\footnote{A secret listener (wiretapper) to private conversations.}, and by Reed \cite{reed1973information} and Yamamoto's \cite{yamamoto1983source} treatment of security and privacy from a lossy source coding standpoint.

\appsubsection{Challenges in Data-Driven Privacy Preservation Mechanisms}

\vspace{-2pt}

Cryptography is a time-honored field that provides a wide range of tools for securing information. However, in today's data-driven economy, traditional cryptographic solutions are often not sufficient to protect privacy. The main difficulty is that disclosed data can still be observed and analyzed by an adversary. In many scenarios, such as when a statistician queries a database containing sensitive information, it is not sufficient to simply encrypt the output. As illustrated by the release of population statistics by the U.S. Census Bureau, significant privacy losses can accumulate over multiple queries, allowing an adversary to infer sensitive information \cite{machanavajjhala2008privacy}. A similar issue arises in machine learning, where user data are needed to train a model: data disclosure can improve model utility, but it can also create risks to the privacy of the individuals from whom the data were obtained. In particular, an adversary may extract information about individual records by analyzing the model's outputs.

The main goal in data release problems is not to prevent all information leakage, which is practically impossible. Instead, the goal is to achieve a level of privacy that is balanced against utility. The privacy threat model for data release includes both computationally bounded and information-theoretic adversaries that attempt to extract information about a dataset and, possibly, about an individual it includes. By analyzing the released data, they may infer sensitive information such as political preferences or whether a particular individual is included in the dataset.

Recent privacy mechanisms have been influenced by advances in computer science and information theory that relax strong assumptions about an adversary's computational capabilities. These mechanisms differ in their adversary goals (e.g. probability of correctly guessing a value versus minimizing the mean-squared error of a reconstructed value) and in their characterization of private information. A major challenge is to balance application-specific utility against privacy needs.

Building on the emergence of data-driven privacy approaches, recent studies have explored privacy mechanisms inspired by Generative Adversarial Networks (GANs). These methods formulate privacy protection as a strategic game between the defender (or privatizer) and the adversary. The goal of the privatizer is to censor or encode a dataset such that the released data limits inference leakage about sensitive variables. On the other hand, the adversary seeks to recover information about private variables from the released data. This interplay between optimizing privacy and maintaining data utility through adversarial training—whether deterministic or stochastic—is a central theme of these approaches.

Machine learning is becoming increasingly prevalent, meaning that reliable data-driven privacy methods are essential for protecting privacy, gaining public trust, and minimizing damage in the event of a data breach. Such breaches can have serious and lasting consequences for individuals and organisations alike, resulting in damage to reputation and financial loss. The need for powerful privacy-preserving methods is becoming increasingly important as we move to a more data-centric world and as machine learning becomes more pervasive in daily life.

\vspace{-5pt}

\appsubsection{Threats to PETs}

In this subsection, we briefly discuss the main threats faced by privacy-enhancing technologies (PETs). In particular, we consider attacks that aim to weaken the privacy or security guarantees provided by PETs and review the main objectives such adversaries may pursue.

\vspace{2pt}

\subsubsection{Adversary Objectives}

\vspace{1pt}

As a high-level taxonomy, we group adversarial objectives into three categories: \textbf{(i)} data reconstruction, \textbf{(ii)} unauthorized access, and \textbf{(iii)} user re-identification.

\vspace{1pt}

\paragraph{Data Reconstruction}
The objective here is to recover original data, or sensitive information about it, from its protected, transformed, or encoded form \cite{agrawal2000privacy, rebollo2009t, sankar2013utility, asoodeh2016information, dwork2017exposed, bhowmick2018protection, ferdowsi2020privacy, stock2022defending, razeghi2023bottlenecks, shiri2024primis}. This objective may take two forms. The first is \textit{attribute inference}, where the adversary seeks to recover specific sensitive attributes or features from the protected data. The second is \textit{full reconstruction}, where the adversary aims to recover the original data record, either exactly or approximately, from the protected representation. Both cases indicate leakage of sensitive information and therefore weaken the privacy guarantees of the protection mechanism.

\vspace{1pt}

\paragraph{Unauthorized Access} 
The objective here is to gain access to protected systems, services, or data without authorization \cite{dunne1994deterring, campbell2003economic, winn2007guilty, mohammed2012analysis, muslukhov2013know, sloan2017unauthorized, razeghi2018privacy, maithili2018analyzing, prokofiev2018method, wang2019longitudinal}. In the context of PETs, this may include bypassing authentication mechanisms, accessing protected records, or exploiting weaknesses in the protection pipeline to obtain privileges or information that should remain inaccessible. The central issue is that the adversary succeeds in circumventing the intended access-control or protection mechanism.

\vspace{1pt}

\paragraph{User Re-identification}
The objective in user re-identification is to link anonymized, pseudonymized, or partially protected data back to a specific individual \cite{el2011systematic, layne2012person, zheng2015scalable, henriksen2016re, zheng2016person, ye2021deep}. This is typically done by combining the protected data with auxiliary information or by linking records across datasets. Even when direct identifiers have been removed, such linkage can reveal the identity of the individual or enable tracking of that individual across records or over time. Re-identification attacks therefore challenge the effectiveness of anonymization and related privacy-preserving mechanisms.

\vspace{1pt}

\subsubsection{Adversary Knowledge}

\vspace{2pt}

\paragraph{Knowledge of the Learning Model}

The adversary may know details of the model used by the system, including its architecture, parameters, training procedure, and implementation choices \cite{wang2018stealing, song2019privacy, oseni2021security, bober2023architectural, yang2023comprehensive}. This may include knowledge of the layer structure, activation functions, loss function, optimization method, and training hyperparameters. Such information can be used to design attacks that target the model more effectively, for example by exploiting known failure modes or by approximating its decision behavior.

\vspace{3pt}

\paragraph{Knowledge of the System Workflow}
The adversary may also know how the overall system operates, including its architecture, data flow, decision pipeline, and validation procedures. This type of knowledge can reveal points at which the system is susceptible to manipulation or information leakage. For example, knowledge of preprocessing steps, intermediate interfaces, or decision thresholds may help the adversary construct more effective attack inputs or identify stages at which the system is most vulnerable.

\vspace{1pt}

\paragraph{Knowledge of the Data}
The adversary may have information about the data used by the system, including data sources, preprocessing steps, feature distributions, class imbalance, and outliers. Such knowledge can support attacks that exploit regularities in the data distribution or weaknesses in data handling. Even partial access to the data, or to representative samples from the same distribution, may help the adversary approximate important properties of the underlying dataset.

\vspace{1pt}

\paragraph{Knowledge of Security Mechanisms}
The adversary may know the security mechanisms used by the system, including authentication procedures, encryption methods, access-control rules, and related protocols. This knowledge can help identify weaknesses in the protection pipeline and support attacks against specific security components or interfaces.

\vspace{1pt}

\paragraph{Insider Operational Knowledge}
The adversary may possess insider knowledge acquired through legitimate access or prior observation of the system. This may include knowledge of internal procedures, deployment practices, access patterns, and system configuration. Such information can reduce uncertainty about how the system is implemented and operated, thereby enabling more targeted attacks.

\vspace{1pt}

\subsubsection{Adversary Strategy}

\vspace{1pt}

Adversaries may employ a range of strategies to weaken the privacy or security guarantees of machine learning systems and privacy-enhancing technologies. These strategies differ in the type of access available to the adversary, the information being exploited, and the attack objective. In the context of machine learning and artificial intelligence, several attack strategies are particularly relevant. Below, we briefly review a few representative examples.

\vspace{1pt}

\paragraph{Gradient-Based Attacks}
Gradient-based attacks exploit gradient information, either directly or indirectly, to analyze or manipulate machine learning models \cite{liu2016delving, papernot2017practical, ilyas2018black, bhagoji2018practical, porkodi2018survey, alzantot2019genattack, guo2019simple, sablayrolles2019white, rahmati2020geoda, tashiro2020diversity}. In the white-box setting, the adversary has access to model parameters or gradients and can use this information to construct targeted attacks, analyze decision boundaries, or infer properties of the training process. In the black-box setting, direct access to the model internals is unavailable, and the adversary instead relies on repeated queries and observable outputs to estimate gradients or approximate the model behavior. These strategies are relevant to attacks such as model extraction and membership inference \cite{tramer2016stealing, batina2019csi, chandrasekaran2020exploring, shokri2017membership}.

\vspace{1pt}

\paragraph{Temporal Analysis Attacks}
Temporal pattern analysis exploits information contained in the time-dependent behavior of a system \cite{kamat2009temporal, xiao2015protecting, backes2016privacy, grover2017digital, leong2020privacy, qi2020privacy, zhang2021synteg, li2023prism}. By analyzing outputs, updates, or verification activity over time, an adversary may identify recurring patterns, update schedules, or periods in which the system is more vulnerable. Such information can then be used to time attacks more effectively or to infer aspects of the system that are not apparent from a single interaction.

\vspace{1pt}

\paragraph{Multi-Source and Data-Poisoning Attacks}
Adversaries may also combine information from multiple sources or manipulate the data used by the system. A prominent example is the data-poisoning attack \cite{biggio2012poisoning, guo2020practical, tian2022comprehensive, wang2022threats, ramirez2022poisoning, carlini2023poisoning}, in which corrupted, misleading, or intentionally mislabeled samples are inserted into the training set in order to alter the learned model. Such attacks can degrade model performance, introduce bias, or induce targeted failure modes. In addition, adversaries may combine observations from multiple modalities or external data sources to support reconstruction, linkage, or impersonation attacks. Related techniques, including multi-modal synthesis \cite{abdullakutty2021review, liu2021face, hu2022m} and denoising-based recovery \cite{voloshynovskiy2000generalized, voloshynovskiy2001attack, lu2002denoising, kloukiniotis2022countering, chen2023advdiffuser}, can further strengthen reconstruction or evasion attacks in some settings.

\vspace{-7pt}

\appsubsection{Biometric PETs}

\vspace{-2pt}

Biometric recognition is an automated process based on certain characteristics of a person, such as behavioral and physiological traits. Systems based on such human features are called biometric recognition systems. Each system includes four basic subsystems: (i) data capture, (ii) signal processing and feature extraction, (iii) comparison, and (iv) data storage. 
Face recognition technology, however, poses serious security and privacy concerns because face images may be reconstructed from stored templates (embeddings).

Recently, a variety of Biometric Privacy-Enhancing Technologies (B-PETs) have emerged to protect privacy-sensitive information contained in biometric templates. This can be achieved through template protection techniques and/or methods that reduce the exposure of sensitive attributes such as age, gender, and ethnicity in biometric data.

The ISO/IEC~24745 standard \cite{ISO24745} sets forth four primary requirements for each biometric template protection scheme, encompassing the principles of \textit{cancelability}, \textit{unlinkability}, \textit{irreversibility}, and the \textit{preservation of recognition performance}. 
These biometric template protection schemes can be categorized into two main groups: \textbf{(i)} cancelable biometrics, which encompasses techniques like Bio-Hashing \cite{jin2004biohashing}, MLP-Hash \cite{shahreza2023mlp}, IoM-Hashing \cite{jin2017ranking}, among others, and rely on transformation functions dependent on keys to generate protected templates \cite{nandakumar2015biometric, sandhya2017biometric, rathgeb2022deep}, and \textbf{(ii)} biometric cryptosystems, which include methodologies such as fuzzy commitment \cite{juels1999fuzzy} and fuzzy vault \cite{juels2006fuzzy}, either binding keys to biometric templates or generating keys from these templates \cite{uludag2004biometric, rathgeb2022deep}. Additionally, some researchers have explored the application of Homomorphic Encryption for template protection in face recognition systems \cite{boddeti2018secure, bassit2021fast, ijcb2022hybrid}.

Face recognition systems, as extensively discussed in prior research \cite{biggio2015adversarial, galbally2010vulnerability, marcel2023handbook}, are not only susceptible to security threats but also face privacy vulnerabilities. These systems rely on facial templates extracted from face images, which inherently contain sensitive information about the individuals they represent. 
%
The B-PETs predominantly focus on protecting identity-related information within face templates through the utilization of template protection schemes \cite{Razeghi2017wifs, boddeti2018secure, Razeghi2019icip, mai2020secureface, hahn2022biometric,  ijcb2022hybrid,  tifs2023measuring, abdullahi2024biometric}, or on minimizing the inclusion of privacy-sensitive attributes, such as age, gender, ethnicity, among others, in these templates \cite{morales2020sensitivenets, melzi2023multi}.
Recent studies have even demonstrated an adversary's capability to reconstruct face images from templates stored within a face recognition system's database \cite{tpami2023faceti3d, neurips2023faceti}.

\vspace{-4pt}

\appsubsection{Related Works}

\vspace{-2pt}

To address the most closely related works to ours, we consider two categories of research, which, while seemingly distinct, are indeed related. The first category encompasses research papers studying and analyzing the privacy funnel model, and the second comprises works addressing disentangled representation learning.

Considering the Markov chain $\mathbf{S} \markov \mathbf{X} \markov \mathbf{Z}$, the authors in \cite{hsu2020obfuscation, de2022funck, huang2024efficient} tackle the privacy funnel problem. In \cite{hsu2020obfuscation}, the authors introduce a method to enhance privacy in datasets by identifying and obfuscating features that leak sensitive information. They propose a framework for detecting these information-leaking features using information density estimation, where features with information densities exceeding a predefined threshold are considered risky and are subsequently obfuscated. This process is data-driven, utilizing a new estimator known as the trimmed information density estimator (TIDE) for practical implementation.

In \cite{de2022funck}, the authors present the conditional privacy funnel with side-information (CPFSI) framework. This framework extends the privacy funnel method by incorporating additional side information to optimize the trade-off between data compression and maintaining informativeness for a downstream task. The goal is to learn invariant representations in machine learning, with a focus on fairness and privacy in both fully and semi-supervised settings. Through empirical analysis, it is demonstrated that CPFSI can learn fairer representations with minimal labels and effectively reduce information leakage about sensitive attributes.

More recently, \cite{huang2024efficient} proposes an efficient solver for the privacy funnel problem by exploiting its difference-of-convex structure, resulting in a solver with a closed-form update equation. For cases of known distribution, this solver is proven to converge to local stationary points and empirically surpasses current state-of-the-art methods in delineating the privacy-utility trade-off. For unknown distribution cases, where only empirical samples are accessible, the effectiveness of the proposed solver is demonstrated through experiments on MNIST and Fashion-MNIST datasets.

The closest work to ours in face recognition is \cite{morales2020sensitivenets}, where the authors presented a privacy-preserving feature representation learning approach that suppresses sensitive information such as gender or ethnicity in the learned representations while maintaining data utility. The core idea was to reformulate the learning objective with an adversarial regularizer to remove sensitive information.

Besides, many other fundamental related works, such as \cite{tran2017disentangled, gong2020jointly, park2021learning, li2022discover, suwala2024face}, focus on learning disentangled representations and improving algorithmic fairness in face recognition systems. These works propose methods to mitigate bias, improve pose-invariant face recognition, and learn representations in which different types of information are separated so as to reduce discriminatory effects in AI systems.

In \cite{tran2017disentangled}, the authors introduce the disentangled representation learning generative adversarial network (DR-GAN) to address the challenge of pose variation in face recognition. Unlike conventional methods that either generate a frontal face from a non-frontal image or learn pose-invariant features, DR-GAN performs both tasks jointly through an encoder-decoder generator structure. This enables it to synthesize identity-preserving faces with arbitrary poses while learning a discriminative representation. The approach disentangles identity representation from other variations, such as pose, using a pose code for the decoder and pose estimation in the discriminator. DR-GAN can process multiple images per subject, fusing them into a single, robust representation and synthesizing faces in specified poses.

\vspace{1pt}

In \cite{gong2020jointly}, the authors present an approach to mitigating bias in automated face recognition and demographic attribute estimation algorithms, focusing on addressing the observed performance disparities across different demographic groups. They propose a de-biasing adversarial network, DebFace, which employs adversarial learning to extract disentangled feature representations for identity and demographic attributes (gender, age, and race) in a way that minimizes bias by reducing the correlation among these feature factors. Their approach combines demographic with identity features to enhance the robustness and accuracy of face representation across diverse demographic groups. The network comprises an identity classifier and three demographic classifiers, trained adversarially to ensure feature disentanglement and reduce demographic bias in both face recognition and demographic estimation tasks.

In \cite{park2021learning}, the authors introduce a fairness-aware disentangling variational auto-encoder (FD-VAE) that aims to mitigate discriminatory results in AI systems related to protected attributes such as gender and age, without sacrificing beneficial information for target tasks. The FD-VAE model achieves this by disentangling data representation into three subspaces: target attribute latent (TAL), protected attribute latent (PAL), and mutual attribute latent (MAL), each designed to contain specific types of information. A decorrelation loss is proposed to appropriately align information within these subspaces, focusing on preserving useful information for the target tasks while excluding protected attribute information.

\vspace{1pt}

In \cite{li2022discover}, the authors introduce Debiasing Alternate Networks (DebiAN) to mitigate biases in deep image classifiers without the need for labels of protected attributes, aiming to overcome the limitations of previous methods that require full supervision. DebiAN consists of two networks, a discoverer and a classifier, trained in an alternating manner to identify and unlearn multiple unknown biases simultaneously. This approach not only addresses the challenges of identifying biases without annotations but also excels in mitigating them effectively. The effectiveness of DebiAN is demonstrated through experiments on both synthetic datasets, such as the multi-color MNIST, and real-world datasets, showing its capability to discover and improve bias mitigation.

\vspace{1pt}

Recently, \cite{suwala2024face} introduces PluGeN4Faces, a plugin for StyleGAN designed to manipulate facial attributes such as expression, hairstyle, pose, and age in images while preserving the person's identity. It employs a contrastive loss to closely cluster images of the same individual in latent space, ensuring that changes to attributes do not affect other characteristics, such as identity.

\vspace{1pt}

In comparison to the research mentioned above, our work begins with a purely information-theoretic formulation of the PF model, which we have named the discriminative PF framework. We then extend the concept of the discriminative PF model to develop a generative PF framework. Building upon our objectives for PF frameworks, as grounded in Shannon's mutual information, we present a tractable variational approximation for both our information utility and information leakage quantities. The variational approximation objectives we have obtained share some connections with the aforementioned research, thereby bridging the gap between information-theoretic approaches to privacy and privacy-preserving machine learning.

\vspace{15pt}

\appsection{Preliminaries}
\label{AppxSec:Preliminaries}

\appsubsection{General Loss Functions for Positive Measures}
\label{ssec:GeneralLossFunctionsForPositiveMeasures}

In many data-science applications, data are represented by positive measures, including probability distributions. Such measures arise in a range of settings and are commonly modeled using either discrete representations, such as histograms, or continuous ones, such as parameterized densities \cite{sejourne2023unbalanced,bishop2006pattern,james2013introduction}.

\vspace{7pt}

\subsubsection{Divergences}

To compare positive measures, one often uses loss functions that quantify the discrepancy between them. An important class of such loss functions is given by divergences, which are generally non-negative and equal to zero when the two measures coincide. A standard example is Csiszár's class of $\mathsf{f}$-divergences \cite{csiszar1967information}, which compare two measures through a pointwise function of their Radon--Nikodym derivative.

\vspace{7pt}

\begin{definition}[$\mathsf{f}$-divergence]
Let $\mathsf{f}:(0,\infty)\to\mathbb{R}$ be a convex function such that $\mathsf{f}(1)=0$. For two probability measures $P$ and $Q$ such that $P \ll Q$, the $\mathsf{f}$-divergence from $P$ to $Q$ is defined as \cite{ali1966general, csiszar1967information}
\begin{equation}
    \D_{\mathsf{f}}(P \| Q) \coloneqq \mathbb{E}_{Q}\!\left[\mathsf{f}\!\left(\frac{\mathrm{d}P}{\mathrm{d}Q}\right)\right].
\end{equation}
\end{definition}

Several specific instances of $\mathsf{f}$-divergences are of particular interest and have different \textbf{operational meanings}. Popular instances are defined as follows \cite{csiszar2004information, polyanskiy2010channel, sharma2013fundamental, polyanskiy2014lecture, duchi2016lecture}:\vspace{4pt}
\begin{enumerate}
  \item \textbf{Kullback-Leibler (KL) Divergence:} The KL-divergence, \( \D_{\text{KL}}(P \| Q) \), is a special case of $\mathsf{f}$-divergence where the function $\mathsf{f}$ is given by \( \mathsf{f}(t) = t \log t \). It is expressed as \( \D_{\mathsf{KL}}(P \| Q) \coloneqq \D_{\mathsf{f}} (P \| Q) \) for \( \mathsf{f}(t) = t \log t \). It quantifies the amount of information lost when $Q$ is used to approximate $P$. It is widely used in scenarios like statistical inference.
  \vspace{2pt}
  \item \textbf{Total Variation Distance:} The total variation distance, denoted as $\mathsf{TV} (P, Q)$, is defined by \( \mathsf{TV}(P, Q) \coloneqq \D_{\mathsf{f}}(P \| Q) \) with the function $\mathsf{f}$ being \( \mathsf{f}(t) = |t - 1| \). It is widely used in hypothesis testing and classification tasks in statistics, providing a bound on the maximum error probability.
  \vspace{2pt}
  \item \textbf{Chi-squared ($\chi^2$) Divergence:} The $\chi^2$-divergence, \( \chi^2(P \| Q) \), is another form of $\mathsf{f}$-divergence given by \( \chi^2(P \| Q) \coloneqq \D_{\mathsf{f}} (P \| Q) \) for the function \( \mathsf{f}(t) = t^2 - 1 \). It is usually used in statistical analysis for feature selection, particularly in the context of evaluating model fit and understanding feature importance. It is also used in estimation problems.
  \vspace{2pt}
  \item \textbf{Squared Hellinger Distance:} This measure, represented as \( H^2(P, Q) \), employs the function \( \mathsf{f}(t) = (\sqrt{2} - \sqrt{t})^2 \) in its definition: \( H^2(P, Q) \coloneqq \D_{\mathsf{f}}(P \| Q) \). This distance is particularly useful in Bayesian statistics. Unlike the KL-divergence, the Hellinger distance is symmetric and bounded.
  \vspace{2pt}
  \item \textbf{Hockey-Stick Divergence:} The hockey-stick divergence, denoted as \( E_\gamma (P \| Q) \), is defined for a specific $\gamma$ (where $\gamma \geq 1$) and employs the function $\mathsf{f}(t) = (t - \gamma)_+$ with \( (a)_+ \coloneqq \max\{a, 0\} \). Therefore, \( E_\gamma(P \| Q)  \coloneqq D_{\mathsf{f}}(P \| Q) \) for \( \mathsf{f}(t) = (t - \gamma)_+ \). This divergence can be particularly useful in decision-making models and risk assessments. The contraction coefficient of this divergence is also equivalent to the local Differential Privacy \cite{asoodeh2021local}.
\end{enumerate}

\vspace{2pt}

Another important related loss is the R\'{e}nyi divergence, which is not an $\mathsf{f}$-divergence but shares a similar purpose in measuring the discrepancy between probability distributions.

\paragraph{R{\'e}nyi Divergence} 
The R{\'e}nyi divergence \cite{renyi1959measures, renyi1961measures} is denoted as \( D_{\mathsf{R}, \alpha}(P \| Q) \) for a parameter \( \alpha \), where \( \alpha \neq 1 \) and \( \alpha > 0 \). It is defined as:
\begin{equation}
    D_{\mathsf{R}, \alpha}(P \| Q) \coloneqq \frac{1}{\alpha - 1} \log \left( \mathbb{E}_Q \left[ \left( \frac{\mathrm{d}P}{\mathrm{d}Q} \right)^\alpha \right] \right).
\end{equation}
This divergence provides a spectrum of metrics between distributions, with the parameter \( \alpha \) controlling the sensitivity to discrepancies. The Kullback-Leibler divergence is a special case of R{\'e}nyi divergence as \( \alpha \rightarrow 1 \). R{\'e}nyi divergence finds extensive application in fields such as information theory, data privacy, cryptography, and machine learning, due to its adaptability and the comprehensive range of distributional differences it can capture.

\vspace{2pt}

\subsubsection{Optimal Transport Distances}

Optimal Transport (OT), a problem introduced by Gaspard Monge in the 18th century in his work `M{\'e}moire sur la th{\'e}orie des d{\'e}blais et des remblais' \cite{monge1781memoire}, emerges as a potent tool for probabilistic comparisons. It provides a uniquely flexible approach to gauge similarities and disparities between probability distributions, regardless of their supports.

\vspace{2pt}

\paragraph{Monge's OT Problem}
Monge's seminal problem seeks an optimal map $T: \mathcal{X} \rightarrow \mathcal{X}$ for transferring mass distributed according to a measure $\mu$ onto another measure $\nu$ on the same space $\mathcal{X}$. This problem can be metaphorically understood as finding the most efficient way to move sand to form certain patterns, with $\mu$ and $\nu$ representing the initial and desired distributions of sand, respectively. The \textit{key constraint} in Monge's formulation is represented by the equation $T_{\#}\mu = \nu$, where $T_{\#}$ denotes the push-forward operator. The integral equation defines the push-forward operator
$\int_{\mathcal{X}} f \circ T \, \mathrm{d} \mu = \int_{\mathcal{X}} f \, \mathrm{d} \nu, \quad \forall f \in \mathcal{C}(\mathcal{X})$,
where $\mathcal{C}(\mathcal{X})$ is the space of continuous functions on $\mathcal{X}$. This condition ensures that the measure $\mu$ is effectively transformed onto $\nu$ through the map $T$. Specifically, it implies that $T_{\#}\delta_{\mathbf{x}} = \delta_{T(\mathbf{x})}$ for Dirac measures $\delta_{\mathbf{x}}$ \cite{villani2008optimal, peyre2019computational, sejourne2023unbalanced}.

In solving Monge's problem, the objective is to find a measurable map $T$ that minimizes the total cost of transportation, subject to the aforementioned constraint. The cost of transporting a unit of mass from location $\mathbf{x}$ to location $\mathbf{y}$ in $\mathcal{X}$ is quantified by a cost function $\mathsf{c} (\mathbf{x}, \mathbf{y})$. A typical choice for $\mathsf{c} (\mathbf{x}, \mathbf{y})$, particularly in Euclidean spaces $\mathcal{X} = \mathbb{R}^d$, is the $p$-th power of the Euclidean distance, $\mathsf{c} (\mathbf{x}, \mathbf{y}) = \|\mathbf{x} - \mathbf{y}\|_p^2$. The original formulation by Monge is associated with linear transport costs, corresponding to $p = 1$. However, the quadratic case where $p = 2$ is often favored in modern applications due to its advantageous mathematical properties, including convexity and differentiability.

\begin{definition}[OT Monge Formulation Between Arbitrary Measures] 
Given two arbitrary (probability) measures $\mu$ and $\nu$ supported on $\mathcal{X}$ and $\mathcal{Y}$, respectively, the optimal transport Monge map $T^{\ast}$, if it exists, solves the following problem:
\begin{equation}
    \inf_{T} \, \left\{ \int_{\mathcal{X}} \mathsf{c} \left( \mathbf{x}, T(\mathbf{x}) \right) \, \mathrm{d} \mu(\mathbf{x}): \quad T_{\#}\mu=\nu \right\},
\end{equation}
over $\mu$-measurable map $T: \mathcal{X} \rightarrow \mathcal{Y}$.
\end{definition}


\paragraph{Kantorovich's OT Problem}
Kantorovich's formulation of the OT problem addresses the scenario of arbitrary measure spaces and introduces the concept of `mass splitting' \cite{villani2008optimal, peyre2019computational, sejourne2023unbalanced}. This innovative approach, initially developed by Kantorovich \cite{kantorovich1942transfer} for applications in economic planning, significantly extends the framework of Monge's problem. In Kantorovich's formulation, the deterministic map $T$ of Monge's problem is replaced by a probabilistic measure $\pi \in \Pi (\mu \times \nu )$, termed as a transport plan. Unlike Monge's formulation where mass moves directly from a point $\mathbf{x}$ to $T(\mathbf{x})$, Kantorovich's approach allows for the dispersion of mass from a single point $\mathbf{x}$ to multiple destinations. This flexibility makes it a generalized, or relaxed, version of Monge's problem.

\begin{definition}[Kantorovich's OT Problem]
Let $\mathcal{X}$ and $\mathcal{Y}$ be two measurable spaces. 
Let $\mathcal{P} (\mathcal{X})$ and $\mathcal{P} (\mathcal{Y})$ be the sets of all positive Radon probability measures on $\mathcal{X}$ and $\mathcal{Y}$, respectively. 
For any measurable non-negative cost function $\mathsf{c}: \mathcal{X} \times \mathcal{Y} \rightarrow \mathbb{R}^+$, the Kantorovich's OT problem between two positive measures $\mu \in \mathcal{P} (\mathcal{X})$ and $\nu \in \mathcal{P} (\mathcal{Y})$ is defined as:
\begin{subequations}
\begin{eqnarray}\label{Eq:KantorovichProblem}
    \mathsf{OT}_{\mathsf{c}} \left( \mu, \nu \right) & \coloneqq  &\mathop{\inf}_{\pi \in \Pi ( \mu, \nu )}  \int_{\mathcal{X} \times \mathcal{Y}} \mathsf{c}(\mathbf{x}, \mathbf{y}) \, \mathrm{d} \pi (\mathbf{x}, \mathbf{y}) \\
    & = & \mathop{\inf}_{\boldsymbol{\pi} \in \Pi ( \mu, \nu )} \mathbb{E}_{\pi} \left[ \, \mathsf{c}(\mathbf{X}, \mathbf{Y}) \, \right],
\end{eqnarray}    
\end{subequations}
where $\Pi ( \mu  , \nu )$ denotes the set of joint distributions (couplings) over the product space $\mathcal{X} \times \mathcal{Y}$ with marginals $\mu$ and $\nu$, respectively. That is, for all measurable sets $\mathcal{A} \subset \mathcal{X} $ and $\mathcal{B} \subset \mathcal{Y}$, we have:
\begin{multline}
    \Pi ( \mu  ,  \nu)   \coloneqq  \left\{ \pi \in \mathcal{P} ( \mathcal{X} \times \mathcal{Y} ): \; \pi(\mathcal{A} \times \mathcal{Y}) \right. \\  \left. = \mu (\mathcal{A}),  \pi (\mathcal{X} \times \mathcal{B}) = \nu (\mathcal{B}) \right\}.     
\end{multline}
\end{definition}

Having established the preliminary concepts of $\mathsf{f}$-divergences and optimal transport distances as foundational tools in data science, we now direct our attention to employing these loss functions for the quantification of privacy leakage and utility performance.

\appsubsection{Measuring Privacy Leakage and Utility Performance}
\label{SubSec:MeasuringPrivacyLeakageUtilityPerformance}

We can define a generic privacy risk loss function as a functional tied to the joint distribution $P_{\mathbf{S}, \mathbf{Z}}$, which quantifies the information leakage about $\mathbf{S}$ when $\mathbf{Z}$ is disclosed. Such a privacy risk loss function can be represented as $\mathcal{C}_{\mathsf{S}} : \mathcal{P}\left( \mathcal{S} \times \mathcal{Z}\right) \rightarrow \mathbb{R}^+ \cup \{ 0\}$. Analogously, a well-characterized and task-specific generic utility performance loss function can be formulated as a functional of the joint distribution $P_{\mathbf{X}, \mathbf{Z}}$, capturing the utility retained about $\mathbf{X}$ through the release of $\mathbf{Z}$. This utility performance loss function is denoted as $\mathcal{C}_{\mathsf{U}}: \mathcal{P}\left( \mathcal{U} \times \mathcal{Z}\right) \rightarrow \mathbb{R}^+ \cup \{ 0\}$.
We can define the $\mathsf{f}$-information between two random objects $\mathbf{X}$ and $\mathbf{Z}$ as $\I_{\mathsf{f}} \left( \mathbf{X} ; \mathbf{Z} \right) = \D_{\mathsf{f}} \left(  P_{\mathbf{X,Z}} \| P_{\mathbf{X}} P_{\mathbf{Z}} \right)$, where $\D_{\mathsf{f}} \left( \cdot \| \cdot \right)$ represents the $\mathsf{f}$-divergence \cite{polyanskiy2014lecture}, serving as a measure for both privacy (obfuscation) and utility.
Expanding this framework, Arimoto's mutual information \cite{arimoto1977information} could also be employed to assess information utility and privacy leakage. In this research, however, we focus on Shannon mutual information as our primary loss function.

\vspace{18pt}

\appsection{Connecting the Privacy Funnel Method with Other Models} 
\label{AppxSec:connection_PF_others}

%
\appsubsection{Connection with Information Bottleneck Model} 


In contrast to the Privacy Funnel (PF) model, which aims to obtain a representation $\mathbf{Z}$ that minimizes information leakage about $\mathbf{S}$ while maximizing information utility about $\mathbf{X}$, the Information Bottleneck (IB) model \cite{tishby2000information} focuses on extracting relevant information from the random variable $\mathbf{X}$ about an associated random variable $\mathbf{U}$ of interest. 
Given two correlated random variables $\mathbf{U}$ and $\mathbf{X}$ with a joint distribution $P_{\mathbf{U,X}}$, the objective of the original IB model is to find a representation $\mathbf{Z}$ of $\mathbf{X}$ through a stochastic mapping $P_{\mathbf{Z}\mid \mathbf{X}}$ that satisfies:
(i) $\mathbf{U} \markov \mathbf{X} \markov \mathbf{Z}$, and
(ii) representation $\mathbf{Z}$ is maximally informative about $\mathbf{U}$ (maximizing $\I \left( \mathbf{U}; \mathbf{Z} \right)$) while being minimally informative about $\mathbf{X}$ (minimizing $\I \left( \mathbf{X}; \mathbf{Z}\right)$).
This trade-off can be expressed by the bottleneck functional:
\begin{eqnarray}\label{Eq:IB_functional}
\mathsf{IB} \left( R^{\mathrm{u}}, P_{\mathbf{U}, \mathbf{X}}\right)  \coloneqq  \!\!\! \!\!\! \mathop{\inf}_{ \substack{P_{\mathbf{Z} \mid \mathbf{X}}:\\ \mathbf{U} \markov \mathbf{X} \markov \mathbf{Z}}}  \!\!\! \!\!\!   \I  \left( \mathbf{X}; \mathbf{Z} \right) \;\;  \mathrm{s.t.} \;\;  \I \left( \mathbf{U}; \mathbf{Z} \right) \geq R^{\mathrm{u}}. 
\end{eqnarray}
In the IB model, $\I \left( \mathbf{U}; \mathbf{Z} \right)$ is referred to as the relevance of $\mathbf{Z}$, and $\I \left( \mathbf{X}; \mathbf{Z} \right)$ is called the complexity of $\mathbf{Z}$. Since mutual information is defined as Shannon information, the complexity here is quantified by the minimum description length of compressed representation $\mathbf{Z}$. 
The IB curve is defined by the values $\mathsf{IB} \left( R , P_{\mathbf{U}, \mathbf{X}}\right)$ for different $R$.
Similarly, by introducing a Lagrange multiplier $\beta \geq 0$, the IB problem can be represented by the associated Lagrangian functional:
\begin{equation}\label{Eq:IB_LagrangianFunctional}
    \mathcal{L}_{\mathrm{IB}} \left( P_{\mathbf{Z}\mid \mathbf{X}}, \beta  \right)   \coloneqq   \I  \left( \mathbf{X}; \mathbf{Z} \right) - \beta \,   \I \left( \mathbf{U}; \mathbf{Z} \right).
\end{equation}

The formulation of the IB method in \cite{tishby2000information} has inspired numerous characterizations, generalizations, and applications \cite{makhdoumi2014information, tishby2015deep, alemi2016deep, strouse2017deterministic, vera2018collaborative, kolchinsky2018caveats, bang2019explaining, amjad2019learning, hu2019information, wu2019learnability, fischer2020conditional, federici2020learning, ding2019submodularity, hafez3information, hafez2020sample, kirsch2020unpacking}. For a review of recent research on IB models, we refer the reader to \cite{voloshynovskiyinformation, goldfeld2020information, zaidi2020information, asoodeh2020bottleneck, razeghi2023bottlenecks}.

%
\appsubsection{Connection with Complexity-Leakage-Utility Bottleneck Model} 

Given three dependent (correlated) random variables $\mathbf{U}$, $\mathbf{S}$ and $\mathbf{X}$ with joint distribution $P_{\mathbf{U ,S , X}}\,$, the goal of the CLUB model \cite{razeghi2023bottlenecks} is to find a representation $\mathbf{Z}$ of $\mathbf{X}$ using a stochastic mapping $P_{\mathbf{Z}\mid \mathbf{X}}$ such that: \textbf{(i)} $\left( \mathbf{U}, \mathbf{S} \right) \markov \mathbf{X} \markov \mathbf{Z}$, and \textbf{(ii)} representation $\mathbf{Z}$ is maximally informative about $\mathbf{U}$ (maximizing $\I \left( \mathbf{U}; \mathbf{Z} \right)$) \textbf{(iii)} while being minimally informative about $\mathbf{X}$ (minimizing $\I \left( \mathbf{X}; \mathbf{Z}\right)$) and \textbf{(iv)} minimally informative about $\mathbf{S}$ (minimizing $\I \left( \mathbf{S}; \mathbf{Z}\right)$). 
We can formulate this three-dimensional trade-off by imposing constraints on the two of them. That is, for a given information complexity and information leakage
constraints, $R^{\mathrm{z}} \geq 0$ and $R^{\mathrm{s}} \geq 0$, respectively, this trade-off can be formulated by a CLUB functional: 
\begin{eqnarray}\label{CLUB_functional}
\mathsf{CLUB} \left( R^{\mathrm{z}}, R^{\mathrm{s}}, P_{\mathbf{U}, \mathbf{S}, \mathbf{X}}\right)  \!\!\! & \coloneqq & \!\!\! \mathop{\sup}_{\substack{ P_{\mathbf{Z} \mid \mathbf{X}}: \\\left( \mathbf{U}, \mathbf{S}\right) \markov \mathbf{X} \markov \mathbf{Z}}} \I  \left( \mathbf{U}; \mathbf{Z} \right) \; \\ &\mathrm{s.t.} &   \I \left( \mathbf{X}; \mathbf{Z} \right) \leq R^{\mathrm{z}}, \;\; 
\I \left( \mathbf{S}; \mathbf{Z} \right) \leq R^{\mathrm{s}}. \nonumber
\end{eqnarray}
Setting $\mathbf{U} \equiv \mathbf{X}$ and $R^{\mathrm{z}} \geq \H \left( P_{\mathbf{X}}\right)$ in the CLUB objective \eqref{CLUB_functional}, the CLUB model reduces to the discriminative (classical) PF model \eqref{Eq:PF_LagrangianFunctional}. 

%
\appsubsection{Connection with Image-to-Image Transition Models}

Consider two measurable spaces $\mathcal{X}$ and $\mathcal{Y}$. Let $\mathbf{X} \sim P_{\mathbf{X}}$ and $\mathbf{Y} \sim P_{\mathbf{Y}}$ be random objects representing random realizations from these spaces, with distributions $P_{\mathbf{X}}$ and $P_{\mathbf{Y}}$ respectively, where $\mathbf{X} \in \mathcal{X}$ and $\mathbf{Y} \in \mathcal{Y}$. 
Let $f : \mathcal{X} \rightarrow \mathcal{Y}$ and $g : \mathcal{Y} \rightarrow \mathcal{X}$ denote appropriate mappings (or functions) that map elements between these spaces.

The objective of the image-to-image translation problem is to find (learn) a mapping $f : \mathcal{X} \rightarrow \mathcal{Y}$ (or vice versa $g : \mathcal{Y} \rightarrow \mathcal{X}$) such that \textbf{(i)} the distribution of the mapped object approximates the distribution of the target object, i.e., 
$P_{f \left(\mathbf{X}\right)} \approx P_{\mathbf{Y}}$ and/or $P_{\mathbf{X}} \approx P_{g\left(\mathbf{Y}\right)}$; and \textbf{(ii)} the mapping preserves or captures specific characteristics or features of the input images.  
This can be formally expressed as a constraint optimization problem, where the mapped images maintain certain predefined properties or metrics of similarity with the input images.
This is a fundamental aspect of tasks like style transfer, domain adaptation, or generative modeling.

Let $\mathcal{C}_{\mathsf{U}} \big( P_{f\left(\mathbf{X}\right) , \mathbf{Y}} \big) = \mathsf{dist} \left( P_{f\left(\mathbf{X}\right)} , P_{\mathbf{Y}} \right)$, where $\mathsf{dist} \left( P_{f\left(\mathbf{X}\right)} , P_{\mathbf{Y}} \right)$ is a discrepancy measure between $P_{f\left(\mathbf{X}\right)}$ and $P_{\mathbf{Y}}$. For instance, one can consider $\mathsf{dist} ( P_{f\left(\mathbf{X}\right)}, P_{\mathbf{Y}} ) = \D_{\mathsf{f}} (P_{f\left(\mathbf{X}\right)}  \Vert P_{\mathbf{Y}} )$, or alternatively, one can use the Maximum Mean Discrepancy (MMD) for a characteristic positive-definite reproducing kernel \cite{tolstikhin2018wasserstein}. 
Now, we can consider an optimization problem where the objective is to minimize a loss function that quantifies both the distributional similarity and the preservation of image characteristics:
\begin{multline}
\label{image-to-image-problem}
 \mathop{\min}_{f, g}\;\;\;
 \mathsf{dist} \left( P_{f\left(\mathbf{X}\right)} , P_{\mathbf{Y}} \right) + 
 \mathsf{dist} ( P_{g\left(\mathbf{Y}\right)} , P_{\mathbf{X}} ) \\
 + \lambda_x \Phi_x ( \mathbf{X}, f \left(\mathbf{X}\right) ) +
 \lambda_y \Phi_y ( \mathbf{Y}, g \left(\mathbf{Y}\right) ) . 
\end{multline}
We can leverage image-to-image transition models from this perspective within a domain-preserving privacy funnel method. This method diverges from traditional obfuscation techniques for the sensitive attribute $\mathbf{S}$. Instead, it involves deliberate manipulation of image attributes in a \textit{random} manner. The defender \textit{generates} and \textit{releases} a manipulated image achieved by \textit{uniformly} selecting a \textit{random} attribute from the set of events pertinent to $\mathbf{S}$.

%

\vspace{10pt}

\appsection{Estimation of Mutual Information via MINE}
\label{AppxSec:MINE}

The Mutual Information Neural Estimation (MINE) method \cite{belghazi2018mutual} employs the Donsker--Varadhan representation of the Kullback--Leibler divergence \cite{donsker1983asymptotic} to estimate mutual information between random variables. This approach is particularly useful in high-dimensional settings, where traditional estimation methods may be less reliable. The Donsker--Varadhan representation of the Kullback--Leibler divergence $\D_{\mathrm{KL}}(P \| Q)$ between two probability distributions $P$ and $Q$ is given by

\begin{theorem}[Donsker-Varadhan Representation]
The KL divergence admits the dual representation \cite{donsker1983asymptotic}:
\begin{equation}
    \D_{\mathrm{KL}} (P \| Q) = \sup_{T \in \mathcal{T}} \mathbb{E}_{P}[T] - \log(\mathbb{E}_{Q}[e^{T}]),
\end{equation}
where $\mathcal{T}$ is a class of measurable functions for which the expectations are finite.
\end{theorem}

Mutual information $\I( \mathbf{X}; \mathbf{Y} )$ between random objects $\mathbf{X}$ and $\mathbf{Y}$ is defined using the KL divergence $\I( \mathbf{X} ; \mathbf{Y} ) = \D_{\mathrm{KL}}(P_{\mathbf{XY}} \| P_{\mathbf{X}} P_{\mathbf{Y}})$. 
In the MINE framework, we utilize a neural network parameterized by $\boldsymbol{\theta}_{\mathsf{MINE}}$\footnote{We use subscript $\mathsf{MINE}$ to distinguish it from our parameterized utility decoder $\boldsymbol{\theta}$ utilized in our DVPF model.}, denoted as \( T_{\boldsymbol{\theta}_{\mathsf{MINE}}} \), to approximate functions in \(\mathcal{T}\). The estimated mutual information $\widehat{\I}_{\boldsymbol{\theta}_{\mathsf{MINE}}}(\mathbf{X}; \mathbf{Y})$ is given by:
\begin{equation}
    \widehat{I}_{\boldsymbol{\theta}_{\mathsf{MINE}}}( \mathbf{X}; \mathbf{Y}) = \sup_{\boldsymbol{\theta}_{\mathsf{MINE}} \,\in \, \boldsymbol{\Theta}} \mathbb{E}_{P_{\mathbf{XY}}}[T_{\boldsymbol{\theta}_{\mathsf{MINE}}}] - \log(\mathbb{E}_{P_{\mathbf{X}} P_{\mathbf{Y}}}[e^{T_{\boldsymbol{\theta}_{\mathsf{MINE}}}}]),
\end{equation}
where \(P_{\mathbf{XY}}\) is the joint distribution of \(\mathbf{X}\) and \(\mathbf{Y}\), and \(P_{\mathbf{X}} P_{\mathbf{Y}}\) is the product of their marginal distributions.

The neural network is trained by maximizing $\widehat{\I}_{\boldsymbol{\theta}_{\mathsf{MINE}}}(\mathbf{X}; \mathbf{Y})$ using stochastic gradient descent. This is done by sampling from $P_{\mathbf{XY}}$ and $P_{\mathbf{X}} P_{\mathbf{Y}}$, and iteratively updating $\boldsymbol{\theta}_{\mathsf{MINE}}$ to maximize the estimated mutual information. The performance of MINE depends on several factors, including the network architecture, the optimization strategy, and the choice of hyperparameters. The capacity of the network and the convergence behavior of the optimization procedure also affect the accuracy of the mutual information estimate.

In our study, we implemented \textit{an improved version of MINE} in PyTorch, with several modifications aimed at practical use. These include a modular code structure, improved network initialization, a revised sampling procedure, an adaptive learning-rate scheduler, and a configurable optimizer. The PyTorch pseudocode for the implementation is given in Algorithm~\ref{Algorithm:PyTorch_Pseudo-Code_MINE}.

\begin{algorithm}
\caption{Pseudocode for Mutual Information Neural Estimation (MINE)}
\label{Algorithm:PyTorch_Pseudo-Code_MINE}
\begin{algorithmic}[1]
\small
\setstretch{1.15}
\Require dim\_x, dim\_y, moving\_average\_rate, hidden\_size, network\_type, batch\_size, n\_iterations, learning\_rate, n\_verbose, n\_window, save\_progress
\Ensure Estimated mutual information between \(X\) and \(Y\)
\State Initialize the neural network (MLP or CNN) according to \texttt{network\_type} and \texttt{hidden\_size}
\State Apply Xavier initialization to the network weights
\State Initialize the network biases
\State \textbf{Class MINE:}
\State \hspace{\algorithmicindent} Define the MINE model using the initialized network
\State \hspace{\algorithmicindent} Initialize \texttt{moving\_average\_exp\_t} as \(1.0\)
\Function{ForwardPass}{x, y}
    \State Concatenate the input tensors \(x\) and \(y\)
    \State Pass the concatenated input through the network
    \State \Return the network output
\EndFunction
\Function{TrainMINE}{dataset}
    \State Set the MINE model to training mode
    \State Initialize the optimizer (Adam or RMSprop) with \texttt{learning\_rate}
    \State Initialize the learning-rate scheduler
    \State Initialize an array to store MI estimates over the last \texttt{n\_window} iterations
    \State Optionally initialize a tensor to store MI progress
    \For{\texttt{iteration} \(= 1\) to \texttt{n\_iterations}}
        \State Sample a joint minibatch \((x, y)\) from the dataset
        \State Construct a marginal minibatch \((x, \tilde{y})\)
        \State Compute \(t = \Call{ForwardPass}{x, y}\)
        \State Compute \(\tilde{t} = \Call{ForwardPass}{x, \tilde{y}}\)
        \State Compute \(\exp(\tilde{t})\) and update
        \texttt{moving\_average\_exp\_t} using \texttt{moving\_average\_rate}
        \State Compute the loss as the negative MINE lower bound
        \State Backpropagate the loss and update the model parameters using the optimizer
        \State Update the learning-rate scheduler
        \State Store the current MI estimate
        \If{\texttt{iteration} \% \texttt{n\_verbose} \(= 0\)}
            \State Print the average MI over the last \texttt{n\_window} iterations
        \EndIf
        \If{\texttt{save\_progress} \(> 0\) and \texttt{iteration} \% \texttt{save\_progress} \(= 0\)}
            \State Save the current MI estimate to \texttt{mi\_progress}
        \EndIf
    \EndFor
    \State \Return the average MI over the last \texttt{n\_window} iterations
\EndFunction
\Function{EvaluateMI}{x, y}
    \State Split \(x\) and \(y\) into batches
    \State Initialize a variable to accumulate MI estimates
    \For{each batch of \(x\) and \(y\)}
        \State Construct the corresponding marginal batch \((x, \tilde{y})\)
        \State Compute the batch MI estimate
        \State Accumulate the batch MI estimate
    \EndFor
    \State \Return the average MI over all batches
\EndFunction
\end{algorithmic}
\end{algorithm}

\vspace{15pt}

\appsection{Training Details}
\label{AppxSec:TrainingDetails}

\appsubsection{The Role of Randomness in DVPF Training}

In the DVPF model, we introduce two additional sources of randomness during training, beyond the stochasticity induced by the reparameterization trick: \textbf{(i)} additive noise in the latent representation, and \textbf{(ii)} dropout in the intermediate layers.

\vspace{3pt}

\subsubsection{Integration of Noise in Latent Representation}

The latent representation vector \( \mathbf{Z} \in \mathbb{R}^n \) is perturbed by additive Gaussian noise. Specifically, we add a noise vector \( \mathbf{N} \in \mathbb{R}^n \) whose entries are i.i.d. Gaussian random variables with variance $\sigma^2 = \frac{1}{2\pi e}$. Hence, $\mathbf{N} \sim \mathcal{N}(0,\sigma^2 \mathbf{I}_n)$, where $\mathbf{I}_n$ denotes the \( n \times n \) identity matrix.
The differential entropy of \( \mathbf{N} \) is
\begin{equation}
\mathrm{h}(\mathbf{N}) = \frac{n}{2}\ln(2\pi e \sigma^2) = \frac{n}{2}\ln\!\left(2\pi e \cdot \frac{1}{2\pi e}\right) = 0.
\end{equation}
This follows directly from the choice $\sigma^2 = \frac{1}{2\pi e}$. Note that the differential entropy is zero should not be interpreted as meaning that there is no randomness. The noise still has nonzero variance and therefore introduces stochasticity into the latent representation. During training, this added stochasticity serves as a regularizer and can help reduce overfitting and improve generalization.

\vspace{3pt}

\subsubsection{Application of Dropout in Intermediate Layers}

The DVPF model also uses Gaussian noise in the latent space and dropout in the intermediate layers. During training, dropout randomly disables a fraction of neurons at each update. This adds randomness to the learning process and helps reduce overfitting.
We use dropout in the hidden layers so that the network does not rely too heavily on any single set of activations. Instead, it is encouraged to learn more distributed representations, which generally improve generalization and are preferable here from a privacy standpoint.

\appsubsection{Alpha Scheduler}

The \texttt{AlphaScheduler} class controls the parameter \(\alpha\) during neural network training. It is initialized with the total number of training epochs (\texttt{num\_epochs}), the initial and final values of \(\alpha\) (\texttt{alpha\_start} and \texttt{alpha\_end}), and the linear increment used in the early stage of training (\texttt{linear\_increment}). The schedule has two phases. In the first phase, which spans roughly the first third of training, \(\alpha\) increases linearly. In the second phase, \(\alpha\) is updated according to a logistic schedule so that it approaches its final value gradually rather than changing too abruptly.

The \texttt{AlphaScheduler} also allows the linear growth rate and the steepness of the logistic curve to be adjusted. In addition, it provides tools to visualize and log the value of \(\alpha\) over training epochs, which helps monitor and tune the training process.

Furthermore, \(\alpha\) is used as a complexity coefficient related to the encoding rate, or equivalently the compression bit rate. Increasing \(\alpha\) gradually allows the model to be trained progressively across different complexity levels. For a given value of \(\alpha\), we evaluate the corresponding utility and privacy-leakage tradeoff. When training the model at a larger value of \(\alpha\), we initialize from a model trained at a smaller value, rather than training again from scratch. This progressive training strategy makes optimization more stable and reduces training cost across complexity levels.

\begin{figure}[t]
    \centering
    \includegraphics[width=0.95\linewidth]{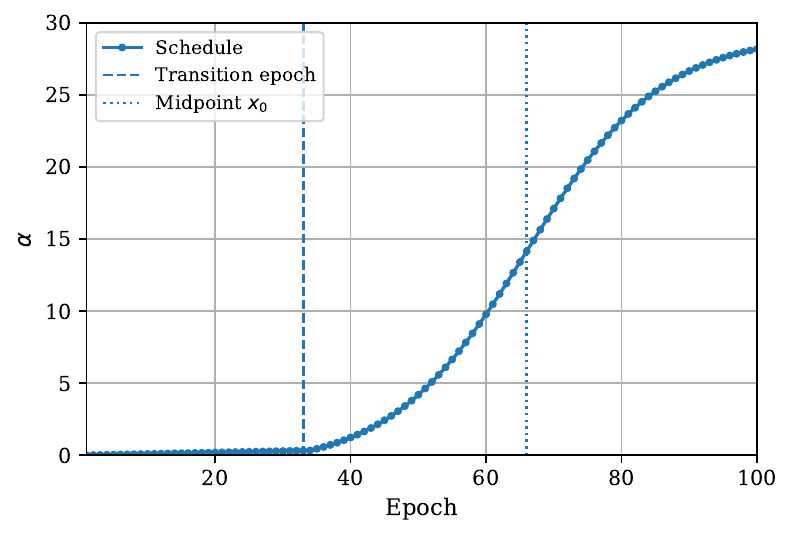}
    \vspace{-5pt}
    \caption{Phase structure of the alpha scheduler. The dashed line marks the end of the linear phase, while the dotted line indicates the logistic midpoint $x_0$.}
    \label{fig:alpha_scheduler}
\end{figure}

Figure~\ref{fig:alpha_scheduler} illustrates the evolution of the scheduling parameter $\alpha$. The scheduler is defined by two successive phases: a linear-growth stage in the early epochs and a logistic-growth stage thereafter. The marked transition point separates these phases, and the midpoint $\mathrm{x}_0$ identifies the region where the logistic increase becomes most pronounced.

\vspace{5pt}

\appsubsection{Uncertainty Decoder (Conditional Generator)}

The decoder uses Feature-wise Linear Modulation (FiLM) to condition the activations of each layer on \(\mathbf{S}\). To do this, the \textsf{\_film\_generator} method uses dedicated gamma and beta networks, implemented as small MLPs, to generate scaling and shifting coefficients from $\mathbf{S}$. These coefficients are then applied to the layer activations, so that the decoder output depends explicitly on the conditioning variable $\mathbf{S}$.

\vspace{10pt}

\appsection{Generative Privacy Funnel in Face Recognition Systems}
\label{AppxSec:DenPF} 

For synthetic data generation targeted at facial recognition, demographic information such as age, gender, ethnicity, and other physical attributes must be carefully incorporated into the data to enhance the system's ability to recognize a large and diverse set of human faces. In addition to these attributes, different expressions (e.g. neutral, happy, etc.) at different orientations (e.g. frontal, profile, etc.) must also be captured and included in the data. Moreover, indoor and outdoor environmental settings and varying lighting conditions (both static and dynamic) must also be included to simulate real-world scenarios as much as possible. High-resolution images (i.e. large input size) are also necessary for effectively extracting fine facial features, and images at lower resolutions are also required in order to handle suboptimal face images effectively.

%
\begin{figure}[t!]
    \centering
    \begin{subfigure}[h]{0.45\textwidth}
        \includegraphics[width=\linewidth]{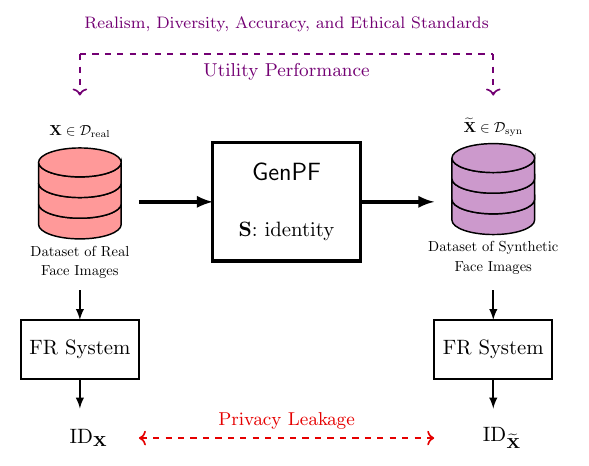}%
        \vspace{-7pt}
        \caption{}
        \label{fig:GPF_Example_FaceRecognition}
    \end{subfigure}%
\\
    \begin{subfigure}[h]{0.45\textwidth}
        ~~~~~~\includegraphics[width=\linewidth]{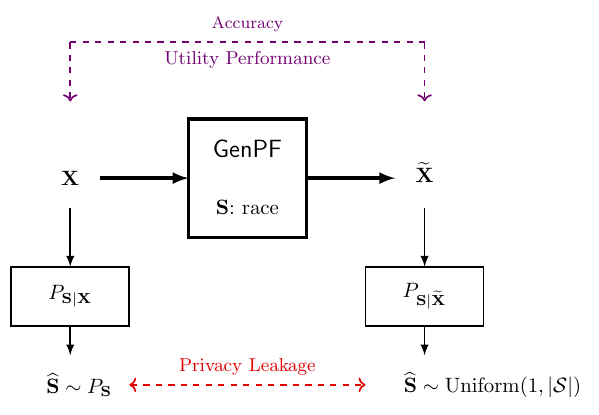}%
        \vspace{-7pt}
        \caption{}
        \label{fig:GPF_Example_FaceAttributeRecognition}
    \end{subfigure}
    \caption{Visualization of the Generative Privacy Funnel in (a) face recognition systems and (b) face attribute recognition.}
    \label{Fig:GPF_Example_FaceImages}
\end{figure}

Images of people wearing glasses, or with part of their face obscured in some other way, should also be included in the database to allow better face recognition in real-world scenes. Another important aspect is to ensure that accurate and consistent labels are associated with the images. Ethical considerations should be taken into account when generating images to avoid introducing bias into the dataset. Realism in the generated images is also critical for the task at hand. If realistic images are not generated, this can significantly affect the performance of the facial recognition system. Thus, it is important to take a holistic approach when generating the dataset.

Incorporating the principles laid out in the comprehensive approach to synthetic dataset generation for facial recognition systems, the $\mathsf{GenPF}$ is aimed to generate synthetic images that not only adhere to the above-mentioned criteria but also protect the sensitive information from real dataset samples. This may include protecting personal \textit{identities} as well as \textit{sensitive attributes} such as gender, race, and emotion inherent in facial images (See Figure~\ref{Fig:GPF_Example_FaceImages}). 
Moreover, $\mathsf{GenPF}$ has the potential to contribute to the creation of a balanced dataset, a crucial step in mitigating biases in face recognition systems. The specifics of this are discussed in Sec.~\ref{SubSec:GenPF_for_BiasMitigation}

\vspace{30pt}

%
\appsection{Face Recognition Experiments}
\label{AppxSec:Experiments}

Face recognition systems represent an important segment of the biometric technology market. They are used to identify or verify a person from a digital image or video frame by analyzing facial features. Biometric face recognition systems (also known as facial recognition systems) identify or verify a person by comparing a facial image or video frame with images or templates stored in a database. Face recognition technology is increasingly used in security and surveillance, as well as in online social media platforms and smartphone apps.

\appsubsection{Face Recognition Leading Models and Their Core Mechanisms}

The evolution of face recognition technology has been significantly influenced by the development of several groundbreaking models, each distinguished by its unique features and mechanisms. Prominent among these are \textbf{DeepFace} \cite{Taigman2014DeepFaceCT}, \textbf{FaceNet} \cite{schroff2015facenet}, \textbf{OpenFace} \cite{amos2016openface}, \textbf{SphereFace} \cite{liu2017sphereface}, \textbf{CosFace} \cite{wang2018cosface}, \textbf{ArcFace} \cite{arcface2019}, and \textbf{AdaFace} \cite{kim2022adaface}. These models have advanced the field through their innovative use of deep learning techniques, setting new standards in accuracy and reliability for face recognition tasks.

\textbf{DeepFace}, developed by Facebook, employs a deep neural network with over 120 million parameters, demonstrating notable robustness against pose variations through advanced 3D modeling techniques. \textbf{FaceNet}, from Google, uses a `triplet loss' function to optimize distances between anchor, positive, and negative images. Despite its effectiveness, FaceNet faces challenges related to the large number of triplets in extensive datasets and complexities in mining semi-hard samples.
\textbf{OpenFace}, a Carnegie Mellon University innovation, offers a lightweight yet efficient alternative, focusing on `TripletHardLoss' for challenging sample selection during training. This model excels in environments with limited computational resources. 
Subsequent to OpenFace, \textbf{SphereFace} introduced an angular margin penalty in its loss function to enhance intra-class compactness and inter-class separation. SphereFace, however, encountered training stability challenges due to the need for computational approximations in its loss function.
Building on these advancements, \textbf{CosFace} added a cosine margin penalty directly to the target logit, simplifying the implementation and improving performance without requiring joint supervision from the softmax loss. This marked a significant step forward in the development of margin-based loss functions. \textbf{ArcFace}, from InsightFace, further refined the approach by introducing an 'Additive Angular Margin Loss', which optimizes the geodesic distance margin on a normalized hypersphere. Known for its ease of implementation and computational efficiency, ArcFace achieved state-of-the-art performance across various benchmarks.
Most recently, \textbf{AdaFace} has represented a significant leap in addressing image quality variations in face recognition. By correlating feature norms with image quality, AdaFace adapts its margin function to emphasize hard samples in high-quality images and de-emphasize them in lower-quality ones. This adaptive approach, blending angular and additive margins based on image quality, represents a notable advancement in the field.

\appsubsection{Backbone Architectures for Feature Extraction}

In face recognition systems, backbone architectures are necessary for extracting and learning high-level features from raw input images. They are a fundamental component of face recognition models and directly affect how well facial features can be learned, which in turn influences recognition performance.
One of the key architectures in this domain is the Improved ResNet, or \textbf{iResNet} \cite{duta2021improved}. As an advanced iteration of the ResNet \cite{resnet2016} model, iResNet integrates modifications that aim to resolve issues related to the degradation of deeper networks. It is characterized by its residual learning framework, which effectively tackles the vanishing gradient problem, a common challenge with deep neural networks. This allows for the training of networks with increased depth, thereby facilitating a more profound extraction of facial features.
The modularity of iResNet, which can be adapted to various depths, provides the flexibility to balance computational efficiency and model accuracy based on the specific requirements of a given task. This adaptability extends the use of iResNet across different face recognition models, each leveraging the architecture's strengths according to their individual design principles.
Other backbone architectures, such as \textbf{VGGNet} \cite{simonyan2014very} and \textbf{MobileNet} \cite{howard2017mobilenets}, are also employed in the design of face recognition models. VGGNet, with its homogeneously stacked convolutional layers, excels in extracting features from input images of varying complexity. On the other hand, MobileNet, with its depthwise separable convolutions, offers an efficient, lightweight solution optimal for mobile and edge computing applications.
The choice of backbone architecture significantly influences the face recognition model's performance, shaping its ability to extract necessary features, adapt to varying task complexities, and function efficiently within the given computational constraints. As such, selecting the most suitable architecture is crucial for the successful deployment of a face recognition system.

\vspace{5pt}

\appsubsection{Datasets Used for Training and Validation}

The performance of face recognition systems depends strongly on the datasets used for training, validation, and evaluation. These datasets should capture a range of variations in facial appearance, such as pose, illumination, expression, occlusion, age, and demographic attributes.

The \textbf{MSCeleb1M} dataset \cite{deng2019lightweight_ms1mv3} has been widely used for training face recognition models. Its large scale and diversity of facial appearance make it useful for learning representations that are robust to variations in pose, expression, illumination, and occlusion.

The \textbf{WebFace} dataset \cite{zhu2021webface260m} provides a large-scale face dataset for training deep models. With nearly half a million images from over 10,000 individuals, it offers a diverse range of facial images sourced from the internet. 
It provides a diverse collection of facial images collected from the internet and is commonly used for large-scale model development and benchmarking.

The \textbf{MORPH} dataset \cite{morph1} distinguishes itself with its focus on longitudinal facial data, charting the progression of facial features over time. The inclusion of aging-related variations makes this dataset crucial for the development of age-invariant face recognition capabilities, an essential attribute for models deployed in dynamic, real-world scenarios.

The \textbf{FairFace} dataset \cite{karkkainenfairface} is an intervention in the realm of equitable face recognition. Designed to mitigate racial and demographic biases, it includes a balanced representation of seven racial groups and a diverse distribution of age and gender within each group. With approximately 100,000 (exactly 108,501) images, FairFace is a valuable resource for training and evaluating face recognition systems, ensuring they perform fairly across different demographics. This dataset is particularly crucial for developing models that can operate justly in multicultural societies, where fairness and inclusivity are paramount.

For unconstrained face recognition, the Labeled Faces in the Wild (\textbf{LFW}) \cite{huang2008labeled} and the IARPA Janus Benchmark-C (\textbf{IJBC}) \cite{ijbc} datasets have made significant contributions. The LFW dataset comprises images collected from the internet, encapsulating the real-world conditions a face recognition system is likely to encounter, including variability in pose, lighting, and expression. IJBC, on the other hand, provides a challenging, large-scale evaluation of face recognition technology under uncontrolled conditions. It includes several variations such as pose, illumination, expression, race, and age, thereby pushing the boundaries of model performance.

\vspace{4pt}

\appsubsection{Metrics Used to Evaluate Face Recognition Model Performance}

In this subsection, we define the metrics used to evaluate the performance of the face recognition models in our experiments. This overview may be helpful for readers who are less familiar with biometric verification and the interpretation of the reported performance measures. Readers already familiar with these concepts may skip this material.

\vspace{7pt}

\subsubsection{False Match Rate ($\mathsf{FMR}$)}

The False Match Rate ($\mathsf{FMR}$), also referred to as the False Acceptance Rate ($\mathsf{FAR}$), measures how often the system incorrectly accepts an impostor attempt as a genuine match. It is computed as the fraction of impostor verification attempts that are falsely accepted:

\begin{equation}
    \mathsf{FMR} = \frac{\mathsf{Number~of~False~Acceptances}}{\mathsf{Total~Imposter~Verification~Attempts}}.
\end{equation}

A lower $\mathsf{FMR}$ indicates a lower risk of falsely accepting impostor attempts. In practice, $\mathsf{FMR}$ depends on the decision threshold: using a stricter threshold typically reduces $\mathsf{FMR}$, but may increase the False Rejection Rate ($\mathsf{FRR}$).

\vspace{7pt}

\subsubsection{True Match Rate ($\mathsf{TMR}$)}

The True Match Rate ($\mathsf{TMR}$), also called the True Acceptance Rate (TAR), measures how often the system correctly accepts genuine matches. It is computed as the fraction of genuine verification attempts that are correctly accepted:

\begin{equation}
    \mathsf{TMR} = \frac{\mathsf{Number~of~True~Acceptances}}{\mathsf{Total~Genuine~Verification~Attempts}}.
\end{equation}

A higher $\mathsf{TMR}$ indicates better performance on genuine verification attempts. As with $\mathsf{FMR}$, its value depends on the decision threshold. Increasing $\mathsf{TMR}$ often comes at the cost of a higher $\mathsf{FMR}$, so both metrics should be considered together.

\subsubsection{Accuracy ($\mathsf{Acc}$)}

Accuracy measures the proportion of correct verification decisions over all attempts. It is computed as the ratio of correct decisions, i.e., true positives and true negatives, to the total number of verification attempts:

\begin{equation}
\small
    \mathsf{Acc} = \frac{\mathsf{Number~of~True~Positives} + \mathsf{Number~of~True~Negatives}}{\mathsf{Total~Verification~Attempts}}.
\end{equation}

A higher accuracy indicates that the system makes fewer incorrect decisions overall. However, accuracy should be interpreted with care, especially when the numbers of genuine and impostor attempts are imbalanced. For this reason, $\mathsf{TMR}$ and $\mathsf{FMR}$ are often more informative in biometric verification settings.

\vspace{8pt}

\subsubsection{Shannon Entropy}

Entropy measures the uncertainty of a random variable. For a discrete random variable $\mathbf{S}$ with probability mass function $P_{\mathbf{S}}$, the Shannon entropy is defined as
\begin{equation}
    \H(\mathbf{S}) = - \sum_{s \in \mathcal{S}} P_{\mathbf{S}}(s)\,\log P_{\mathbf{S}}(s).
\end{equation}
In our setting, $\H(\mathbf{S})$ quantifies the uncertainty in the distribution of the sensitive labels $\mathbf{S}$. The maximum entropy of a discrete random variable with alphabet $\mathcal{S}$ is $\log |\mathcal{S}|$, and it is attained when $\mathbf{S}$ is uniformly distributed over $\mathcal{S}$. For example, if $\mathbf{S}$ denotes gender with two categories, then the maximum entropy is $\log_2 2 = 1$; if $\mathbf{S}$ has four categories, then the maximum entropy is $\log_2 4 = 2$.

When the entropy is smaller than $\log |\mathcal{S}|$, the distribution of $\mathbf{S}$ is not uniform. In that case, some labels occur more frequently than others, so the variable is more predictable than in the uniform case.

\vspace{8pt}

\subsubsection{Mutual Information}

Mutual information quantifies how much knowing one variable reduces uncertainty about another. In our setting, it measures how much information the embedding \(\mathbf{Z}\) contains about the sensitive label \(\mathbf{S}\). Since \(\mathbf{S}\) is discrete, it can be written as
\begin{equation}
\I(\mathbf{S}; \mathbf{Z}) = \H(\mathbf{S}) - \H(\mathbf{S}\mid \mathbf{Z}),
\end{equation}
where \(\H(\mathbf{S})\) is the entropy of \(\mathbf{S}\) and \(\H(\mathbf{S}\mid \mathbf{Z})\) is the remaining uncertainty about \(\mathbf{S}\) after observing \(\mathbf{Z}\). Thus, \(\I(\mathbf{S}; \mathbf{Z})\) represents the reduction in uncertainty about the sensitive labels due to the embeddings. When \(\I(\mathbf{S}; \mathbf{Z})\) is close to \(\H(\mathbf{S})\), the embeddings reveal a large amount of information about the labels; when it is close to zero, they reveal little. Mutual information is symmetric, i.e., $\I(\mathbf{S}; \mathbf{Z}) = \I(\mathbf{Z}; \mathbf{S})$, and, since conditioning cannot increase entropy, $\I(\mathbf{S}; \mathbf{Z}) \leq \H(\mathbf{S})$.

\vspace{2pt}

\appsubsection{Experimental Setup}

\begin{figure}[!t]
    \centering
    \begin{subfigure}[h]{0.38\textwidth}
    \centering
    \includegraphics[height=3.3cm]{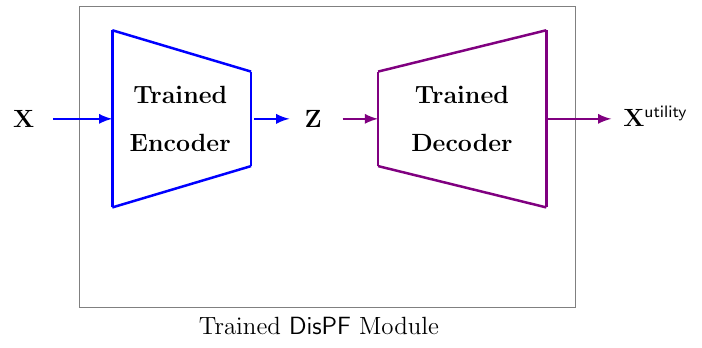}%
    \caption{}
    \label{Fig:Trained_DisPF_module}
    \end{subfigure}%
\\
%
%
    \begin{subfigure}[h]{0.38\textwidth}
    \centering
    ~\includegraphics[height=3.3cm]{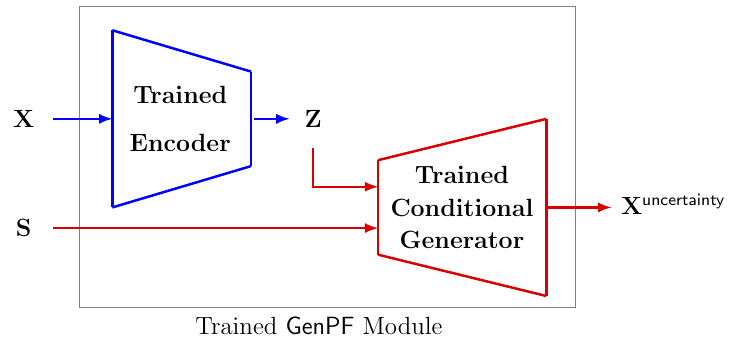}%
    \caption{}
    \label{Fig:Trained_GenPF_module}
    \end{subfigure} 
    %
    \caption{The $\mathsf{DisPF}$ and $\mathsf{GenPF}$ modules have been trained and are designed for integration in a plug-and-play manner. These modules are characterized by a set of specific parameters: `dataset name' (for example, FairFace), which denotes the dataset utilized; `sensitive attribute name' (e.g., Race); `alpha' (e.g., 0.1); `latent $\mathbf{Z}$ dimension' (e.g., 128); `backbone' (e.g., iResNet~50); `loss function' (e.g., arcface); and `backbone trained dataset' (e.g., WebFace12M).}
    \label{Trained_DisPF_GenPF_modules}
 \vspace{-5pt} 
\end{figure}

We consider the state-of-the-art FR backbones with three variants of iResNet \cite{resnet2016,arcface2019} architecture (iResNet100, iResNet50, and iResNet18). These architectures have been trained using either the MS1MV3 \cite{deng2019lightweight_ms1mv3} or WebFace4M/12M \cite{zhu2021webface260m} datasets.
For loss functions, ArcFace \cite{arcface2019} and AdaFace \cite{kim2022adaface} methods were employed. 
For the training phase, we utilized pre-trained models sourced from the aforementioned studies. All input images underwent a standardized pre-processing routine, encompassing alignment, scaling, and normalization. This was in accordance with the specifications of the pre-trained models. We then trained our networks using the Morph dataset \cite{morph1} and FairFace \cite{karkkainenfairface}, focusing on different demographic group combinations such as race and gender.
\autoref{TrainDVPF_FaceRecognition_EmbeddingBased} depicts our framework during the \textit{training} phase for a specific setup, which we will explain later. \autoref{Trained_DisPF_GenPF_modules} shows the trained modules. \autoref{InferenceDVPF_FaceRecognition_EmbeddingBased} illustrates our framework during the \textit{inference} phase. 

\vspace{2pt}

%

\subsubsection{Learning Scenarios}
\vspace{1pt}

We consider two forms of input data for $\mathbf{X}$: \textbf{(i)} \textbf{raw images}, and \textbf{(ii)} \textbf{feature representations}, commonly referred to as \textit{embeddings}, extracted from facial images. When raw images are used, we consider two encoder types: \textbf{(i)} a \textbf{custom encoder} trained from scratch, and \textbf{(ii)} a \textbf{backbone encoder} based on a pre-trained network that is further fine-tuned during training. When embeddings are used as input, we employ a custom MLP encoder trained from scratch.  
Based on the objectives of the utility and uncertainty decoders, we consider two decoder tasks: \textbf{(i)} \textbf{reconstruction}, and \textbf{(ii)} \textbf{classification}. Combining these design choices, we study \textit{three learning scenarios}:

\vspace{1pt}

\textit{End-to-End Raw Data Scratch Learning:}
In this setting, we train a custom encoder model, together with the other networks, from scratch using raw data samples as input. The model learns representations directly from the input data without relying on a pre-trained model. This setting is appropriate when the dataset is sufficiently large and diverse to support end-to-end training from scratch.

\vspace{1pt}

\textit{Raw Data Transfer Learning with Fine-Tuning:}
In this setting, we use a pre-trained model as the backbone and fine-tune it on the target dataset. A selected intermediate layer of the backbone is used as the latent representation. This setting is appropriate when the target dataset is relatively small or specialized, and fine-tuning a pre-trained model is more effective than training a model from scratch.


\textit{Embedding-Based Data Learning:}
In this setting, we use a MLP projector as the encoder, with pre-extracted face embeddings as input. This approach is appropriate when a face recognition model has already learned informative features from a large and diverse dataset. Using these embeddings can reduce the computational cost of end-to-end training on raw images while still providing useful input representations. \autoref{TrainDVPF_FaceRecognition_EmbeddingBased} shows an example of our training framework for this setting.

\vspace{-3pt}

\appsubsection{Extended Results: Visualizing DVPF Effects on FairFace}

\begin{figure}[h]
    \centering
    \begin{subfigure}[h]{0.43\linewidth}
        \includegraphics[width=\linewidth]{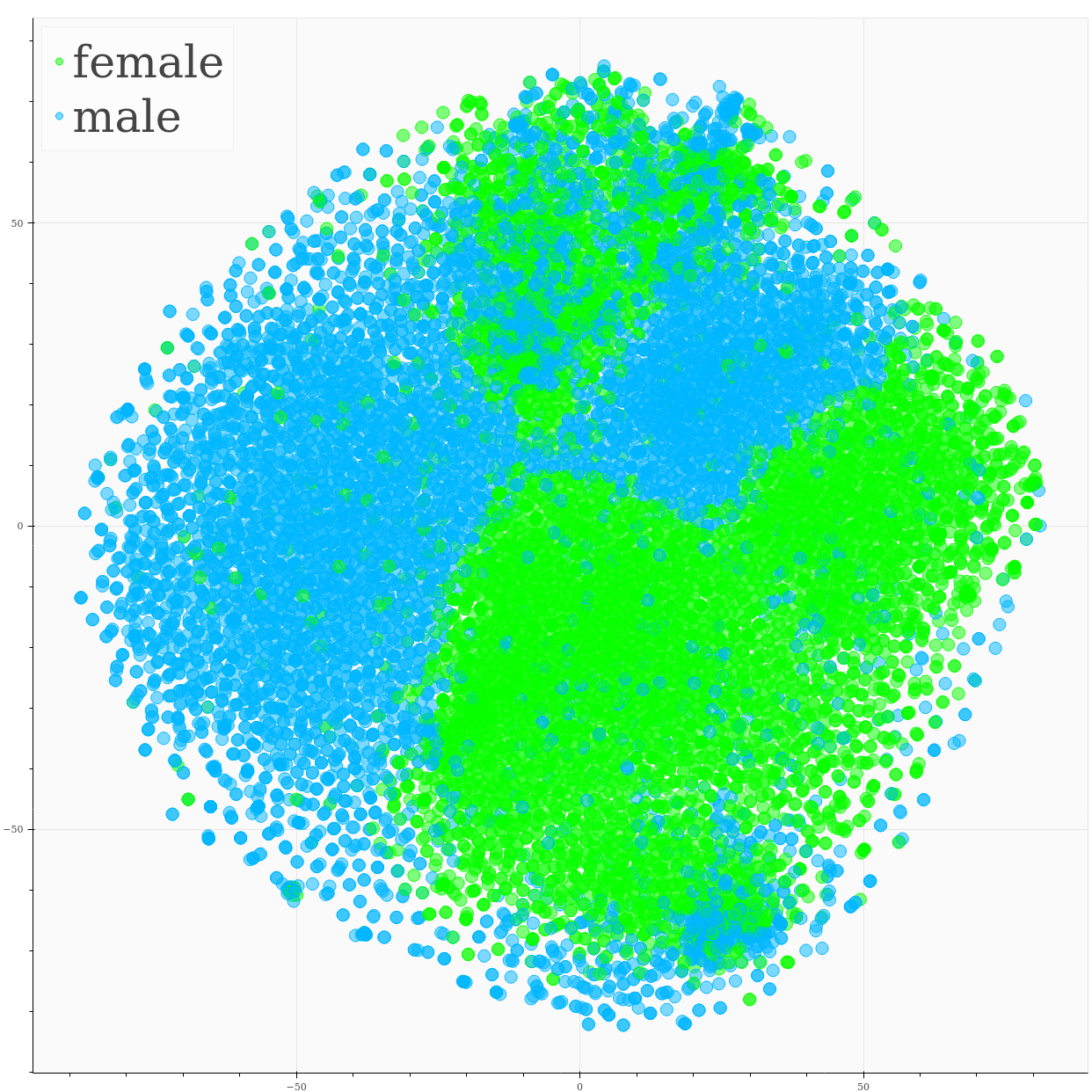}%
        \caption{}
        \label{fig:tsne_adaface_ir50_webface4m_gender}
    \end{subfigure}%
    \begin{subfigure}[h]{0.43\linewidth}
        \includegraphics[width=\textwidth]{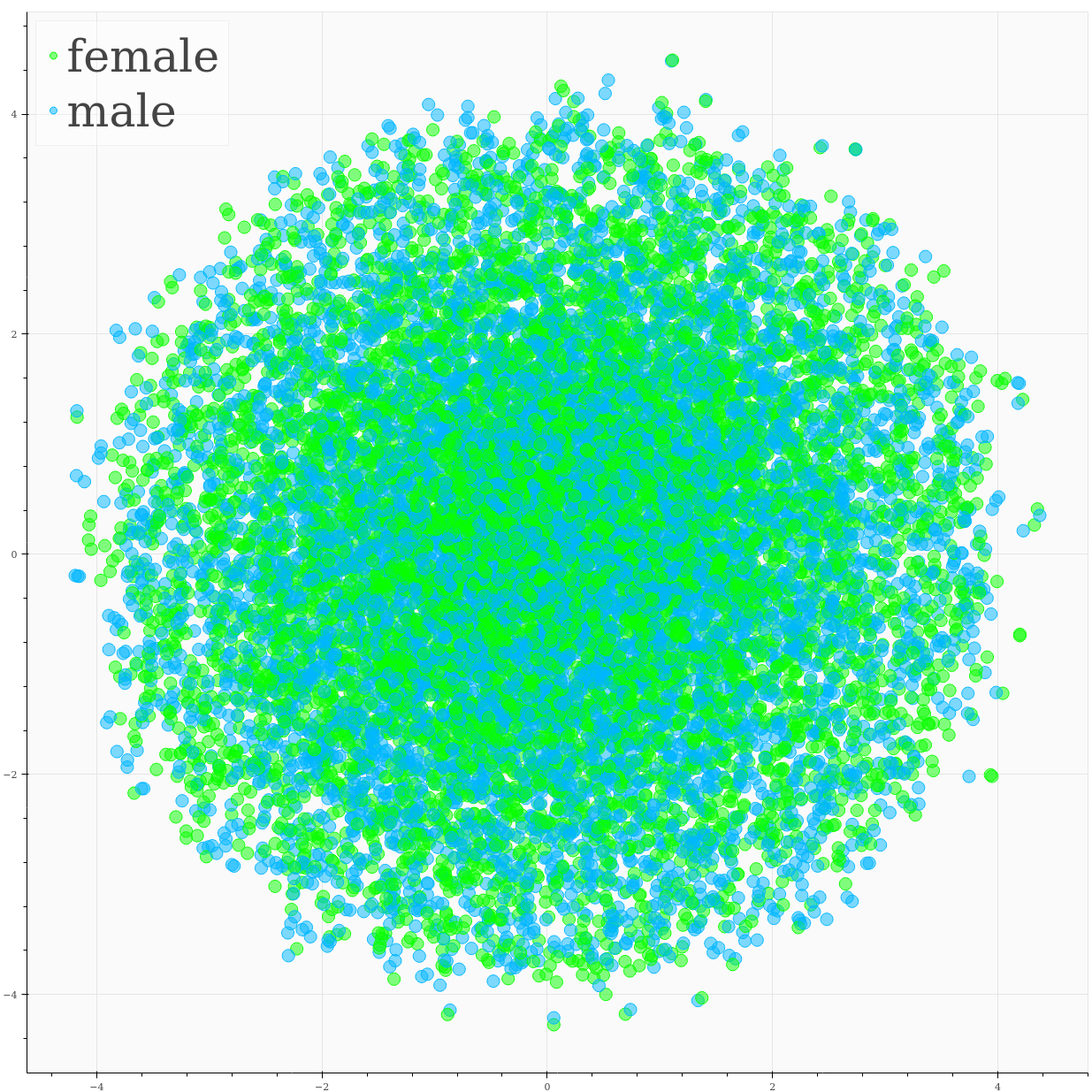}%
        \caption{}
        \label{fig:tsne_dvpf_adaface_ir50_webface4m_fairface_gender_10_UB_128_adaface_ir50_webface4m_gender}
    \end{subfigure}%
    \\
    \begin{subfigure}[h]{0.43\linewidth}
        \includegraphics[width=\textwidth]{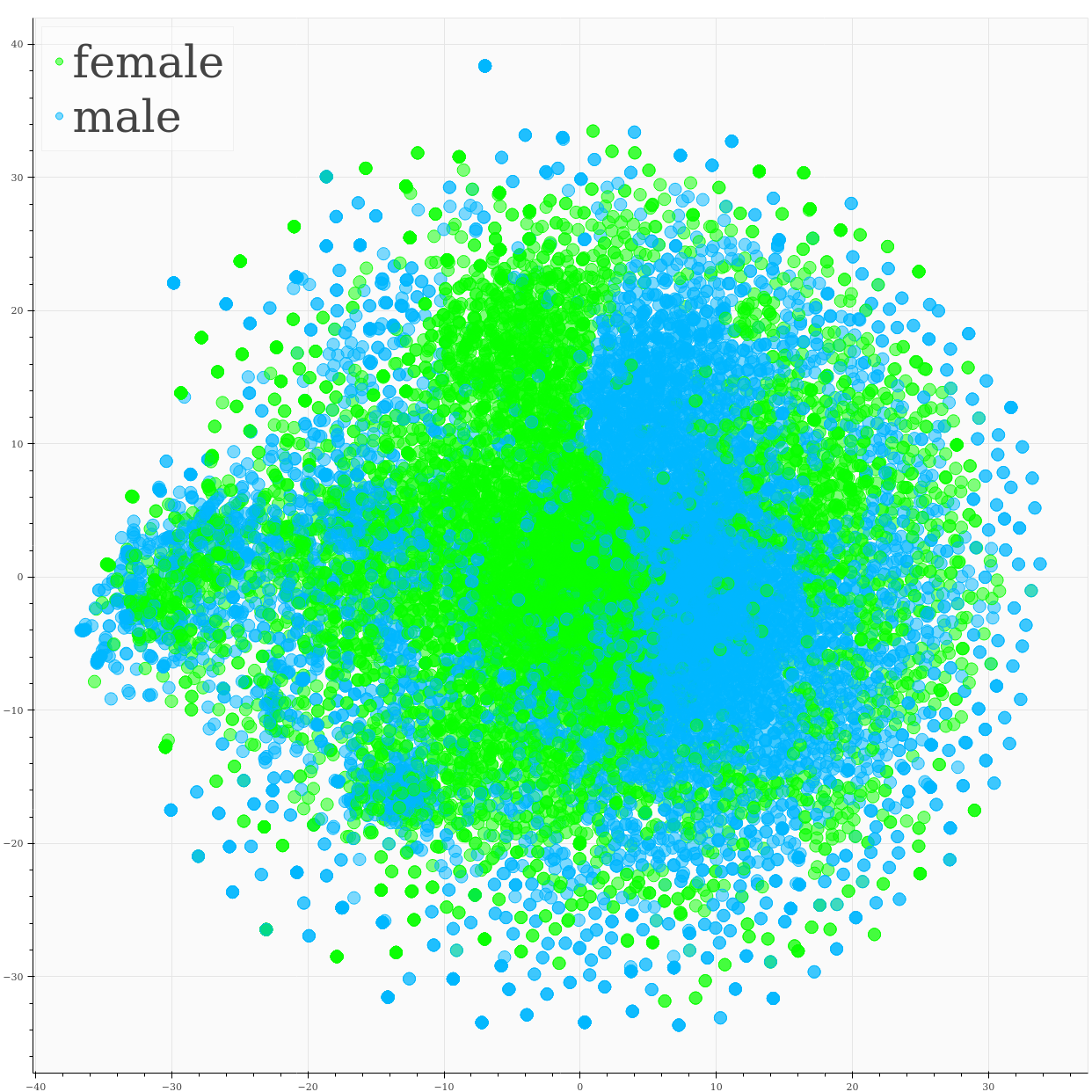}%
        \caption{}
        \label{fig:tsne_arcface_r50_gender}
    \end{subfigure}%
    \begin{subfigure}[h]{0.43\linewidth}
        \includegraphics[width=\textwidth]{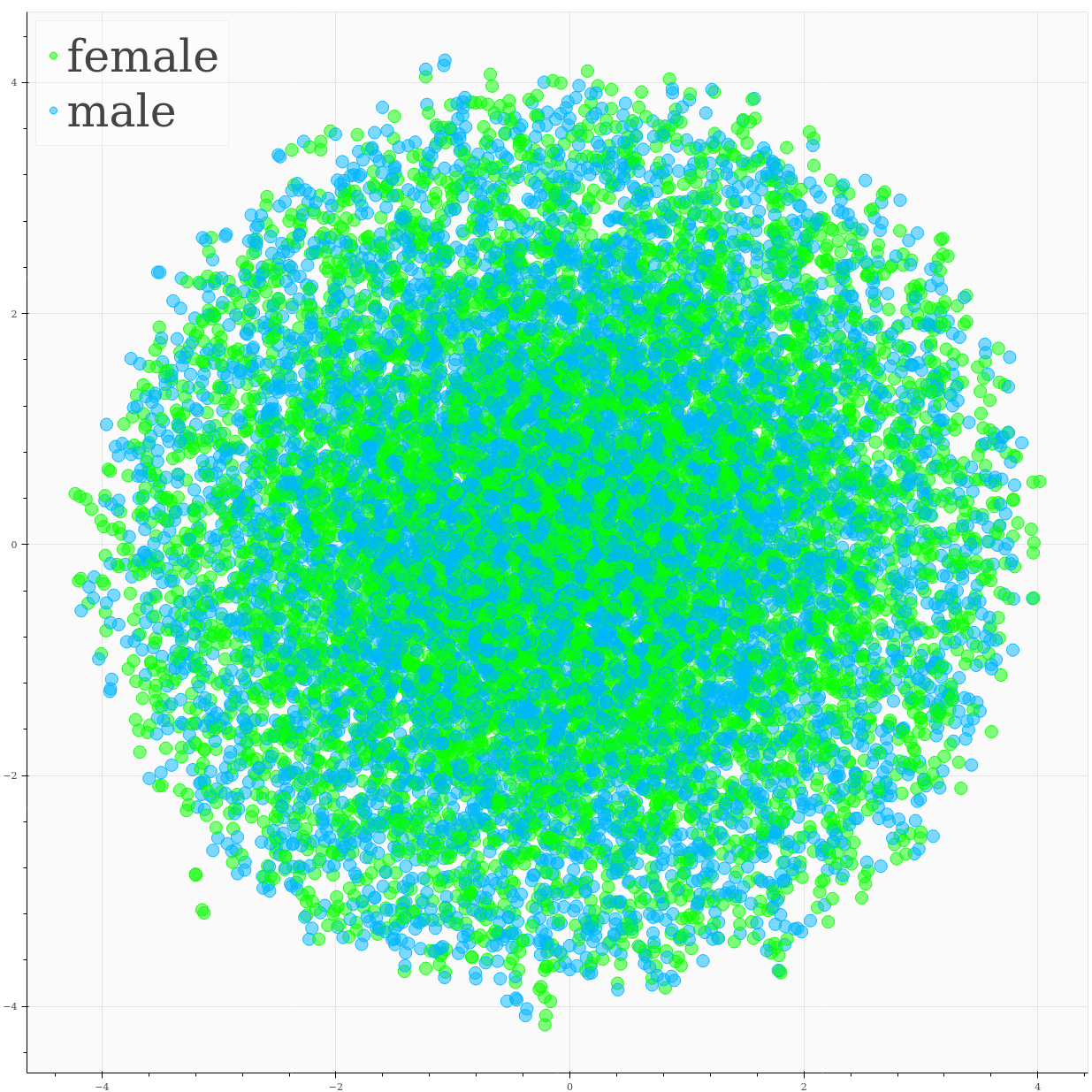}%
        \caption{}
        \label{fig:tsne_dvpf_arcface_r50_fairface_gender_10_UB_128_arcface_r50_MS1Mv3_gender}
    \end{subfigure}%
    %
    \caption{
    t-SNE visualizations of the FairFace dataset with $\mathbf{S}$ representing `gender', using the (\textsf{P2}) model, setting $\alpha = 10$ and $d_{\mathbf{z}} = 128$. The visualizations include:
    (a) AdaFace original (clean) embeddings, 
    (b) Post-DVPF AdaFace embeddings,
    (c) ArcFace original (clean) embeddings, 
    and 
    (d) Post-DVPF ArcFace embeddings.}
    \label{Fig:tSNE_on_sensitive_attribute_gender}
\end{figure}

\autoref{Fig:tSNE_on_sensitive_attribute_gender} provide qualitative visualization of the leakage in sensitive attribute classification on the FairFace database, both before and after applying the DVPF model with $\mathbf{S}$ set as gender.

%


\clearpage
\onecolumn

\appsection{Training Algorithms}

%
%
%
\begin{center}
\centering
\begin{spacing}{1}
\begin{algorithm}
    \setstretch{0.85}
    \begin{algorithmic}[1]
        \State \textbf{Input:} Training Dataset: $\{ \left( \mathbf{s}_n, \mathbf{x}_n  \right) \}_{n=1}^{N}$; Hyper-Parameter: $\alpha$
        \State $\boldsymbol{\phi}, \boldsymbol{\theta}, \boldsymbol{\psi} , \boldsymbol{\varphi}, \boldsymbol{\eta}, \boldsymbol{\omega}\; \gets$ Initialize Network Parameters
        
         \Repeat
        
         \hspace{-15pt}(1) {\footnotesize\textbf{\textsf{Train the Encoder $\boldsymbol{\phi}$, Utility Decoder $\boldsymbol{\theta}$, Uncertainty Decoder}} $\boldsymbol{\xi}$}
         \State  ~~Sample a mini-batch $\{ \mathbf{x}_m, \mathbf{s}_m \}_{m=1}^{M} \sim P_{\mathsf{D}} (\mathbf{X}) P_{\mathbf{S} 
         \mid \mathbf{X}}$
         \State  ~~Compute encoder outputs $\boldsymbol{\mu}_m^{\mathsf{enc}}, \boldsymbol{\sigma}_m^{\mathsf{enc}} = f_{\boldsymbol{\phi}} (\mathbf{x}_m), \forall m \in [M]$
         \State  ~~Apply reparametrization trick $\mathbf{z}_m^{\mathsf{enc}} = \boldsymbol{\mu}_m^{\mathsf{enc}} + \boldsymbol{\epsilon}_m  \odot \boldsymbol{\sigma}_m^{\mathsf{enc}}, \; \boldsymbol{\epsilon}_m \sim \mathcal{N}(0, \mathbf{I}) , \; \forall m \in [M]$
         %
         %
         \State  ~~Sample $\{ \mathbf{n}_m \}_{m=1}^{M} \sim \mathcal{N}(\boldsymbol{0}, \mathbf{I})$%
         \State  ~~Compute $\boldsymbol{\mu}_{m}^{\mathsf{prior}}, \boldsymbol{\sigma}_m^{\mathsf{prior}} = g_{\boldsymbol{\psi}} (\mathbf{n}_m), \forall m \in [M]$%
         \State  ~~Compute $\mathbf{z}_m^{\mathsf{prior}} \!\!=\! \boldsymbol{\mu}_m^{\mathsf{prior}} \! + \boldsymbol{\epsilon}_m'  \odot \boldsymbol{\sigma}_m^{\mathsf{prior}}, \boldsymbol{\epsilon}_m' \! \sim \! \mathcal{N}(0, \mathbf{I}), \forall m \! \in \! [M]\!$%
         \State  ~~Compute $\mathbf{\widehat{x}}_m =  g_{\boldsymbol{\theta}} ( \mathbf{z}_m^{\mathsf{enc}} ), \forall m \in [M]$%
         \State  ~~Compute $\mathbf{\widehat{s}}_m =  g_{\boldsymbol{\xi}} (\mathbf{z}_m^{\mathsf{enc}}), \forall m \in [M]$
         \State  ~~Back-propagate loss:\vspace{-10pt}
                %
                {\footnotesize \begin{equation}
                \quad \mathcal{L} \left( \boldsymbol{\phi},  \boldsymbol{\theta}, \boldsymbol{\xi} \right) = \! - \frac{1}{M} \sum_{m=1}^{M} \! \Big( \mathsf{dis}( \mathbf{x}_m , \mathbf{\widehat{x}}_m ) 
                 -  \, \alpha \; \log P_{\boldsymbol{\xi}} (\mathbf{s}_m \! \mid \! \mathbf{z}_m^{\mathsf{enc}}) 
                 \Big) \nonumber
                \end{equation}}

        \vspace{-3pt}
        
        \hspace{-15pt}(2) {\footnotesize\textbf{\textsf{Train the Latent Space Discriminator}} $ \boldsymbol{\eta} $}
        \State  ~~Sample $\{ \mathbf{x}_m \}_{m=1}^{M} \sim P_{\mathsf{D}} (\mathbf{X})$
        \State  ~~Sample $\{ \mathbf{n}_m \}_{m=1}^{M} \sim \mathcal{N}(\boldsymbol{0}, \mathbf{I})$
        \State  ~~Compute $\mathbf{z}_m^{\mathsf{enc}}\!$ from $\!f_{\! \boldsymbol{\phi}} (\mathbf{x}_m)\!$ with reparametrization, $\forall m \!\in \![M]\!$
        \State  ~~Compute $\mathbf{z}_m^{\mathsf{prior}}\!\!$ from $\! g_{\boldsymbol{\psi}} (\mathbf{n}_m \!)\!$ with reparametrization, $\! \forall m \! \in \![M]\!$
        \State  ~~Back-propagate loss:\vspace{-5pt}
                {\footnotesize \begin{equation}
                \;\;\; \mathcal{L} \left( \boldsymbol{\eta} \right) =  - \frac{\alpha}{M} \; \sum_{m=1}^{M}  \log D_{\boldsymbol{\eta}} (\mathbf{z}_m^{\mathsf{enc}}) + \log \big( 1- D_{\boldsymbol{\eta}} (\, \mathbf{z}_m^{\mathsf{prior}} \,) \big) \nonumber 
                \end{equation}}
          
        \vspace{-3pt}
        
        \hspace{-15pt}(3) {\footnotesize\textbf{\textsf{Train the Encoder $\boldsymbol{\phi}$ and Prior Distribution Generator $\boldsymbol{\psi} $ Adversarially}}} 
        \State  ~~Sample $\{ \mathbf{x}_m \}_{m=1}^{M} \sim P_{\mathsf{D}} (\mathbf{X})$
        \State  ~~Compute $\mathbf{z}_m^{\mathsf{enc}}\!$ from $\!f_{\! \boldsymbol{\phi}} (\mathbf{x}_m)\!$ with reparametrization, $\forall m \!\in \![M]$
        \State  ~~Sample $\{ \mathbf{n}_m \}_{m=1}^{M} \sim \mathcal{N}(\boldsymbol{0}, \mathbf{I})$
        \State  ~~Compute $\mathbf{z}_m^{\mathsf{prior}}\!\!$ from $\! g_{ \boldsymbol{\psi}} (\mathbf{n}_m \!)\!$ with reparametrization, $\! \forall m \! \in \![M]\!$
        \State  ~~Back-propagate loss:\vspace{-5pt}
                {\footnotesize \begin{equation}
                \;\;\; \mathcal{L} \left( \boldsymbol{\phi}, \boldsymbol{\psi} \right) = \frac{\alpha}{M} \;   \sum_{m=1}^{M}  \log D_{\boldsymbol{\eta}} (\mathbf{z}_m^{\mathsf{enc}})  + \log \big( 1- D_{\boldsymbol{\eta}} (\, \mathbf{z}_m^{\mathsf{prior}} \,) \big) \nonumber 
                \end{equation}}
        
        \vspace{-3pt}  
        
        \hspace{-15pt}(4) {\footnotesize\textbf{\textsf{Train the Utility Output Space Discriminator}} $ \boldsymbol{\omega} $}
        \State  ~~Sample $\{ \mathbf{x}_m \}_{m=1}^{M} \sim P_{\mathsf{D}} (\mathbf{X})$
        \State  ~~Sample $\{ \mathbf{n}_m \}_{m=1}^{M} \sim \mathcal{N} \! \left( \boldsymbol{0}, \mathbf{I}\right)$
        \State ~~Compute $\mathbf{z}_m^{\mathsf{prior}}\!\!$ from $\! g_{ \boldsymbol{\psi}} (\mathbf{n}_m \!)\!$ with reparametrization, $\! \forall m \! \in \![M]\!$
        \State  ~~Compute $\mathbf{\widehat{x}}_m = g_{\boldsymbol{\theta}} (\mathbf{z}_m^{\mathsf{prior}} ), \forall m \in [M]$
        \State  ~~Back-propagate loss:\vspace{-6pt}
                {\footnotesize \begin{equation}
                \mathcal{L} \left( \boldsymbol{\omega} \right) = -  \frac{1}{M} \sum_{m=1}^{M} \log D_{\boldsymbol{\omega}} (\mathbf{x}_m)  +  \log \left( 1- D_{\boldsymbol{\omega}} (\, \mathbf{\widehat{x}}_m \,) \right) \nonumber
                \end{equation}}
        
        \vspace{-3pt} 
        
        \hspace{-15pt}(5) {\footnotesize\textbf{\textsf{Train the Prior Distribution Generator $\boldsymbol{\psi}$, Utility Decoder $\boldsymbol{\theta}$, and Uncertainty Decoder $\boldsymbol{\xi}$ Adversarially}}}
        %
        \State  ~~Sample $\{ \mathbf{n}_m \}_{m=1}^{M} \sim \mathcal{N} \! \left( \boldsymbol{0}, \mathbf{I}\right)$
        \State ~~Compute $\mathbf{z}_m^{\mathsf{prior}}\!\!$ from $\! g_{ \boldsymbol{\psi}} (\mathbf{n}_m \!)\!$ with reparametrization, $\! \forall m \! \in \![M]\!$
        \State  ~~Compute $\mathbf{\widehat{x}}_m = g_{\boldsymbol{\theta}} \left( \mathbf{z}_m^{\mathsf{prior}} \right), \forall m \in [M]$
        \State  ~~Compute $\mathbf{\widehat{s}}_m = g_{\boldsymbol{\xi}} \left( \mathbf{z}_m^{\mathsf{prior}} \right), \forall m \in [M]$
        \State  ~~Back-propagate loss:\vspace{-5pt}
                {\footnotesize \begin{equation}
                \;\;\; \;\mathcal{L} \left( \boldsymbol{\psi}, \boldsymbol{\theta}, \boldsymbol{\xi} \right)\! = \! \frac{1}{M} \!\! \sum_{m=1}^{M} \!\! \log \left( 1 \! - \! D_{\boldsymbol{\omega}} (\, \mathbf{\widehat{x}}_m \,) \right) +  \log \left( 1 \! - \! D_{\boldsymbol{\tau}} (\, \mathbf{\widehat{s}}_m \,) \right)\!\!\!\! \nonumber
                \end{equation}}
        
        \vspace{-3pt}

        \hspace{-15pt}(6) {\footnotesize\textbf{\textsf{Train Uncertainty Output Space Discriminator $\boldsymbol{\omega}$}}}
        \State  ~~Sample a mini-batch $\{ \mathbf{s}_m, \mathbf{x}_m \}_{m=1}^{M} \sim P_{\mathsf{D}} (\mathbf{X}) P_{\mathbf{S} \mid \mathbf{X}}$
        \State  ~~Sample $\{ \mathbf{n}_m \}_{m=1}^{M} \sim \mathcal{N}(\boldsymbol{0}, \mathbf{I})$
        \State  ~~Compute $\mathbf{z}_m^{\mathsf{prior}}\!\!$ from $\! g_{ \boldsymbol{\psi}} (\mathbf{n}_m \!)\!$ with reparametrization, $\! \forall m \! \in \![M]\!$
        \State  ~~Compute $\mathbf{\widehat{s}}_m \sim g_{\boldsymbol{\xi}} \left( \mathbf{z}_m^{\mathsf{prior}} \right), \forall m \in [M]$
        \State  ~~Back-propagate loss:\vspace{-5pt}
                {\footnotesize \begin{equation}
                \mathcal{L} \left( \boldsymbol{\tau} \right) =  \frac{1}{M} \sum_{m=1}^{M} \log D_{\boldsymbol{\tau}} (\, \mathbf{s}_m \,)  +  \log \left( 1- D_{\boldsymbol{\tau}} (\, \mathbf{\widehat{s}}_m \,) \right) \nonumber
                \end{equation}}

        \vspace{-8pt}
        \Until{Convergence}
        \State \textbf{return} $\boldsymbol{\phi}, \boldsymbol{\theta}, \boldsymbol{\psi}, \boldsymbol{\varphi}, \boldsymbol{\eta}, \boldsymbol{\omega}$
    \end{algorithmic}
    \caption{\small Deep Variational $\mathsf{DisPF}$ training algorithm associated with $\textsf{DisPF\text{-}MI}\; (\textsf{P1})$.}
    \label{Algorithm:P1_DisPF}
\end{algorithm}
\end{spacing}
\end{center}

%
%
%
\begin{center}
\centering
\vspace{-40pt}
\begin{spacing}{1}
\begin{algorithm}
    \begin{algorithmic}[1]
        \State \textbf{Input:} Training Dataset: $\{ \left( \mathbf{s}_n, \mathbf{x}_n  \right) \}_{n=1}^{N}$; Hyper-Parameter: $\alpha$
        \State $\boldsymbol{\phi}, \boldsymbol{\theta}, \boldsymbol{\psi} , \boldsymbol{\varphi}, \boldsymbol{\eta}, \boldsymbol{\omega}\; \gets$ Initialize Network Parameters
        
         \Repeat
        
         \hspace{-15pt}(1) {\footnotesize\textbf{\textsf{Train the Encoder $\boldsymbol{\phi}$, Utility Decoder $\boldsymbol{\theta}$, Uncertainty Decoder}} $\boldsymbol{\varphi}$}
         \State  ~~Sample a mini-batch $\{ \mathbf{x}_m, \mathbf{s}_m \}_{m=1}^{M} \sim P_{\mathsf{D}} (\mathbf{X}) P_{\mathbf{S} 
         \mid \mathbf{X}}$
         \State  ~~Compute encoder outputs $\boldsymbol{\mu}_m^{\mathsf{enc}}, \boldsymbol{\sigma}_m^{\mathsf{enc}} = f_{\boldsymbol{\phi}} (\mathbf{x}_m), \forall m \in [M]$
         \State  ~~Apply reparametrization trick $\mathbf{z}_m^{\mathsf{enc}} = \boldsymbol{\mu}_m^{\mathsf{enc}} + \boldsymbol{\epsilon}_m  \odot \boldsymbol{\sigma}_m^{\mathsf{enc}}, \; \boldsymbol{\epsilon}_m \sim \mathcal{N}(0, \mathbf{I}) , \; \forall m \in [M]$
         %
         %
         \State  ~~Sample $\{ \mathbf{n}_m \}_{m=1}^{M} \sim \mathcal{N}(\boldsymbol{0}, \mathbf{I})$%
         \State  ~~Compute $\boldsymbol{\mu}_{m}^{\mathsf{prior}}, \boldsymbol{\sigma}_m^{\mathsf{prior}} = g_{\boldsymbol{\psi}} (\mathbf{n}_m), \forall m \in [M]$%
         \State  ~~Compute $\mathbf{z}_m^{\mathsf{prior}} \!\!=\! \boldsymbol{\mu}_m^{\mathsf{prior}} \! + \boldsymbol{\epsilon}_m'  \odot \boldsymbol{\sigma}_m^{\mathsf{prior}}, \boldsymbol{\epsilon}_m' \! \sim \! \mathcal{N}(0, \mathbf{I}), \forall m \! \in \! [M]\!$%
         \State  ~~Compute $\mathbf{\widehat{x}}_m =  g_{\boldsymbol{\theta}} ( \mathbf{z}_m^{\mathsf{enc}} ), \forall m \in [M]$%
         \State  ~~Compute $\mathbf{\widetilde{x}}_m =  g_{\boldsymbol{\varphi}} (\mathbf{z}_m^{\mathsf{enc}}, \mathbf{s}_m), \forall m \in [M]$
         \State  ~~Back-propagate loss:\vspace{-10pt}
                %
                 \begin{equation}
                \quad \mathcal{L} \left( \boldsymbol{\phi},  \boldsymbol{\theta}, \boldsymbol{\varphi} \right) = \! - \frac{1}{M} \sum_{m=1}^{M} \! \Big( \mathsf{dis}( \mathbf{x}_m , \mathbf{\widehat{x}}_m ) 
                 -  \alpha  \, \D_{\mathrm{KL}} \! \left( P_{\boldsymbol{\phi}} (\mathbf{z}_m^{\mathsf{enc}} \! \mid \! \mathbf{x}_m ) \Vert  Q_{\boldsymbol{\psi}} (\mathbf{z}_m^{\mathsf{prior}}) \right) 
                 +  \, \alpha \, \mathsf{dis}( \mathbf{x}_m , \mathbf{\widetilde{x}}_m )  \Big) \nonumber
                \end{equation}

        \vspace{-3pt}
        
        \hspace{-15pt}(2) {\footnotesize\textbf{\textsf{Train the Latent Space Discriminator}} $ \boldsymbol{\eta} $}
        \State  ~~Sample $\{ \mathbf{x}_m \}_{m=1}^{M} \sim P_{\mathsf{D}} (\mathbf{X})$
        \State  ~~Sample $\{ \mathbf{n}_m \}_{m=1}^{M} \sim \mathcal{N}(\boldsymbol{0}, \mathbf{I})$
        \State  ~~Compute $\mathbf{z}_m^{\mathsf{enc}}\!$ from $\!f_{\! \boldsymbol{\phi}} (\mathbf{x}_m)\!$ with reparametrization, $\forall m \!\in \![M]\!$
        \State  ~~Compute $\mathbf{z}_m^{\mathsf{prior}}\!\!$ from $\! g_{\boldsymbol{\psi}} (\mathbf{n}_m \!)\!$ with reparametrization, $\! \forall m \! \in \![M]\!$
        \State  ~~Back-propagate loss:\vspace{-5pt}
                {\footnotesize \begin{equation}
                \;\;\; \mathcal{L} \left( \boldsymbol{\eta} \right) =  - \frac{\alpha}{M} \; \sum_{m=1}^{M}  \log D_{\boldsymbol{\eta}} (\mathbf{z}_m^{\mathsf{enc}}) + \log \big( 1- D_{\boldsymbol{\eta}} (\, \mathbf{z}_m^{\mathsf{prior}} \,) \big) \nonumber 
                \end{equation}}
          
        \vspace{-3pt}
        
        \hspace{-15pt}(3) {\footnotesize\textbf{\textsf{Train the Encoder $\boldsymbol{\phi}$ and Prior Distribution Generator $\boldsymbol{\psi} $ Adversarially}}} 
        \State  ~~Sample $\{ \mathbf{x}_m \}_{m=1}^{M} \sim P_{\mathsf{D}} (\mathbf{X})$
        \State  ~~Compute $\mathbf{z}_m^{\mathsf{enc}}\!$ from $\!f_{\! \boldsymbol{\phi}} (\mathbf{x}_m)\!$ with reparametrization, $\forall m \!\in \![M]$
        \State  ~~Sample $\{ \mathbf{n}_m \}_{m=1}^{M} \sim \mathcal{N}(\boldsymbol{0}, \mathbf{I})$
        \State  ~~Compute $\mathbf{z}_m^{\mathsf{prior}}\!\!$ from $\! g_{ \boldsymbol{\psi}} (\mathbf{n}_m \!)\!$ with reparametrization, $\! \forall m \! \in \![M]\!$
        \State  ~~Back-propagate loss:\vspace{-5pt}
                {\footnotesize \begin{equation}
                \;\;\; \mathcal{L} \left( \boldsymbol{\phi}, \boldsymbol{\psi} \right) = \frac{\alpha}{M} \;   \sum_{m=1}^{M}  \log D_{\boldsymbol{\eta}} (\mathbf{z}_m^{\mathsf{enc}})  + \log \big( 1- D_{\boldsymbol{\eta}} (\, \mathbf{z}_m^{\mathsf{prior}} \,) \big) \nonumber 
                \end{equation}}
        
        \vspace{-3pt}  
        
        \hspace{-15pt}(4) {\footnotesize\textbf{\textsf{Train the Utility Output Space Discriminator}} $ \boldsymbol{\omega} $}
        \State  ~~Sample $\{ \mathbf{x}_m \}_{m=1}^{M} \sim P_{\mathsf{D}} (\mathbf{X})$
        \State  ~~Sample $\{ \mathbf{n}_m \}_{m=1}^{M} \sim \mathcal{N} \! \left( \boldsymbol{0}, \mathbf{I}\right)$
        \State ~~Compute $\mathbf{z}_m^{\mathsf{prior}}\!\!$ from $\! g_{ \boldsymbol{\psi}} (\mathbf{n}_m \!)\!$ with reparametrization, $\! \forall m \! \in \![M]\!$
        \State  ~~Compute $\mathbf{\widehat{x}}_m = g_{\boldsymbol{\theta}} (\mathbf{z}_m^{\mathsf{prior}} ), \forall m \in [M]$
        \State  ~~Back-propagate loss:\vspace{-6pt}
                {\footnotesize \begin{equation}
                \mathcal{L} \left( \boldsymbol{\omega} \right) = -  \frac{1}{M} \sum_{m=1}^{M} \log D_{\boldsymbol{\omega}} (\mathbf{x}_m)  +  \log \left( 1- D_{\boldsymbol{\omega}} (\, \mathbf{\widehat{x}}_m \,) \right) \nonumber
                \end{equation}}
        
        \vspace{-3pt} 
        
        \hspace{-15pt}(5) {\footnotesize\textbf{\textsf{Train the Prior Distribution Generator $\boldsymbol{\psi}$, Utility Decoder $\boldsymbol{\theta}$, and Uncertainty Decoder $\boldsymbol{\varphi}$ Adversarially}}}
        %
        \State  ~~Sample $\{ \mathbf{n}_m \}_{m=1}^{M} \sim \mathcal{N} \! \left( \boldsymbol{0}, \mathbf{I}\right)$
        \State ~~Compute $\mathbf{z}_m^{\mathsf{prior}}\!\!$ from $\! g_{ \boldsymbol{\psi}} (\mathbf{n}_m \!)\!$ with reparametrization, $\! \forall m \! \in \![M]\!$
        \State  ~~Compute $\mathbf{\widehat{x}}_m \sim g_{\boldsymbol{\theta}} \left( \mathbf{z}_m^{\mathsf{prior}} \right), \forall m \in [M]$
        \State  ~~Compute $\mathbf{\widetilde{x}}_m \sim g_{\boldsymbol{\varphi}} \left( \mathbf{z}_m^{\mathsf{prior}} , \mathbf{s}_m \right), \forall m \in [M]$
        \State  ~~Back-propagate loss:\vspace{-5pt}
                {\footnotesize \begin{equation}
                \;\;\; \;\mathcal{L} \left( \boldsymbol{\psi}, \boldsymbol{\theta}, \boldsymbol{\varphi} \right)\! = \! \frac{1}{M} \!\! \sum_{m=1}^{M} \!\! \log \left( 1 \! - \! D_{\boldsymbol{\omega}} (\, \mathbf{\widehat{x}}_m \,) \right) +  \log \left( 1 \! - \! D_{\boldsymbol{\omega}} (\, \mathbf{\widetilde{x}}_m \,) \right)\!\!\!\! \nonumber
                \end{equation}}
        
        \vspace{-3pt}

        \hspace{-15pt}(6) {\footnotesize\textbf{\textsf{Train Uncertainty Output Space Discriminator $\boldsymbol{\omega}$}}}
        \State  ~~Sample a mini-batch $\{ \mathbf{s}_m, \mathbf{x}_m \}_{m=1}^{M} \sim P_{\mathsf{D}} (\mathbf{X}) P_{\mathbf{S} \mid \mathbf{X}}$
        \State  ~~Sample $\{ \mathbf{n}_m \}_{m=1}^{M} \sim \mathcal{N}(\boldsymbol{0}, \mathbf{I})$
        \State  ~~Compute $\mathbf{z}_m^{\mathsf{prior}}\!\!$ from $\! g_{ \boldsymbol{\psi}} (\mathbf{n}_m \!)\!$ with reparametrization, $\! \forall m \! \in \![M]\!$
        \State  ~~Compute $\mathbf{\widetilde{x}}_m \sim g_{\boldsymbol{\varphi}} \left( \mathbf{z}_m^{\mathsf{prior}} , \mathbf{s}_m \right), \forall m \in [M]$
        \State  ~~Back-propagate loss:\vspace{-5pt}
                {\footnotesize \begin{equation}
                \mathcal{L} \left( \boldsymbol{\omega} \right) =  \frac{1}{M} \sum_{m=1}^{M} \log D_{\boldsymbol{\omega}} (\, \mathbf{x}_m \,)  +  \log \left( 1- D_{\boldsymbol{\omega}} (\, \mathbf{\widetilde{x}}_m \,) \right) \nonumber
                \end{equation}}

        \vspace{-8pt}
        \Until{Convergence}
        \State \textbf{return} $\boldsymbol{\phi}, \boldsymbol{\theta}, \boldsymbol{\psi}, \boldsymbol{\varphi}, \boldsymbol{\eta}, \boldsymbol{\omega}$
    \end{algorithmic}
    \caption{\small Deep Variational $\mathsf{GenPF}$ training algorithm associated with $\mathsf{GenPF\text{-}MI}\; (\textsf{P2})$}
    \label{Algorithm:P2_GenPF}
\end{algorithm}
\end{spacing}
\end{center}

\clearpage
\onecolumn
%
%
\appsection{Deep Private Feature Extraction/Generation Experiment}

%
\begin{figure*}[h]
    \centering
    \includegraphics[width=0.95\textwidth]{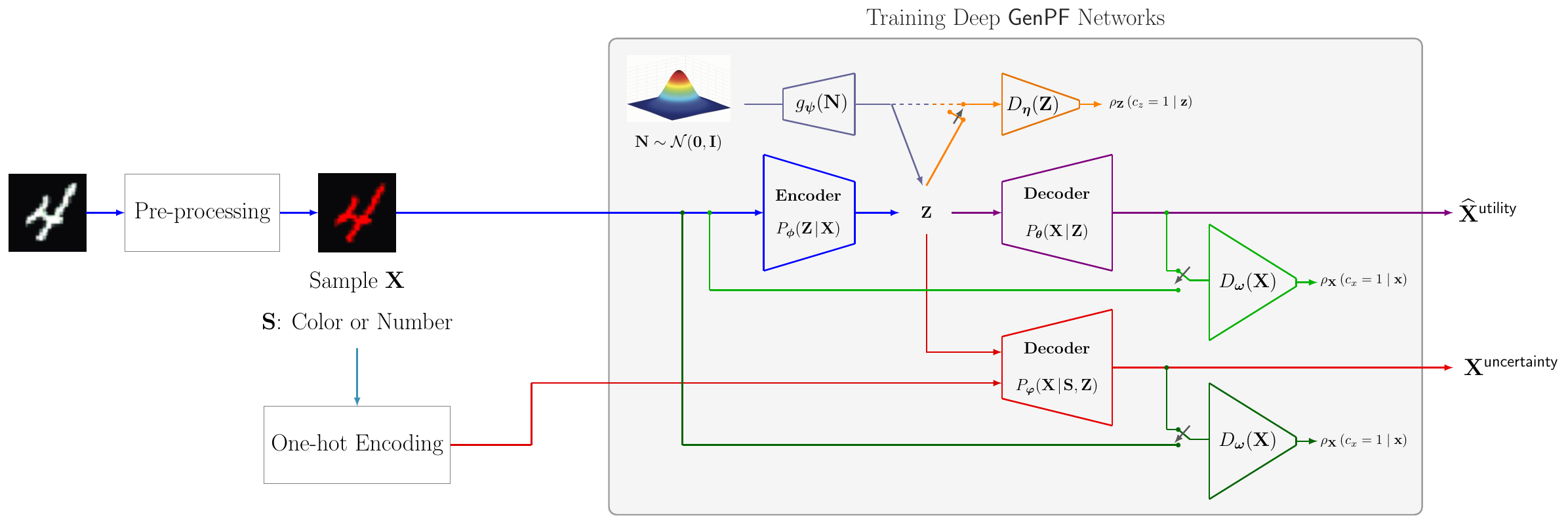}%
    \vspace{-5pt} 
    \caption{Training the deep variational $\mathsf{GenPF}$ model on Colored-MNIST dataset, employing the learning scenario `End-to-End Raw Data Scratch Learning'.}
    \label{TrainDVPF_CooredMNIST_RawDataScratch}
\vspace{-5pt} 
\end{figure*} 

\vspace{10pt}

%
\begin{figure*}[h]
    \centering
    \includegraphics[width=0.95\textwidth]{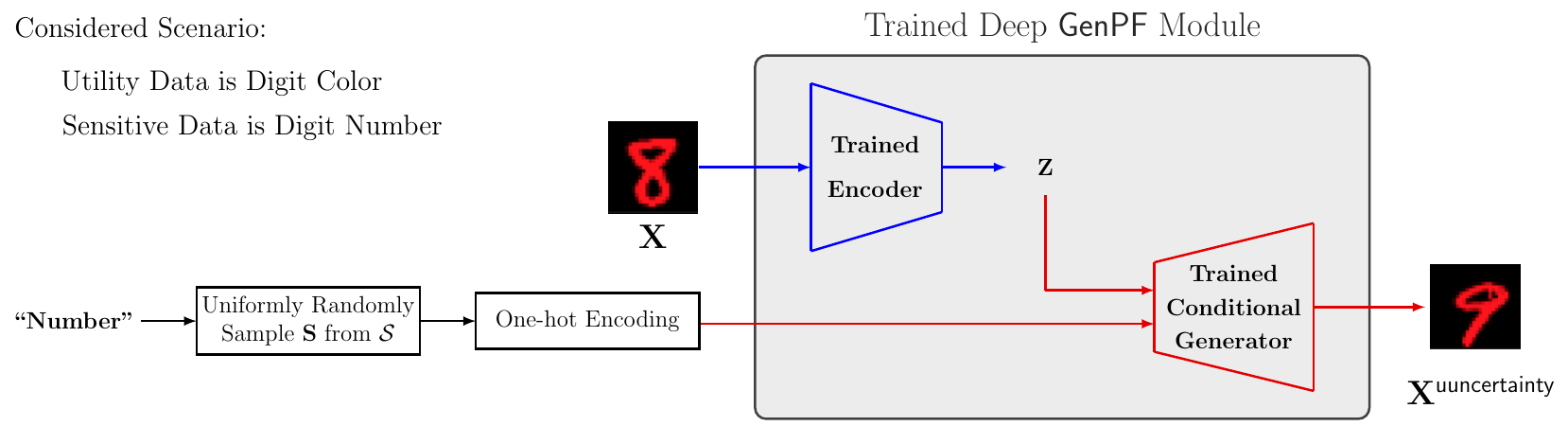}%
    \vspace{-5pt} 
    \caption{Evaluating the performance of the deep variational GenPF model, trained on the Colored-MNIST dataset, with the ``digit number'' as the ``sensitive attribute'' and the ``digit color'' as the ``useful data''.}
    \label{InferenceGenPF_ColoredMNIST_RawDataScratch_SNumber}
\vspace{-5pt} 
\end{figure*} 

\vspace{10pt}

%
\begin{figure*}[h]
    \centering
    \includegraphics[width=0.95\textwidth]{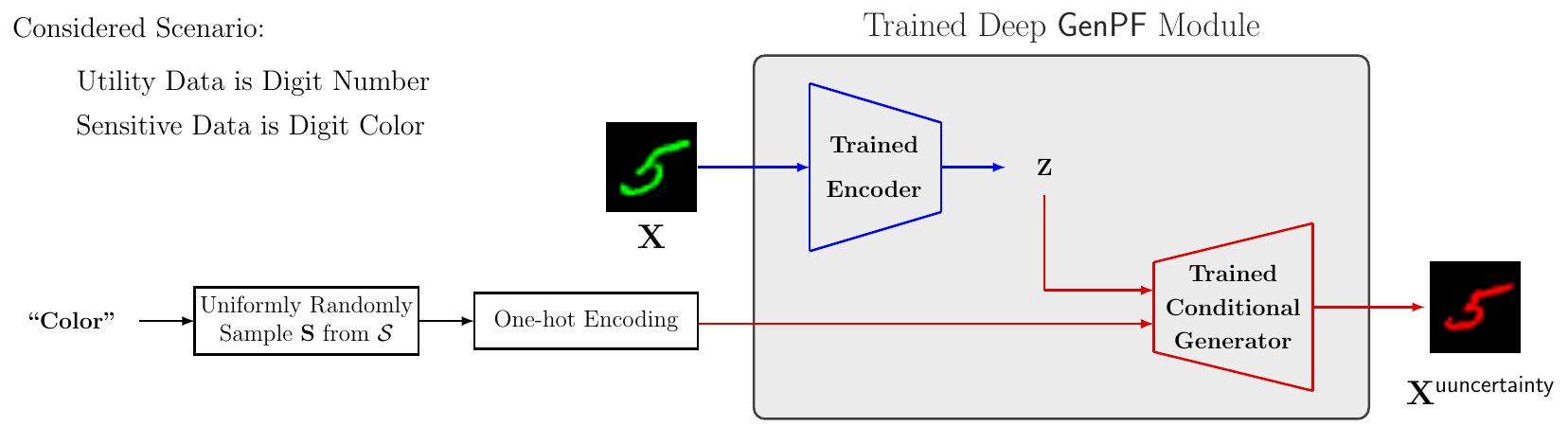}%
    \vspace{-5pt} 
    \caption{Evaluating the performance of the deep variational GenPF model, trained on the Colored-MNIST dataset, with the ``digit color'' as the ``sensitive attribute'' and the ``digit number'' as the ``useful data''.}
    \label{InferenceGenPF_ColoredMNIST_RawDataScratch_SColor}
\vspace{-5pt} 
\end{figure*} 


\clearpage
%
\begin{figure*}[h]
\centering
\includegraphics[width=0.68\textwidth]{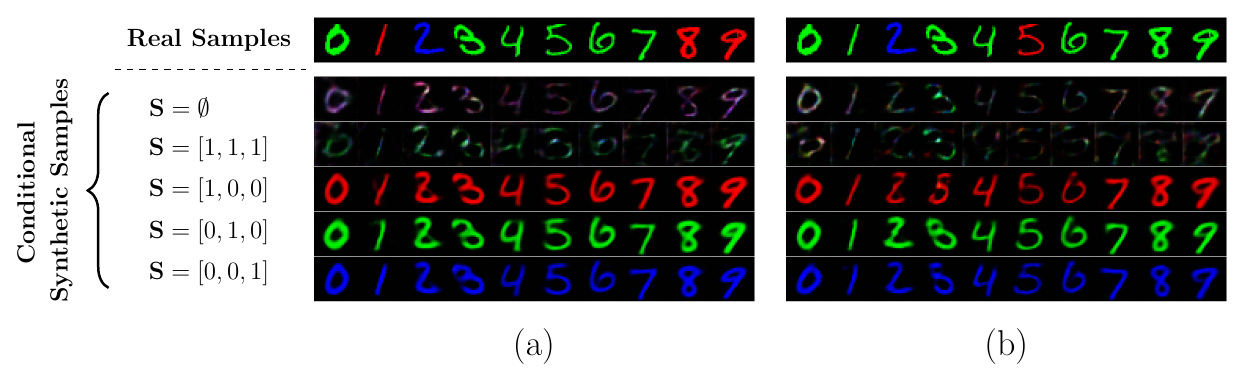}
\vspace{-5pt}
\caption{Qualitative evaluation of privacy-preserving synthetic samples $\mathbf{X}^{\texttt{uncertainty}}$ generated by the conditional generator $g_{\boldsymbol{\varphi}}$, using a custom Colored-MNIST dataset, where the sensitive attribute under consideration is the digit color.
The setting is defined with $d_{\mathbf{z}}=8$. For scenario (a), the color probabilities are set as $P_S (\mathsf{Red}) = \frac{1}{2}$, $P_S (\mathsf{Green}) = \frac{1}{6}$, and $P_S (\mathsf{Blue}) = \frac{1}{3}$. In scenario (b), all probabilities are equal with $P_S (\mathsf{Red}) = P_S (\mathsf{Green}) = P_S (\mathsf{Blue}) = \frac{1}{3}$.}
\label{Fig:QualitativeEvaluation_Colored-MNIST_S_DigitColor}
\end{figure*}

%
\begin{figure*}[h]
\centering
\includegraphics[width=0.68\textwidth]{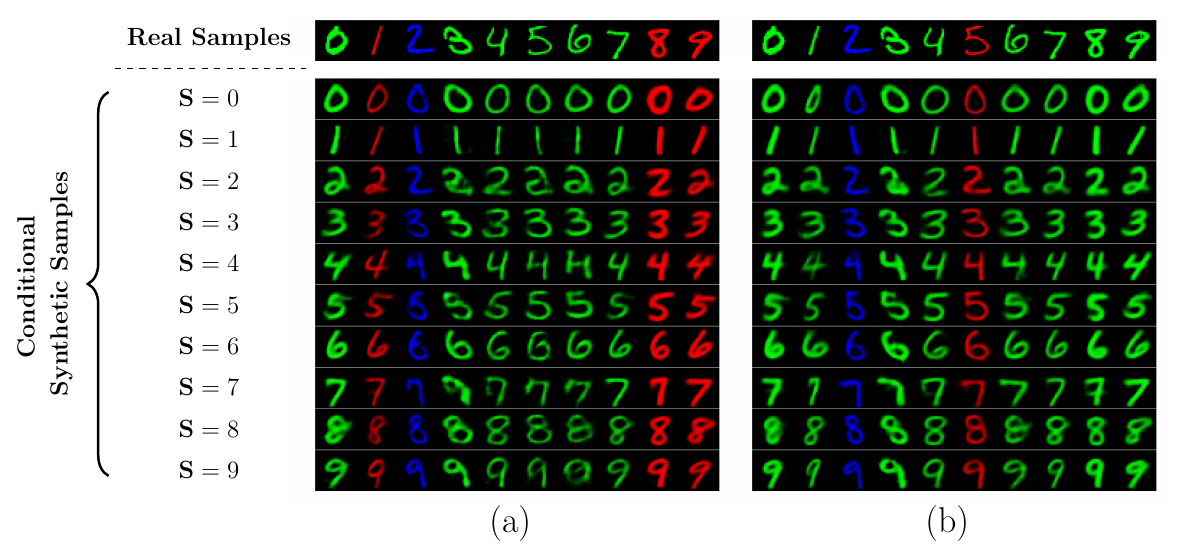}
\vspace{-5pt}
\caption{Qualitative evaluation of privacy-preserving synthetic samples $\mathbf{X}^{\texttt{uncertainty}}$ generated by the conditional generator $g_{\boldsymbol{\varphi}}$, using a custom Colored-MNIST dataset, where the sensitive attribute under consideration is the digit number.
The setting is defined with $d_{\mathbf{z}}=8$. For scenario (a), the color probabilities are set as $P_S (\mathsf{Red}) = \frac{1}{2}$, $P_S (\mathsf{Green}) = \frac{1}{6}$, and $P_S (\mathsf{Blue}) = \frac{1}{3}$. In scenario (b), all probabilities are equal with $P_S (\mathsf{Red}) = P_S (\mathsf{Green}) = P_S (\mathsf{Blue}) = \frac{1}{3}$.}
\label{Fig:QualitativeEvaluation_Colored-MNIST_S_DigitNumber}
\end{figure*}

\twocolumn

\clearpage

\renewcommand{\refname}{Appendix References}

\putbib

\end{bibunit}

\end{document}